\documentclass[mnsc,non-blindrev]{cls/informs3} 

 \DoubleSpacedXI 

\usepackage{tikz}
\usetikzlibrary{arrows.meta, positioning}

\usepackage[textsize=tiny]{todonotes}
\PassOptionsToPackage{usenames,dvipsnames}{xcolor} 

\newcommand{\Jingwei}[1]{{\textcolor{green}{{\textsf{Jingwei: #1}}}}}

\usepackage{booktabs}
\usepackage{endnotes}
\usepackage{bbm}
\usepackage[ruled]{algorithm}
\usepackage{algorithmicx}
\usepackage{algpseudocode}
\usepackage{graphicx}
\usepackage{epstopdf}
\usepackage{subfigure,caption,enumitem,thm-restate}
\usepackage{enumitem}
\setlist{leftmargin=*}
\let\footnote=\endnote

\usepackage{natbib}
 \bibpunct[, ]{(}{)}{,}{a}{}{,}%
 \def\bibfont{\small}%
\usepackage{bm}
\usepackage{amsmath}
\usepackage{amssymb}
\usepackage{comment}

\usepackage[colorlinks=true, linkcolor=blue, citecolor=blue]{hyperref} 

\newenvironment{myproof}{ \paragraph{Proof: } } {\hfill$\square$}
\newenvironment{myproof2}[1]{ \paragraph{Proof of #1: } } {\hfill$\square$}

\newcommand{\ie}{\textit{i.e.}}

\newcommand{\eg}{\textit{e.g.}}
\newcommand{\reg}{\mathcal{R}}
\newcommand{\cA}{\mathcal{A}}
\newcommand*{\midcap}{\mathbin{\scalebox{1.66}{\ensuremath{\cap}}}}

\newcommand{\defeq}{\mathrel{\overset{\mathrm{def}}{=}}}

\newcommand{\E}[1]{\mathbb{E} \left[ {#1} \right]}

\newcommand{\one}[1]{\mathbbm{1} \left[ {#1} \right]}

\newcommand{\paren}[1]{\left({#1}\right)}
\newcommand{\brackets}[1]{\left[{#1}\right]}
\newcommand{\braces}[1]{\left\{{#1}\right\}}
\newcommand{\abs}[1]{\left|{#1}\right|}
\newcommand{\normtwo}[1]{\left \| {#1} \right \|_{2}}

\newcommand{\inner}[1]{\left \langle {#1} \right \rangle }
\newcommand{\p}[1]{\mathbf{Pr} \left[ {#1} \right]}

\definecolor{darkolivegreen}{rgb}{0.33, 0.42, 0.18}

\DeclareSymbolFont{extraup}{U}{zavm}{m}{n}
\DeclareMathSymbol{\varheart}{\mathalpha}{extraup}{86}
\DeclareMathSymbol{\vardiamond}{\mathalpha}{extraup}{87}

\newcommand{\rise}{\texttt{RISE}}
\newcommand{\risepp}{\texttt{RISE++}}

\newcommand{\bftheta}{\boldsymbol{\theta}}

\TheoremsNumberedThrough     
\ECRepeatTheorems

\EquationsNumberedThrough    

\allowdisplaybreaks
\begin{document}



\RUNTITLE{Risk-Aware Linear Bandit}

\TITLE{\Large Risk-Aware Linear Bandits: Theory and Applications in Smart Order Routing}


\ARTICLEAUTHORS{%
\AUTHOR{Jingwei Ji}
\AFF{
    Management Science and Engineering, 
    Stanford University, 
    \EMAIL{jingwei.ji@stanford.edu}, 
    \URL{}
}

\AUTHOR{Renyuan Xu}
\AFF{Management Science and Engineering, 
    Stanford University,  
    \EMAIL{renyuanxu@stanford.edu}
    }

\AUTHOR{Ruihao Zhu}
\AFF{SC Johnson College of Business, Cornell University, \EMAIL{ruihao.zhu@cornell.edu}}

} 

\ABSTRACT{%
Motivated by practical considerations in machine learning for financial decision-making, such as risk aversion and large action space, we consider risk-aware bandits optimization with applications in smart order routing (SOR). 
Specifically, based on {preliminary} observations of linear price impacts made from the NASDAQ ITCH dataset, we initiate the study of risk-aware linear bandits. In this setting, we aim at minimizing regret, {which measures our performance deficit compared to the optimum's}, under the mean-variance metric when facing a set of actions whose rewards are linear functions of (initially) unknown parameters. 
Driven by the variance-minimizing globally-optimal (G-optimal) design, we propose the novel instance-independent \underline{Ris}k-Aware \underline{E}xplore-then-Commit (\rise) algorithm 
and the instance-dependent \underline{Ri}sk-Aware \underline{S}uccessive \underline{E}limination (\risepp) algorithm. 
Then, we analyze their near-optimal regret upper bounds to show that, by leveraging the linear structure, our algorithms can dramatically reduce the regret when compared to existing methods. 
Finally, we demonstrate the performance of the algorithms by conducting extensive numerical experiments in the SOR setup using both synthetic datasets and the NASDAQ ITCH dataset.
Our results reveal that 1) The linear structure assumption can {indeed} be well-supported by the Nasdaq dataset; and more importantly 2) Both \rise~and \risepp~can significantly outperform the competing methods, in terms of mean-variance regret, especially in complex decision-making scenarios. 

}%


\maketitle

%

\section{Introduction}\label{sec:introduction}
The increasing amount of financial data  has revolutionized the techniques on data analytics, and brought new theoretical and computational challenges to the finance industry. In contrast to classic stochastic control theory and other analytical approaches, which typically rely heavily on model assumptions, machine learning based approaches can fully leverage the large amount of financial data. Moreover, they require much fewer model assumptions when utilized to improve financial decision-making \citep{hambly2021recent}.

Among others,  multi-armed bandit (MAB) is one of the most popular learning-based paradigms for sequential decision-making. At a high level, MAB concerns the scenario where an agent iteratively chooses one action/arm, among many, and then receives an action-specific random reward sampled from an (initially) unknown distribution. 
Rewards of other un-chosen actions remain unseen. The performance of the agent is (typically) measured by the notion of \emph{regret}, which is the difference between the maximum expected total reward and the agent's expected total reward. Here, the agent faces the classical \emph{exploration-exploitation dilemma}, where she has to accurately learn the reward distribution of each action while collecting high rewards. Due to its flexibility and simplicity, MAB has found a wide variety of applications in revenue analytics \citep{li2010contextual,agrawal2019mnl} and portfolio management \citep{shen2015portfolio}.

In finance, a fundamental but largely overlooked problem of optimal executions across multiple venues, often referred to as smart order routing (SOR), could be naturally formulated into the MAB setup (see  \citep{almgren2001optimal,cartea2016incorporating,lin2015trade} for developments in the single-venue case). 
SOR is an automated trading procedure that seeks to find the best available way to execute orders among a range of different trading venues, including  both \textit{lit pools} and \textit{dark pools}. 
A lit pool often refers to a public stock exchange where the order book is openly displayed and available for all participants. Dark pools are private exchanges only available to institutional investors. These private exchanges are known as ``dark pools'' due to their lack of transparency. When market participants have access to multiple venues and can split their orders and route the child orders to different venues for execution,  the overall execution price and quantity could be improved significantly. 
Similar to the MAB setting, one of the key challenges of this problem is the partial observability and censored feedback of the orders submitted to dark pools \citep{agarwal2010optimal}. 
More specifically, only when an order is submitted to a given dark pool, can the market participant observe the corresponding executed amount (possibly smaller than the submitted order amount) and the execution price. This makes the MAB framework particularly suitable for modeling the SOR problem, as there is a {\it natural exploration versus exploitation tradeoff}. 

Despite these, several challenges remain for the MAB to capture some {\it main characteristics} of the SOR problem and to be applied to a broader class of decision-making problems in finance:
\begin{itemize}
    \item \textbf{Challenge 1. Risk and Uncertainty:} Most of the existing works in the MAB framework focus on maximizing the {\it expected total reward} for a {\it risk-neutral} agent. However, in many financial applications, reducing the {\it risk and uncertainty} of the outcome is equally important. In the SOR example, many institutional traders and brokers face the requirement of reducing the risk  of their profits and losses (PnLs) or the worst-case losses made during a given period of time;
    \item \textbf{Challenge 2. Large Action Space:} For many financial decision-making problems, the action space can grow to be prohibitively large, which prevents the direct adoption of the canonical MAB setup. Again in the SOR problem, for instance, splitting $S$ shares among $d$ venues results in $K= \Theta(S^d)$ different actions.
    This can easily make regret guarantee of many existing MAB algorithms vacuous.
\end{itemize}
To address \textbf{Challenge 1}, a recent line of works has started to develop learning algorithms that simultaneously achieve reward maximization and risk minimization. In these works, a commonly adopted measure that strikes the right balance between these two (potentially conflicting) goals is \emph{mean-variance} \citep{markowitz1952}. Since its inception, the mean-variance metric has been widely adopted in many finance applications such as asset management and portfolio selection \citep{rubinstein2002markowitz,zhou2003markowitz}. In the mean-variance measure, the risk is quantified by the variance of the reward (\textit{i.e.}, deviation from the expected reward) and the objective is to minimize a difference between the total variance and the (weighted) expected total reward. 

Regret minimization for MAB under the mean-variance measure is first considered by \cite{sani2012risk} and then further studied in \cite{vakili2016}. Formally, let $T$ be the length of the entire time horizon of decision-making, and $X^{\pi}_t$ be the observed reward at time $t$. The cumulative mean-variance (MV) of policy $\pi$ is \begin{eqnarray}\label{eq:mean-variance}
\mathbb{E} \left[\sum_{t=1}^T\left(X_t^\pi-\frac{1}{T}\sum_{s=1}^T X_{s}^\pi \right)^2 - \rho  \sum_{t=1}^{T} X^{\pi}_{t}\right],
\end{eqnarray}
where $\rho\geq0$ is the risk tolerance level.  The regret is defined with respect to a policy which selects the best single action which attains the smallest MV. 
Compared with the canonical regret (without the variance term) considered in literature, the risk-aware regret is more amenable to financial applications, where the goal is not only to identify an optimal policy but also to ensure that the reward sequence collected remains stable throughout the learning process.
This difference also gives rise to immediate difficulties. 
Under the measure of MV, regret can no longer be written as the sum of immediate performance loss at each time instant. 
Hence, a regret decomposition which involves higher-order statistics of the random time spent on each action is called upon.

\begin{figure}[H]
	\centering
	\subfigure[AMZN NASDAQ  mean with R-squared $R^2=0.99$]{\includegraphics[width=8cm,height=5cm]{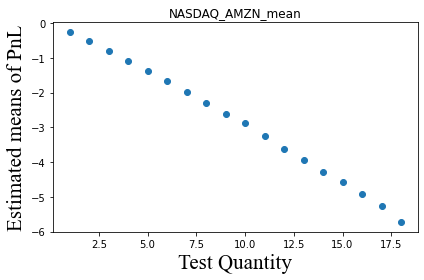} \label{fig:NASDAQ_SPY_mean} }
	\subfigure[AMZN NASDAQ  var with R-squared $R^2=0.98$]{\includegraphics[width=8cm,height=5cm]{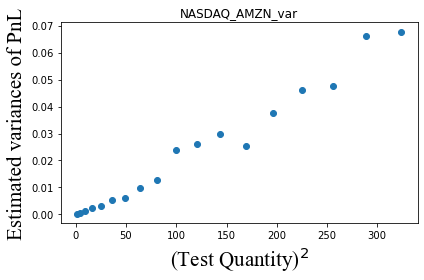} \label{fig:NASDAQ_SPY_var}	}
	\caption{\label{fig:intro}
The estimated means and variances of PnL for Amazon at NASDAQ exchange, at various test quantity to liquidate at each time point. 
This is an empirical justification for our linear form approximation. 
Further details can be found in Online Companion~\ref{sec:figures_tables}. 
 }
\label{fig:spy}
\end{figure}

However, as pointed out in {\bf Challenge 2}, the number of actions $K$ can be prohibitively large in real-world applications.
A direct application of methods from \cite{sani2012risk,vakili2016} may not yield a favorable performance guarantee, as the regret bounds can deteriorate rapidly with increasing $K$.
To address this issue, many of the existing works propose to leverage the structure of the action space (see, \eg, part V of \cite{lattimore2020bandit} for a more thorough discussion). 
For example, a stream of works \citep{dani2008stochastic,abbasi2011improved,AgrawalG13} assumes that the expected reward of each action is an inner product between its feature vector and a common (initially) unknown parameter. {This framework is known as the multiarmed bandit problem under a linear structure, or more succinctly as the linear bandit. Despite its name, this structure can integrate {\it nonlinear information} by constructing appropriate and possibly nonlinear feature/basis functions of our choice.


It turns out that in the SOR, our target application, the {\it linear structure} assumption aligns well with many empirical studies on price impacts observed in centralized limit order books \citep{bouchaud2009price}.
For instance, the practice of dividing large orders into smaller ones to minimize price impact is well-documented, with evidence indicating that the price impact of small orders is a polynomial of low degree relative to the order size (i.e., the action) \citep{bouchaud2009price,cont2014price}. More specifically, as shown in Figure \ref{fig:intro} using Amazon data,   we observe a linear relationship between the average execution price and the order size (see Figure \ref{fig:NASDAQ_SPY_mean}) as well as another linear (up to noise perturbations) relationship between the variance of the execution price and the square of the order size (see Figure \ref{fig:NASDAQ_SPY_var}). These evident relationships can be incorporated into a linear structure by introducing the order size and its square into the action vector (i.e., feature basis).

} 

{Motivated by the above discussions, this paper focuses on regret minimization for bandits with a linear structure under the mean-variance metric, which appears to be the first of its kind to the best of our knowledge. In this case, the mean and variance of each action are assumed to have a linear structure on the action vector.} Our main contributions can be summarized as follows:
\begin{itemize}
\item \textbf{Modeling:} {We propose a modeling and learning framework designed to capture the key characteristics of many financial decision-making problems, notably a risk-sensitive objective and a large action space. To the best of our knowledge, this framework has not yet been investigated in the literature.} In the context of SOR, as seen in Figure \ref{fig:intro},  we can introduce first, second, and even higher moments of the order size to capture the potentially nonlinear relationship between the reward and the action. 
\item \textbf{Efficient Learning Algorithms:} For the linear bandit setting, we propose two algorithms with near-optimal instance-independent regret upper bound and instance-dependent regret upper bound, respectively. Specifically, we first propose the \underline{Ris}k-Aware \underline{E}xplore-then-Commit (\rise) algorithm with instance-independent (\ie, worst case) regret upper bound of order $\tilde{O}(d T^{2/3} ).$ 
To overcome the above-mentioned challenges, the algorithm follows the variance-minimizing G-optimal design (see, \eg, Chapter 21 of \cite{lattimore2020bandit}) to compute a distribution over a small subset of actions through which the unknown reward parameters can be learned efficiently. Then, \rise~learns the unknown reward parameters by deploying this distribution. Afterwards, it sticks to a single action after a reasonable estimate of the unknown parameters is acquired. 

On the technical end, our analysis critically relies on a novel temporal decomposition of expected mean-variance regret (see Proposition~\ref{prop:temporal_decomposition}). 
This is in sharp contrast to prior works, which would lead to polynomial dependence on the number of actions (which is typically large in our setting). This result helps to eliminate this dependence and dramatically reduce the mean-variance regret.
Moreover, in Online Companion~\ref{appendix:non-stationary}, our analysis further underscores the effectiveness of the temporal decomposition, highlighting its potential to address dynamic environments with abrupt, unknown changes.


{To further exploit the possibly benign problem instances, we introduce the \underline{Ri}sk-Aware \underline{S}uccessive \underline{E}limination (\texttt{RISE++}) algorithm with instance-dependent bound of order $O(d^3\log(d) \log^{3} (T) )$. 
The algorithm runs in phases. In each phase, it explores the remaining actions using the re-computed G-optimal design (\textit{w.r.t.} the remaining actions). Afterwards, it estimates the reward parameters with the collected data, and eliminates actions with poor estimated mean-variance values. }
    
While the instance-independent regret upper bound guarantees the growth rate of the expected regret (\textit{w.r.t.} $T$) for all possible problem parameters, the instance-dependent regret upper bound can provide exponential improvement (\textit{w.r.t.} $T$) for the growth rate of expected regret when the problem parameters are in our favor. 
We remark that both algorithms' regret upper bounds successfully decouple the dependence on $K$ from that of $T$, which greatly reduce the regret when compared to existing approaches (see \eg, \cite{sani2012risk,vakili2016}), especially when the number of actions grows large. Hence, our algorithms are potentially more suitable for practical usage. 
When compared to the lower bounds developed in \cite{vakili2016}, both \rise~and \risepp~enjoy nearly optimal (up to poly-logarithmic factors) dependence on $T$.


\item \textbf{Numerical Experiments for Smart Order Routing:} To complement the above, we implement \rise~and \risepp~on both synthetic datasets and the NASDAQ ITCH dataset for the SOR problem. 
Along the way, we first formally delineate the linear approximation of the SOR problem, with empirical evidence showing that the approximation error is small on the NASDAQ ITCH dataset. 
Then, we show that our proposed algorithms can significantly outperform competing methods in the literature for SOR in terms of mean-variance regret. 

\item \textbf{Comparisons with the Preliminary Version:}  A preliminary version of this paper  \citep{ji2022risk} has been published in Proceedings of the Third ACM International Conference on AI in Finance, in which only instance-independent regret upper bound (\ie, \rise) is established and the proposed algorithm is only tested on synthetic datasets. 
This version extends the preliminary version in several directions with substantial developments. In particular: 
\begin{enumerate}
    \item {We eliminate an additional additive factor of $O(d^2)$ in the upper bound of \rise~ by deriving a novel temporal decomposition of the regret; } 
    \item We further develop \risepp~that is capable of providing a near-optimal instance-dependent regret upper bound. It helps to capture the potentially more benign decision-making environment in our setting; and
    \item Driven by the practical necessity to adjust decisions according to evolving market conditions, we also study the extension of a non-stationary setting where the environment changes at unknown switching times. 
    Not only has this setting attracted attention in literature \cite{garivier2008upper, yu2009piecewise, auer2008near},  but also it reflects practical industry practices, where models are typically retrained on a weekly basis or at even longer intervals; 
    \item More importantly, we also provide a more thorough numerical experiment based on the NASDAQ ITCH dataset. This demonstrates the practicality of our proposed methods on real-world scenarios.
    What is more, in Online Companion~\ref{append:auot_cor}, we conduct additional numerical experiments to evaluate the robustness of our algorithms under autocorrelated noise with varying correlation levels. The results show that our algorithms are robust to correlated noise and perform comparably to, or better than, the benchmarks across all tested scenarios. 
\end{enumerate}

\end{itemize}

\subsection{Related Literature} 
In this section, we discuss the existing results in literature and highlight the technical difficulty in our problem. 

The linear bandit problem was first studied by \cite{auer2002using} under the name of linear reinforcement learning. Improved and optimal algorithms for this setting are later proposed in \cite{dani2008stochastic} and \cite{abbasi2011improved}. In this work, we examine linear bandits through an optimal design perspective \citep{lattimore2020bandit,ZhuB22}. 
{This line of classical works does not consider risk aversion. }

Regret minimization under the mean-variance measure has been studied in several existing works {for the MAB setting}. \cite{sani2012risk} proposed an explore-then-commit algorithm with 
instance-independent regret $O(KT^{2/3})$ and a UCB-type algorithm with instance-dependent regret $O(K^2\sqrt{T})$.
Later on, \cite{vakili2016} established an $\Omega(K\log T)$ lower bound for the instance-dependent regret and an $\Omega( T^{2/3})$ lower bound for the instance-independent regret when $K=2$, alongside an improved analysis of the UCB-type algorithm. 
\cite{ZhuT20} also propose a Thompson sampling algorithm for mean-variance bandits.
We remark that the above-mentioned works on risk-aware multi-armed bandits focus exclusively on the {\it small action space} scenario. 
{ Other related works include \cite{saux2023risk}, where they consider a class of elicitable risk measures, applied to the per-round rewards. } 
{Interested readers are referred to the survey \cite{TanAJ22} for a more thorough discussion of existing works in risk-aware bandits, including the adaptation of other risk measures \citep{simchi2023stochastic, si2023distributionally}.  } 

To derive a desirable regret bound in risk-aware MAB studies, a common step is to decompose the risk-aware regret into higher-order statistics of actions and number of times each action is taken (see, e.g., {\color{blue} Lemma~1 in \cite{sani2012risk}, Theorem~1 in \cite{vakili2016} }).
It turns out that in the linear bandit setting, existing decompositions in the literature do not facilitate a desirable analysis of the instance-independent regret. 
To this end, we propose a new temporal decomposition of the expected regret, and demonstrate how it streamlines the analysis of the explore-then-commit algorithm. 

%


For the risk-neutral setting, going from linear bandit \citep{abbasi2011improved} to the {contextual bandit \citep{chu2011contextual, blanchet2023delay}} is relatively straightforward. However, in the risk-aware case, the total regret does not adhere to a mere additive structure based on the rewards of each round, hence not as trivial to go from the linear model to the contextual model.

The SOR problem across multiple lit pools was first studied in \cite{cont2017optimal} under a convex optimization framework. \cite{baldacci2020adaptive} extended the single-period model to multiple trading periods and adopted a Bayesian framework for  updating the model parameters. 
For allocation across dark pools,  \cite{laruelle2011optimal} adopted a stochastic control approach to solve the optimal order splitting strategy, \cite{ganchev2010censored} formulated the problem as an online learning problem with censored feedback and applied the Kaplan-Meier estimator to estimate the reward distribution.  \cite{agarwal2010optimal} proposed an exponentiated gradient-style algorithm and proved an optimal regret guarantee $O(K\sqrt{T})$  under the adversarial setting. 
\cite{bernasconi2022dark} applied an existing combinatorial MAB algorithm to the context of SOR and conducted numerical testing on it.
It is worth noting that there are no existing results on learning to route across both dark pools and lit pools \citep{hambly2021recent}.
{Finally, bandit algorithms have become increasingly popular in financial application domains. Examples include market making \citep{abernethy2013adaptive},  robo-advising \citep{alsabah2021robo} and portfolio selection \citep{huo2017risk}.}

\subsection{Organization}
The paper is organized as follows. 
We first focus on the theoretical analysis for the linear bandit case (Sections~\ref{sec:formulation}-\ref{sec:risepp}) and then demonstrate the numerical performance of our proposed algorithm (Section \ref{sec:numerical}).
Section~\ref{sec:formulation} formally describes the problem formulation for the linear bandit.
In Section~\ref{sec:rise} and Section~\ref{sec:risepp}, we introduce and analyze the \rise~ and \risepp~algorithms for the linear bandit, respectively. 
In Section~\ref{sec:numerical}, we explain the details of how the SOR problem can be formulated as a linear bandit problem and then showcase the superiority of our proposed algorithms for the SOR problem in a practical environment. 
Finally, to maintain a smooth reading flow, we defer the results regarding the non-stationary environment to Online Companion~\ref{appendix:non-stationary}. 

\section{Problem Formulation: {Linear Bandit}} \label{sec:formulation} 
In this section, we introduce the notations and the setup of our risk-aware {linear bandit problem}. 

\noindent\textbf{Notations:} For any positive integer $n,$ we use $[n]$ to denote the set $\{1,2,\ldots,n\}.$ For any non-negative integer $m,$ $\|\cdot\|_m$ is the $\ell_m$-norm. We denote $\one{\cdot}$ as the indicator variable. We adopt the asymptotic notations $O(\cdot),\Omega(\cdot),$ and $\Theta(\cdot)$ \citep{CLRS09}. When logarithmic factors are omitted, we use $\tilde{O}(\cdot),\tilde{\Omega}(\cdot),$ $\tilde{\Theta}(\cdot),$ respectively.
The symbol $x \lesssim y$ means that there exists some absolute constant $C$ such that $x \leq C y$. 
Given a matrix $A$, $A^{\dagger}$ denotes its pseudo-inverse. 
Given a positive semi-definite matrix $A$, $\| x \|_A $ denotes $\sqrt{x^\top A x}$.  

\noindent\textbf{Setup:} Let $\mathcal{A} \subset \mathbb{R}^d$ be a fixed and finite set of actions with size $K$. We assume each $a \in \mathcal{A}$ has a bounded $\ell_2$ norm, \textit{i.e.}, $\| a \|_2 \leq 1$. For notational convenience, we assume that $\mathcal{A} $ spans $ \mathbb{R}^d$ (we show how to relax this assumption in Section \ref{sec:risepp}).
At each time step t, we select an action $A_t \in \mathcal{A}$ and observe a reward:
\begin{equation}
    X_t = \langle A_t, \theta_{*} \rangle + \eta_t(A_t),
\end{equation} 
where the noise $\eta_t(A_t)$ satisfies the following: 
\begin{itemize}
    \item $\eta_t(A_t)$ is sampled independently across time steps $t$; 
    \item Let $\mathcal{N}(0, \sigma^2)$ denote a Gaussian distribution with zero mean and variance $\sigma^2$, and $\omega$ be a constant ensuring that the variance is non-negative. 
    The variance of the noise depends on action $A_t$: 
    \begin{equation} \label{eq:var_def}
    \eta_t(A_t) \sim \mathcal{N} \left(0,  \langle \phi_{*}, A_t \rangle + \omega \right),
    \end{equation}
    whose functional form is known by the decision maker. 

    \item Vectors $\theta_{*}, \phi_{*} \in \mathbb{R}^d$ are unknown parameters with bounded $\ell_2$-norm, \textit{i.e.}, $\| \theta_{*} \|_2 \leq 1 , \| \phi_{*} \|_2 \leq \omega  ,$ so that $ \operatorname{Var} [ \eta_t(A_t) ]   \in [\sigma_{\min}^2, \sigma_{\max}^2]$.
    We assume that the decision maker is aware of the upper bound $\sigma_{\max}$. 
\end{itemize}  

We use $\rho\geq0$ to denote the (user-specified) risk tolerance level. For ease of exposition, the expected reward of an action $a$ is denoted as $\mu_a = \inner{a, \theta_{*}}$ and the variance is denoted as
$\sigma_a^2 = \inner{ \phi_{*}, a  } + \omega$.
We also define 
\begin{align}
\Gamma_{a,b} = \mu_a-\mu_b \leq \Gamma_{\max} \ , \quad \Delta_{a,b} = \sigma_a^2 - \rho \mu_a - (\sigma_b^2 - \rho \mu_b)
\end{align} 
as the difference between the expected reward of actions $a$ and $b$, and the difference of the mean-variance (MV) of these two actions, respectively. 
Finally, we call $\Delta_a =  \Delta_{a, a_*} $, where $a_* = \underset{ a \in \mathcal{A} }{\arg\min} ~ \{ \sigma_a^2 - \rho \mu_a \} $ the mean-variance gap of action $a$. 

\noindent\textbf{Regret and the Optimal Single Arm Policy:} Let $\mathbb{F}=\left(\mathcal{F}_{t}\right)_{t=0}^{\infty}$ be the filtration with $\mathcal{F}_{t}:=\sigma\left(A_{1}, X_{1}, \ldots, A_{t}, X_{t}\right)$.
A policy $\pi$ maps $\mathcal{F}_{t-1}$, the history available at time $t$, to the action set $\mathcal{A}$. 
We denote by $X^{\pi}_t$ the reward obtained by a policy $\pi$ at time $t$. The mean-variance of policy $\pi$ over $T$ time periods is defined to be 
\begin{equation} \label{eq:mv_def}
\xi_{\pi} (T) = \sum_{t=1}^T\left(X_t^\pi-\frac{1}{T}\sum_{s=1}^T X_{s}^\pi \right)^2 - \rho \sum_{t=1}^{T} X^{\pi}_{t} \ .
\end{equation}

Let $\pi^*$ be the optimal single-arm policy, which has the complete knowledge of $\theta_*, \phi_*$ and hence always selects the action with the smallest mean-variance. 
We follow existing works on risk-aware MABs \citep{vakili2016,sani2012risk,ZhuT20} to define the expected (mean-variance) regret of a policy $\pi$ as
\begin{equation} \label{eq:def_pseudo_regret}
    \E{ \reg_{\pi}(T)} = \E {\xi_{\pi} (T) - \xi_{\pi^{*}} (T)} \ .
\end{equation}
Our goal is to devise an efficient algorithm which obtains a sublinear regret.

\section{Risk-Aware Explore-then-Commit (\rise) Algorithm {for Linear Bandit}} \label{sec:rise}
In this section, we begin with the description of the Risk-Aware Explore-then-Commit (\rise) algorithm and then establish its instance-independent regret upper bound. {The key idea behind the Explore-then-Commit algorithm is to split the entire horizon into an exploration phase and the exploitation phase. The agent learns to eliminate sub-optimal arms in the exploration phase and then fully exploits the identified ``best arm'' in the exploitation phase. The structure of the algorithm design works particularly well for applications that have concerns with too much exploration such as financial decisions \citep{hambly2021recent} and personalized healthcare recommendations \citep{yu2021reinforcement}.}

\subsection{The Algorithm}
\noindent\textbf{Additional Notations:} 
For any $\delta>0$, we denote 
$\alpha^{(1)} (\delta) = 4 \sqrt{2} \sigma_{\max}  \sqrt{ d + \log \left(\delta^{-1} \right)}$, $\alpha^{(2)}(\delta)=\sqrt{2 d\left(\sigma_{\max }^{2}\right)^{2}\left(\left(C^{-1}_{2} \log \left(C_{1}\delta^{-1}\right)\right)^{2}+1\right)}$, $\alpha^{(3)}(\delta)=\sqrt{2 d \sigma_{\max }^{2} \log \left(d\delta^{-1}\right)}$ where constants $C_1$ and $C_2$ are prescribed in Lemma~3 of \cite{stay2019}. 
Defining $\zeta(\delta) =  \alpha^{(2)}({\delta}/{3}) + 2 \alpha^{(3)}({\delta}/{3}) \alpha^{(1)}({\delta}/{3}) + \alpha^{(1)}({\delta}/{3})^2
= C_3 d \sqrt{  \log{(d)} }  \log{\delta^{-1}} $ for some absolute constant $C_3,$
we let $\tilde{C}$ be a constant which ensures that 
\begin{equation} \label{eq:C_tilde}
    \rho \sigma_{\max} \sqrt{2 \log \left( 4\delta^{-1}\right)}  + 
C_3 d\sqrt{\log(d)}\log(4\delta^{-1}) \leq \tilde{C} d \sqrt{  \log{(d)} }  \log \left( {\delta^{-1}} \right) .
\end{equation}

\noindent\textbf{Algorithm:} The algorithm splits the entire horizon of $T$ rounds into the exploration phase  and the exploitation phase. 
At the beginning of the exploration phase, the algorithm first utilizes the G-optimal design (\textit{i.e.}, globally-optimal design, see, \eg, Chapter 21 of \cite{lattimore2020bandit}) to identify a distribution $Q(\cdot)$ over $\cA$ by solving the following optimization problem
\begin{align}  
    \nonumber&\underset{Q(\cdot)}{\operatorname{minimize}} ~\max_{x \in\cA} ~ x^{\top}\left(\sum_{a\in\cA}Q(a)aa^{\top}\right)^{-1} x \\
    \label{eq:g-optimal}&\text{s.t.,}~\sum_{a\in\cA}Q(a)=1\,,\ Q(a)\geq 0~\forall a\in\cA.
\end{align}
For convenience, we define $g(Q) := \underset{ x \in \cA }{\max}  ~ x^{\top}\left(\sum_{a\in\cA}Q(a)aa^{\top}\right)^{-1} x$.
It has been shown in \cite{kiefer1960equivalence} (or Theorem 21.1 of \cite{lattimore2020bandit}) that there exists an optimal solution $Q^*$ for the above problem such that $Q^*$ is only non-zero for at most $d(d+1)/2$ entries. 
This avoids an exhaustive enumeration over all actions in $\cA$ during the exploration phase thus leading to a low regret.
In addition, compared to other techniques like barycentric spanner, G-optimal design yields a better dependence on $d$. 
Furthermore, $g(Q^*)\leq d$. 
Numerically, with a proper implementation, one can find a 2-approximate G-optimal design $Q^*$ with support size $\tilde{O}(d)$.
Then, the algorithm selects each action $a$ by
\begin{align}\label{eq:n_a}
    n_a=\left \lceil  \frac{ \tilde{C}^2 d^2 \log{(d)}\log^2 \left( \delta^{-1}\right)  g(Q^*) Q^* (a)  }{\varepsilon^2}   \right \rceil
\end{align} 
number of times, where $\delta$ is an input parameter to be specified. We denote the length of the exploration phase by 
\begin{equation} \label{eq:n}
    n=\sum_{a\in\cA}n_a.
\end{equation}
Finally, the parameters $\theta_*$ and $\phi_*$ are estimated based on the collected data. Specifically, we define 
\begin{align}
    V_n = \sum_{s=1}^{n} A_s A_s^{\top} 
\end{align}
as the design matrix at the end of the exploration phase and the least-square estimates (see, \eg, \cite{stay2019}) for $\theta_*$  and $\phi_*$ are thus
\begin{align}
    &\hat{\theta}_n = V_{n}^{-1} \sum_{s=1}^{n} X_s A_s, \label{eq:theta_estimate_MV}\\
    &\hat{\phi}_{n} = V_{n}^{-1} \sum_{s=1}^{n} \left( \left ( X_s - \inner{\hat{\theta}_{n} , \,A_s} \right )^2 - \omega \right) A_s.\label{eq:phi_estimate_MV}
\end{align}
In the exploitation phase, the algorithm simply follows the action $\hat{a}^*$ with the best mean-variance value \textit{w.r.t.} the estimated $\hat{\theta}_{n}$ and $\hat{\phi}_{n}$, \textit{i.e.},
\begin{align}\label{eq:a_star}
    \hat{a}^* = \arg \min_{a \in \mathcal{A}} ~ \inner{  \hat{\phi}_n -\rho \hat{\theta}_n,\,a} .
\end{align} The formal description of the \rise~algorithm is presented in Algorithm \ref{algo:ExpExp}. 

\begin{algorithm}[htbp]
  \caption{The Risk-Aware Explore-then-Commit (\rise) Algorithm.} \label{algo:ExpExp}
\begin{algorithmic}[]
\footnotesize 
\State (1) Given $\mathcal{A}$, find the (approximate) $G$-optimal design $Q^*$ (with size of support at most $O(d^2)$) as the solution to  
\begin{align}
    \nonumber&\underset{Q(\cdot)}{\operatorname{minimize}} ~g(Q):=\max_{a\in\cA} ~ x^{\top}\left(\sum_{a\in\cA}Q(a)aa^{\top}\right)^{-1} x \\
    \label{eq:g-optimal}&\text{s.t.,}~\sum_{a\in\cA}Q(a)=1\,,\ Q(a)\geq 0~~~\forall a\in\cA.  \nonumber 
\end{align}
\State (2) Select each action $a \in \mathcal{A}$ for $n_a$ times, where $n_a$ 
is 
$$
n_a=\left \lceil  \frac{ \tilde{C}^2 d^2 \log{(d)}\log^2 \left( \delta^{-1}\right)  g(Q^*) Q^* (a)  }{\varepsilon^2}   \right \rceil. 
$$
Let $n=\underset{a\in\cA}{\sum} n_a$.
\State (3) Compute least squares estimates according to 
\begin{align}
    &\hat{\theta}_n = V_{n}^{-1} \sum_{s=1}^{n} X_s A_s, \nonumber  \qquad
    \hat{\phi}_{n} = V_{n}^{-1} \sum_{s=1}^{n} \left( \left ( X_s - \inner{\hat{\theta}_{n} , \,A_s} \right )^2 - \omega \right) A_s. \nonumber 
\end{align}
\State (4) For the remaining $T - n$ rounds, select the action that maximizes the estimated MV
$$
 \hat{a}^* = \arg \min_{a \in \mathcal{A}} \inner{  \hat{\phi}_n -\rho \hat{\theta}_n,\,a} . 
$$ 

\end{algorithmic}
\end{algorithm}

\subsection{Regret Analysis}
We are now ready to present the regret upper bound of the \rise~algorithm. We begin by stating a high probability deviation bound between the empirically best mean-variance value and the true optimal mean-variance value.
\begin{proposition} \label{prop:bound_ip}
For any fixed $n$ and $\varepsilon,\delta > 0$, the algorithm \rise~ensures
\begin{equation}
    \p{ \left | \inner{a, \phi_{*} - \rho \theta_{*}} - \inner{a, \hat{\phi}_{n} - \rho \hat{\theta}_{n}  } \right | \leq \varepsilon ,~~\forall~a \in \mathcal{A}    } \geq 1 - \delta. 
\end{equation}
\end{proposition}
The proof of this proposition is provided in Online Companion \ref{sec:prop:bound_ip}.

To proceed, we provide an upper bound on the expected regret.
The proof of this upper bound hinges on a meticulous analysis of the instantaneous regret and is elaborated upon in Online Companion~\ref{sec:prop:temporal_decomposition}.

\begin{proposition}[Temporal Decomposition of Mean-Variance Regret]\label{prop:temporal_decomposition}
The expected regret of policy $\pi$ is bounded by 
\begin{equation}
    \E{\reg_\pi(T)} =  \E{ \sum_{t=1}^{T} \inner{\phi_* - \rho \theta_*, A_{t} - a_*} } 
    + \E{ \sum_{t=1}^{T-1}  \frac{1}{ (t+1)}  \sum_{s=1}^{t} \left( \inner{\theta_*, A_{t+1} - A_{s} } \right)^2 }  
    + O\paren{\sqrt{T}}  \ .   \label{eq:expected_regret_upper_bound_sqrtT}
\end{equation}    
\end{proposition}
By this decomposition, we see that in order to bound the expected regret, it suffices to control: 
(i) the accumulated loss in terms of MVs with respect to the optimal one, and 
(ii) the variance caused by switching back and forth from different arms. 
The last $ O\paren{\sqrt{T}}$ term in \eqref{eq:expected_regret_upper_bound_sqrtT} accounts for the variance arising from the interaction between future actions and current noises. Its order serves the purpose for the theorem which comes next.

Now we are ready to show that \rise~ enjoys the instance-independent bound. 
By Proposition~\ref{prop:bound_ip}, with probability at least $1-\delta$, the loss in terms of expected MV will be at most $2 \varepsilon$ in each round, and hence \rise~ has 
\begin{equation}
     \sum_{t=1}^{T} \inner{\phi_* - \rho \theta_*, A_{t}^{\rise} - a_*}  \leq n \cdot 2 \paren{ \sigma_{\max}^{2} - \sigma_{\min}^{2} + 2 \rho } + (T-n) 2 \varepsilon  \ .  \label{eq:thm_instance_independent_1}
\end{equation}
Next, we argue that the variance from the second term in \eqref{eq:expected_regret_upper_bound_sqrtT} over the \textit{whole} horizon contributes only of order $\tilde{O}(n)$  to the regret. 
To see this, we name the actions in $Q^*$ such that \rise~ pulls arms $a_1$ for $n_1$ times, then action $a_2$ for $n_2$ times, and so forth. 
Let $\bar{A}$ be the (random) index of the action that is selected in the exploitation phase. 
By design of the algorithm, we have 
\begin{eqnarray}
& & \E{ \sum_{t=1}^{T-1} \frac{1}{ (t+1)}  \sum_{s=1}^{t}  \left( \inner{\theta_*, A_{t+1}^{\rise} - A_{s}^{\rise} } \right)^2 }  \nonumber \\ 
&=&  \E{ \sum_{t=1}^{n-1} \frac{1}{ (t+1)}  \sum_{s=1}^{t}  \left( \inner{\theta_*, A_{t+1}^{\rise} - A_{s}^{\rise} } \right)^2 } 
    + \E{\sum_{i=1}^{|Q^*|} \sum_{j=1}^{K} \one{j=\bar{A}}  n_i \paren{\frac{1}{n+1}+\frac{1}{n+2}+\cdots +\frac{1}{T}} \Gamma_{i, j}^2 }  \nonumber \\ 
&\leq&  \Gamma_{\max}^{2} ( n - \sum_{j=n_1+1}^{ n } \frac{1}{j} )
+  \Gamma_{\max}^{2} \sum_{i=1}^{|Q^*|} \sum_{j=1}^{ K } \p{j=\bar{A}} n_i  \paren{\frac{1}{n+1}+\frac{1}{n+2}+\cdots +\frac{1}{T}}  \nonumber \\ 
&\leq& \Gamma_{\max}^{2} n +  \Gamma_{\max}^{2}  \paren{\frac{1}{n+1}+\frac{1}{n+2}+\cdots +\frac{1}{T}}  \sum_{i=1}^{|Q^*|} n_i \sum_{j=1}^{ K } \p{j=\bar{A}}  \nonumber \\ 
&\leq& \Gamma_{\max}^{2} n +  \Gamma_{\max}^{2} \log\paren{\frac{T}{n}} n \ , \label{eq:thm_instance_independent_2}
\end{eqnarray}
where in the first equality we decompose the regret to the exploration and exploitation phase, respectively. 
We defer the reasoning of the first inequality to Online Companion~\ref{sec:lemma:trivial_rise_exploration_bound}.
Therefore, in view of \eqref{eq:expected_regret_upper_bound_sqrtT}, \eqref{eq:thm_instance_independent_1}, \eqref{eq:thm_instance_independent_2} and by setting $\delta=\frac{1}{T}$, we have 
\begin{eqnarray}
    \E{\reg_{\rise}(T)} &\leq& n \cdot 2 \paren{ \sigma_{\max}^{2} - \sigma_{\min}^{2} + 2 \rho } + (T-n) 2 \varepsilon 
    +  \Gamma_{\max}^{2} n +  \Gamma_{\max}^{2} \log\paren{\frac{T}{n}} n + O\paren{\sqrt{T}} \nonumber \\ 
    &\lesssim& \log\paren{T} n + (T-n) \varepsilon + \sqrt{T}  \ . 
\end{eqnarray}
Recall that 
$n=\sum_{a\in\cA}n_a=\tilde{\Theta}\left({d^3}/{\varepsilon^2}\right),$ by setting $\varepsilon=\Theta(dT^{-1/3})$, hence we have shown the following regret bound.

\begin{theorem}
    With $\delta=T^{-1}$ and $\varepsilon =\Theta(dT^{-1/3})$, the algorithm \rise~ensures
\begin{equation}
     \E{\reg_{\rise}(T)} = \tilde{O} \left( dT^{2/3} \right).
\end{equation}
\end{theorem}

\begin{remark} \label{rmk:difficulty}
    While LinUCB \citep{chu2011contextual}  or successive elimination \citep{camilleri2021high} achieve sublinear regret for linear rewards in the risk-neutral case, they do not apply here because the mean-variance objective \eqref{eq:mv_def} is intrinsically non-additive across time.
    Namely, it cannot be directly decomposed into a simple sum of per-round regrets. In our setting, the dependence among all $X_t^\pi$ terms make the analysis fundamentally different from that of risk-neutral case.
    As a result, even though \rise~ has an elimination mechanism similar to \cite{camilleri2021high}, the standard analysis cannot be applied directly.  
\end{remark}

\begin{remark}[{\bf Technical Difference Compared to Existing Works}]

Just as in all previous studies on risk-aware MAB problems \citep{sani2012risk, vakili2016, ZhuT20} (we remark that \cite{sani2012risk} use the time-average regret), a regret decomposition in terms of both first-order and second-order moments or actions is imperative. Should we follow the regret decomposition developed in \cite{ZhuT20}, there would be an {\it additive} $O(K^2)$ term in the total regret bound. 
Similarly, the decomposition in \cite{sani2012risk} would also yield an {\it additive} term linear in $K$ in the regret. 
This is because both of these decompositions are carried out action-wise, instead of time. 
However, unlike decompositions used in the existing literature that typically focus on arms, our decomposition is conducted over time, hence termed ``temporal decomposition'' in Proposition \ref{prop:temporal_decomposition}. 
Further discussion is deferred to Online Companion~\ref{sec:further_discussion}, which details the regret bounds obtained by directly applying existing mean–variance MAB results to our context.  
In the forthcoming Online Companion~\ref{appendix:non-stationary}, we will see that this decomposition also has a unique advantage for analyzing the non-stationary case. 
\end{remark}

\begin{remark}[\textbf{Near-Optimal Regret Upper Bound and the Advantages}]
Existing regret upper bound for risk-aware bandits scales as $O(KT^{2/3})$ \citep{sani2012risk}, which could easily become linear in $T$ when $K$ becomes large (\eg, when $K=\Omega(T^{1/3})$). In contrast, our regret upper bound critically decouples the dependence on $K$ and $T.$ This implies that, even when the action space becomes large and complex, our regret bound would still be sub-linear in terms of $T.$ Moreover, compared to the regret lower bound developed in \cite{vakili2016}, we can conclude that our dependence on $T$ is indeed optimal up to a logarithmic factor.
\end{remark}

\section{Risk-Aware Successive Elimination (\texttt{RISE++}) Algorithm}
\label{sec:risepp}
{ 

In the previous section, \rise~is shown to be able to ensure a regret of $\tilde{O} \left( d T^{2/3} \right)$ for any underlying parameters of our problem. However, as already noticed in the MAB setting (see, \eg, Section 7 of \cite{lattimore2020bandit}), it is in fact possible to attain a much smaller regret (in terms of dependence on $T$) when the problem parameters are set in favor of learning (especially when $\Delta_a$'s are relatively large when compared to $1/T$). 

In this section, we exploit this possibility for risk-aware linear bandits. We first introduce the Risk-Aware Successive Elimination (\risepp) algorithm, and then establish its instance-dependent regret upper bound.

\subsection{The Algorithm}


\noindent\textbf{Algorithm:} 
The \underline{Ri}sk-Aware \underline{S}uccessive \underline{E}limination (\texttt{RISE++}) algorithm successively eliminates undesirable actions in phases. 
In each phase, the remaining actions are explored in accordance with a G-optimal design.

To be specific, in phase $\ell$ for $\ell = 1, 2, \cdots$, let $\mathcal{A}_{\ell}$ be the set of remaining promising actions to explore. 
We initialize $\mathcal{A}_{1}$ to be $\mathcal{A}$. 
For notational  convenience, for any finite set $\mathcal{H} \subseteq \mathbb{R}^m$ which spans $\mathbb{R}^m$, we define 
$
\operatorname{G\text{-}opt}(\mathcal{H}) \defeq 
\braces{ 
\underset{Q(\cdot)}{\arg\min} ~\underset{x \in \mathcal{H}}{\max}  ~ x^{\top} \left( \underset{a \in \mathcal{H}}{\sum}  Q(a) aa^{\top} \right)^{-1} x ~ \mid~  \underset{a \in \mathcal{H}}{\sum}  Q(a) = 1,~ Q(a) \geq 0 \text{~for all~} a \in \mathcal{H}
}  
$. 
As arms are eliminated in each phase, the remaining arms may not span $\mathbb{R}^d$ any more. 
Hence, to identify the exploration distribution, we need to consider two cases depending on the dimension of the span of $\mathcal{A}_{\ell}$: 
\begin{enumerate}
    \item If $\operatorname{dim}( \operatorname{range} (\mathcal{A}_{\ell}) ) = d$: We determine the optimal design by computing $Q_{\ell}^{*} (\cdot) = \operatorname{G-opt}( \mathcal{A}_{\ell})$. 
    \item If $\operatorname{dim}( \operatorname{range} (\mathcal{A}_{\ell}) ) = m_{\ell} < d$: Let $B_{\ell} \in \mathbb{R}^{ m_\ell \times d}$ be a matrix whose rows form an orthonormal basis for $\operatorname{span}(\mathcal{A_{\ell}})$.  
    We construct a lower dimensional action set $\tilde{\mathcal{A}}_{\ell} \defeq  \braces{ \tilde{a} : \tilde{a} = B_{\ell} a, a \in \mathcal{A}_{\ell}  } $. 
    Let $\tilde{Q}^* = \operatorname{G-opt}( \tilde{\mathcal{A}}_{\ell})$. 
    We can convert this into a distribution on $\mathcal{A}_\ell$ by the following procedure
    $$
   Q_{\ell}^{*} (a) =
    \begin{cases}
     \tilde{Q}^* (\tilde{a})  , & \text{if } a \in \mathcal{A}_{\ell} , \\
    0 , & \text{otherwise } . 
    \end{cases}
    $$
\end{enumerate}

Then, the algorithm selects each action $a \in \mathcal{A}_{\ell}$ by 
\begin{equation} \label{eq:phase_elimination_phase_length}
    n_{\ell,a}=\left\lceil \hat{C}^2 d^{2} \log(d) \log^2 \left(\frac{  T }{\delta}\right)
    \frac{Q_{\ell}^{*} (a) d }{\varepsilon_{\ell}^{2}} \right\rceil 
\end{equation}
number of times, where $\delta$ is an input parameter and $\varepsilon_{\ell}=2^{- \ell}$ is the tolerance level in phase $\ell$.
To ensure a smooth flow of reading, we present the definition of $\hat{C}$ (which is a universal constant) in Electronic Companion~\ref{sec:proof_prop_phase_elimination_confidence_bound_theta_phi_any_t}.
We denote by $n_{(\ell)} = \sum_{a \in \mathcal{A}_{\ell} } n_{\ell,a}$ the length of phase $\ell$.
Let $ V_{(\ell)} = \lambda I + \sum_{a \in \mathcal{A}_{\ell}} n_{\ell,a}  a a^{\top}$ be the design matrix used in phase $\ell$, where $I$ is the $d$-dimensional identity matrix. 
Let $t_{\ell}$ be the timestep of the beginning of phase $\ell$. 
Next, parameters $\theta_{*}$ and $\phi_{*}$ are estimated by $\ell_2$-regularized OLS estimators
\begin{equation} \label{eq:phase_elimination_theta_hat}
    \hat{\theta}_{(\ell)}=V_{(\ell)}^{-1} \sum_{t=t_{\ell}}^{t_{\ell}+n_{(\ell)}} A_{t} X_{t}  
\end{equation}
and 
\begin{equation} \label{eq:phase_elimination_phi_hat}
\hat{\phi}_{(\ell)} = V_{(\ell)}^{-1} 
\sum_{t= t_\ell}^{ t_{\ell} +n_{(\ell)} }  \left( \left( X_t - \inner{\hat{\theta}_{(\ell)} , A_t} \right)^2 - \omega \right) A_t . 
\end{equation}

Underperforming actions are eliminated based on these estimates with the tolerance level $\varepsilon_{\ell}$.
Thus,  the action set to explore for phase $\ell + 1$ is 
\begin{equation} \label{eq:phase_elimination_new_action_set}
\mathcal{A}_{\ell+1}=\left\{a \in \mathcal{A}_{\ell}: \max_{b \in \mathcal{A}_{\ell}}~\left\langle
\rho \hat{\theta}_{(\ell)} - \hat{\phi}_{(\ell)}
, b-a\right\rangle \leq 2 \varepsilon_{\ell}\right\} .
\end{equation}
The formal description of the \rise~algorithm is presented in Algorithm \ref{algo:phase_ elimination}.

\begin{algorithm}[htbp]
  \caption{Risk-Aware Successive Elimination Algorithm.} \label{algo:phase_ elimination}
  \begin{algorithmic}[]
  \footnotesize 
    \State Set $\ell = 1$, and $\mathcal{A}_1 = \mathcal{A}$.
\While { true  }
\State (1) Compute the G-optimal design:
        \State \quad \textbf{If} $\operatorname{dim}(\operatorname{span}(\mathcal{A}_{\ell})) = d$: 
        \State \quad \quad Compute $Q_{\ell}^{*} (\cdot) = \operatorname{G\text{-}opt}(\mathcal{A}_{\ell})$.
        
        \State \quad \textbf{Else} ($\operatorname{dim}(\operatorname{span}(\mathcal{A}_{\ell})) = m_{\ell} < d$):
        \State \quad \quad Compute an orthonormal basis $B_{\ell} \in \mathbb{R}^{m_{\ell} \times d}$ of $\operatorname{span}(\mathcal{A}_{\ell})$.
        \State \quad \quad Construct the lower-dimensional action set:
        $
        \tilde{\mathcal{A}}_{\ell} = \left\{ \tilde{a} = B_{\ell} a \mid a \in \mathcal{A}_{\ell} \right\} \ .
        $
        \State \quad \quad Compute $\tilde{Q}^*(\cdot) = \operatorname{G\text{-}opt}(\tilde{\mathcal{A}}_{\ell})$.
        \State \quad \quad Convert $\tilde{Q}^*$ into a distribution on $\mathcal{A}_{\ell}$:
        $
        Q_{\ell}^{*} (a) =
        \begin{cases}
            \tilde{Q}^* (\tilde{a})  , & \text{if } a \in \mathcal{A}_{\ell} , \\
            0 , & \text{otherwise } . 
        \end{cases}
        $
    \State (2) Select each action $a \in \mathcal{A}_{\ell}$ for 
    $
    n_{\ell, a} = \left\lceil \hat{C}^2 d^{2} \log(d) \log^2 \left(\frac{  T }{\delta}\right)
    \frac{Q_{\ell}^{*} (a) d }{\varepsilon_{\ell}^{2}} \right\rceil 
    $ times. 
    \State (3) Compute the design matrix $ V_{(\ell)} = \lambda I + \sum_{a \in \mathcal{A}_{\ell}} n_{\ell,a}  a a^{\top}$. Then compute least squares estimates according to 
\begin{eqnarray*}
\hat{\theta}_{(\ell)}&=&V_{(\ell)}^{-1} \sum_{t=t_{\ell}}^{t_{\ell}+n_{(\ell)}} A_{t} X_{t} ,  \\ 
\text{ and  } \hat{\phi}_{(\ell)} &=& V_{(\ell)}^{-1} 
\sum_{t= t_\ell}^{ t_{\ell} +n_{(\ell)} }  \left(\left( X_t - \inner{\hat{\theta}_{(\ell)} , A_t} \right)^2 - \omega \right) A_t . 
\end{eqnarray*}
    \State (4) Eliminate underperforming actions in terms of MV according to 
$$
\mathcal{A}_{\ell+1}=\left\{a \in \mathcal{A}_{\ell}: \max _{b \in \mathcal{A}_{\ell}}\left\langle
\rho \hat{\theta}_{(\ell)} - \hat{\phi}_{(\ell)}
, b-a\right\rangle \leq 2 \varepsilon_{\ell}\right\} .    
$$
    \State (5) $\ell = \ell + 1 $.
\EndWhile
\end{algorithmic}
\end{algorithm}

\begin{remark}[\textbf{Comparisons with the Design of \rise}]
Although it is evident that the design of \risepp~shares some similarities with \rise~(\eg, both of them are driven by the variance-minimizing G-optimal design), there are several critical differences: 
(1) \risepp~could have potentially more than one phase and keeps eliminating underperforming actions as long as there is more than one action remaining. Meanwhile, \rise~only permits a single phase; 
(2) Associated with the previous point, \risepp~has to re-compute the G-optimal design (\textit{w.r.t.} the remaining actions) at the beginning of each phase. 
In contrast, \rise~only needs to compute the G-optimal design once. Altogether, one can immediately conclude that \risepp~virtually never stops exploration (as long as there are more than one action left).  
As shown in the forthcoming Theorem~\ref{thm:risepp_regret}, this is critical in achieving low instance-dependent regret. The reason is that the continuously exploring nature of \risepp~prevents it from stopping prematurely (as what \rise~might), and can truly exploit the benign problem parameters.
\end{remark}

\subsection{Regret Analysis}

Our proof strategy is to conduct a careful analysis of the algorithm based on the temporal decomposition Proposition~\ref{prop:temporal_decomposition}. 
It turns out that by carefully examining the behavior of \risepp, we can provide a tighter bound on a term which we could only bound by $O(\sqrt{T})$ for a generic algorithm. 

To proceed, we first need to quantify how well \risepp~keeps refining its estimates of MVs of each action over the entire horizon.

\begin{proposition}
\label{lemma:phase_elimination_confidence_bound_theta_phi_any_t}
For any $\delta > 0$, we have 
\begin{equation}
   \p {
    \left | \inner{ \rho \theta_{*} - \phi_{*} - \left( \rho \hat{\theta}_{(\ell)} - \hat{\phi}_{(\ell)} \right) , a } \right | 
    \leq \varepsilon_{\ell},~ \forall~ a \in \mathcal{A}_{\ell}, ~ \forall~\ell \in \mathbb{N}
    } \geq 1 - \delta .
\end{equation}
\end{proposition}
The proof of this proposition is provided in Electronic Companion \ref{sec:proof_prop_phase_elimination_confidence_bound_theta_phi_any_t}.
While built upon existing concentration bounds \citep{abbasi2011improved, stay2019}, it still requires careful analysis of the G-optimal design outlined in \risepp. 

With underperforming actions removed in each phase, we can show that with high probability, a suboptimal action $a$ with gap $\Delta_a$ will be eliminated before phase $\ell_a = \min \{\ell \geq 1: 2 \varepsilon_\ell < \Delta_a  \} = \lceil \log_2(2 / \Delta_a ) \rceil$.
Denote 
$
\Delta = \underset{a \in \mathcal{A}, ~a \neq a_*  }{\min} ~ \Delta_a \ .
$
In such case, after $\bar{t}$ periods, there will be only one arm left, where 
\begin{eqnarray}
    \bar{t}  &=&  \sum_{\ell = 1}^{ \lceil \log_2(2 / \Delta ) \rceil }  
    n_{(\ell)}   \ . 
\end{eqnarray}
By exploiting this fact, we can refine the \( O\paren{\sqrt{T}} \) term in Proposition~\ref{prop:temporal_decomposition}, which accounts for the variance arising from the interaction between future actions and current noise. 
Although we can only guarantee an \( O\paren{\sqrt{T}} \) bound over this term for a \textit{generic} algorithm, the analysis simplifies on the event that the algorithm pulls only a single arm, allowing us to achieve a much sharper estimate.
We manage to prove the following theorem.

\begin{theorem}\label{thm:risepp_regret}
With $\delta=T^{-1}$, the expected regret of \risepp~is
\begin{equation}
    \E{\reg_{\risepp}(T)} = O \paren{ \max_{a\in\cA} \left(\frac{ \Delta_{a}+\Gamma_{a, *}^{2} }{\Delta^2_a}\right)  d^3 \log(d) \log^3(T)  + d^2 \log_2 \paren{\frac{1}{\Delta}} + d^2 \Delta }  \ . 
\end{equation}
\end{theorem}

To this end, we make a couple of discussions to compare the regret bounds of prior works and that of \rise 's.

\begin{remark}[\textbf{Comparisons with \cite{sani2012risk,vakili2016}}]
In view of the instance-dependent regret lower bound established in \cite{vakili2016}, we can conclude that our dependence on $T$ is nearly optimal up to a factor of $O(\log^2(T))$ (after ignoring dependence on $d$ and $K$). 
Moreover, compared to \cite{sani2012risk,vakili2016}, \risepp~removes any polynomial dependence on parameters $K$ and $T$, which makes it suitable for applications with large action sets. 
\end{remark}

\begin{remark}[\textbf{Comparisons with Instance-Independent Regret}]
We notice that the $O(d^3\log(d)\log^3(T))$ instance-dependent regret bound can potentially provide an exponential improvement in terms of the dependence on $T$ when compared with the $\tilde{O} (d T^{2/3} ) $ instance-independent regret bound. Of course, there is a slightly worse dependence on $d\log(d)\,.$ This is expected as it is similar to the canonical MAB setting \citep{lattimore2020bandit}. Following the regret bounds, the benefits of instance-dependent regret becomes more obvious when $T$ grows large. This also aligns with prior understandings in the canonical MAB settings, that instance-dependent regret upper bounds become more meaningful when $\Delta$'s are relatively large when compared to $1/T.$ Nevertheless, we emphasize that, by taking advantage of the linear structure, both of them decouple the dependence on $K$ and $T$. 
\end{remark}

}

\section{Application: Smart Order Routing}
\label{sec:numerical}
{In this section, we demonstrate the performance of {\rise } algorithm and \risepp~algorithm to the smart order routing (SOR) problem under the linear bandit setting. When additional contextual information is available, the numerical experiment can be easily modified to the contextual setting.}
In what follows, we first explain the problem in-depth and model it in the framework of linear bandits. 
Then, we test our algorithms on both a synthetic dataset and the NASDAQ ITCH dataset. 

Usually, the SOR problem arises for big institutions who need to liquidate a large amount of stocks. 
To do so, they can 
split the trade and submit orders to different venues, including both lit pools and dark pools. 
This could potentially improve the overall execution price and minimize the total market impact. 
Both the decision and the outcome are influenced by the characteristics of different venues as well as the structure of transaction fees and rebates across different venues.
Interested readers are referred to for instance, \cite{JP_SOR}, for further information.

\noindent{\bf Dark Pools:} 
Dark pools are private exchanges for trading securities that are not accessible by the investing public.  These exchanges, referred to as ``dark pools of liquidity'' due to their lack of transparency, were established to allow institutional investors to trade large amounts of securities without triggering too much impact on the market and facing adverse prices for their trades.

\noindent{\bf Lit Pools and Limit Order Books:} Unlike dark pools, where prices at which participants are willing to trade are not revealed, lit pools do display bid offers and ask offers for different assets. Primary exchanges operate in such a way that available liquidity is displayed at all times.
They form the bulk of the lit pools available to traders through the so-called  Limit Order Books (LOB). An LOB is a list of orders that a trading venue (\eg, the NASDAQ exchange) uses to record the interest of buyers and sellers in a particular financial asset. There are two types of orders the buyers (sellers) can submit: a \emph{limit buy (sell) order} with a preferred price for a given volume of the asset or a \emph{market buy (sell) order} with a given volume which will be immediately executed with the best available limit sell (buy) orders. 
Therefore, limit orders have a price guarantee but are not guaranteed to be executed, whereas market orders are executed immediately at the best available price. The lowest price of all  limit sell orders is called the \emph{best ask} price, and the highest price of all limit buy orders is called the \emph{best bid} price. The difference between the best ask and best bid is known as the \emph{spread} and the average of the best ask and best bid is known as the \emph{mid-price}. 
    
A matching engine uses the LOB to store pending orders and match them with incoming buy and sell orders. This typically follows the price-time priority rule \citep{preis2011price}, whereby orders are first ranked according to their price. Multiple orders having the same price are then ranked according to the time they were entered.
According to \cite{rosenblatt_report}, dark pools executed approximately 15.03\% of US equity volume in December 2022.

For the SOR problem, the most important characteristics of different dark pools are the chances of being  matched with a counterparty and the price (dis)advantages; whereas the relevant characteristics of lit pools include the order flows, queue sizes, and cancellation rates. These characteristics are often not available (initially) and could be learned by trial and error.

\subsection{Modeling SOR in the Framework of Linear Bandits} 
\label{sec:model_SOR_linear_bandit}

Consider a trader who faces a task of selling certain shares of a given  asset in a repeated setting. The trader has access to $(J+N)$ venues where $J$ of them are lit pools and  $N$ of them are  dark pools.

In each iteration $t$, the trader receives a task to sell $S$ shares of the given asset where $S$ is a known constant.
In the context of SOR problem, we overload the action vector $A_t$ to be 
$
A_t = (M_t^1,\cdots,M_t^J, L_t^1,\cdots,L_t^J, D_t^1,\cdots,D_t^N ,
{M_t^1}^2 , \cdots, {M_t^J}^2 , {L_t^1}^2 ,\cdots, {L_t^J}^2 , {D_t^1}^2 , \cdots, {D_t^N}^2 
)
\in\mathbb{R}^{2 \times (J+J+N) }
$ as the allocation taken by the trader at time $t$.
Therefore, each action corresponds to a unique integer combination for allocating $S$ units across $d$ venues. 
Here, $M_t^j$ is the number of market orders submitted to the $j$th lit pool which will be executed against the limit buy orders, 
$L_t^j$ is the number of limit sell orders submitted to the $j$th lit pool to join the queue of the best ask price level, 
and $D_t^n$ is the quantity submitted to the $n$th dark pool. 
Motivated by the phenomenon observed in Figure~\ref{fig:intro}, we also \textit{concatenate the squared terms} in the action vector $A_t$. 
In view of the linear bandit framework, the dimension of the actions is the number of venues, \ie, $d = 2 \times ( J + J + N )$. 
For $A_t$ to be admissible, we require that $ \sum_{j=1}^{J} M_t^j + \sum_{j=1}^{J} L_t^j  + \sum_{n=1}^{N} D_t^n = S$.

The reward, more specifically profit and loss (PnL) of the allocation, is assumed to have a linear form in $A_t$:
\begin{eqnarray}
X_t &=&  
\sum_{j=1}^J \, p^{\texttt{m},j} M_t^j 
+ \sum_{j=1}^J \, p^{\texttt{l},j} L_t^j 
+ \sum_{n=1}^N \, p^{\texttt{d},n} D_t^n    
+ \sum_{j=1}^J \, \tilde{p}^{\texttt{m},j} {M_t^j}^2 
+ \sum_{j=1}^J \, \tilde{p}^{\texttt{l},j} {L_t^j}^2 
+ \sum_{n=1}^N \, \tilde{p}^{\texttt{d},n} {D_t^n}^2 
+ \eta_t(A_t), \label{eq:sor_mean_model}
\end{eqnarray}
where $ p^{\texttt{m},\cdot } , p^{\texttt{l},\cdot}, p^{\texttt{d},\cdot}, \tilde{p}^{\texttt{m},\cdot} , \tilde{p}^{\texttt{l},\cdot}, \tilde{p}^{\texttt{d},\cdot}$ are the coefficients to learn. 
In view of Eq.~\eqref{eq:var_def}, we assume that the noise term in the reward follows $\eta_t(A_t)\sim \mathcal{N}(0,f(A_t))$ with
\begin{equation} 
f(A_t) = 
\sum_{j=1}^J \, (\sigma^{\texttt{m},j})^2 M_t^j 
+ \sum_{j=1}^J \, (\sigma^{\texttt{l},j})^2 L_t^j 
+ \sum_{n=1}^N \, (\sigma^{\texttt{d},n})^2 D_t^n 
+ \sum_{j=1}^J \, ( \tilde{\sigma}^{\texttt{m},j})^2 {M_t^j}^2 
+ \sum_{j=1}^J \, ( \tilde{\sigma}^{\texttt{l},j})^2 {L_t^j}^2 
+ \sum_{n=1}^N \, ( \tilde{\sigma}^{\texttt{d},n})^2 {D_t^n}^2
+ \omega, \label{eq:sor_var_model}
\end{equation}
where $\omega>0$ is a positive constant to guarantee that the variance is well-defined.  
This formulation captures \textit{both the linear and the quadratic effect} of the trading quantity in each venue on the PnL. 
These assumptions are supported by empirical observations from real data. See the discussion in Section \ref{sec:introduction} and Figures \ref{fig:spy}, \ref{fig:tsla} for the details.

In light of Theorem \ref{thm:risepp_regret}, we mention in passing that one can show that under very mild assumption, the gap $\Delta$ in our SOR setting does not directly depend on $K$. 
We refer interested readers to Online Companion~\ref{sec:discussion_gap_K} for a more detailed discussion.

\subsection{A Synthetic Example}
\label{subsec:synthetic}

In this section, we first briefly showcase the superiority of \rise~and \risepp~solving the SOR problem using synthetic data. 

For simplicity, we consider only non-negative integer allocations of $S$ orders into $d$ venues. 
Throughout the experiments, we set the risk-aversion parameter $\rho=2$  and vary the choices of $d$ and $S$. We set $\omega=1$ and scale the action vectors accordingly so that the norms of actions are less than one.

\begin{figure}[!ht]
	\centering
	\subfigure[$d=4, S=4, K=35$.] 
 {\includegraphics[width=8cm,height=5cm]{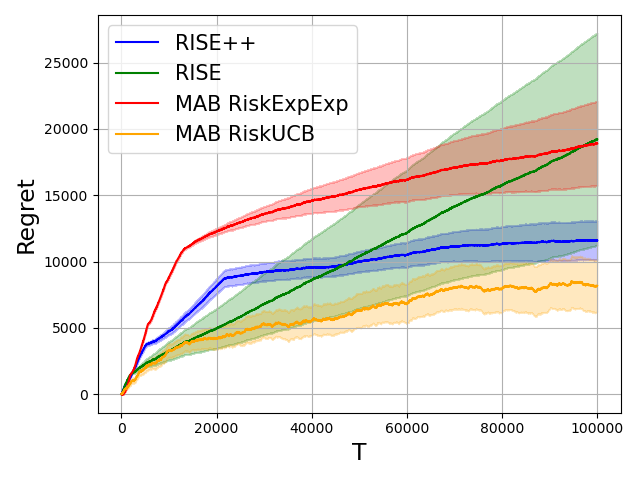} \label{fig:1_d4S4} }
	\subfigure[$d=5, S=4, K=70$.] 
 {\includegraphics[width=8cm,height=5cm]{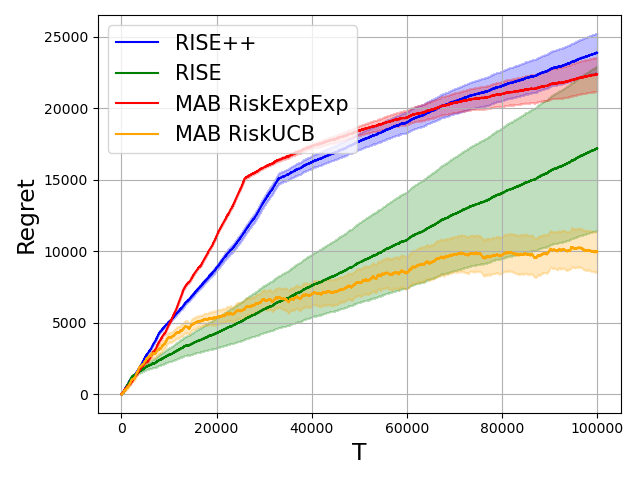} \label{fig:2_d5S4}	}
	\subfigure[$d=6, S=6, K = 462$.] 
 {\includegraphics[width=8cm,height=5cm]{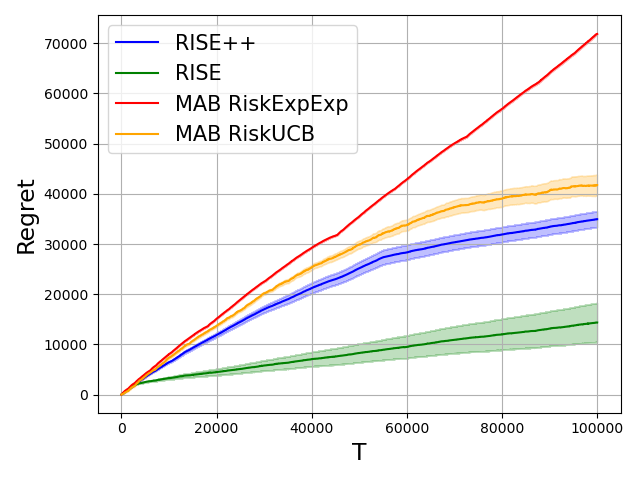} \label{fig:3_d6S4}	}
	\subfigure[$d=6, S=8, K=1287$.] 
 {\includegraphics[width=8cm,height=5cm]{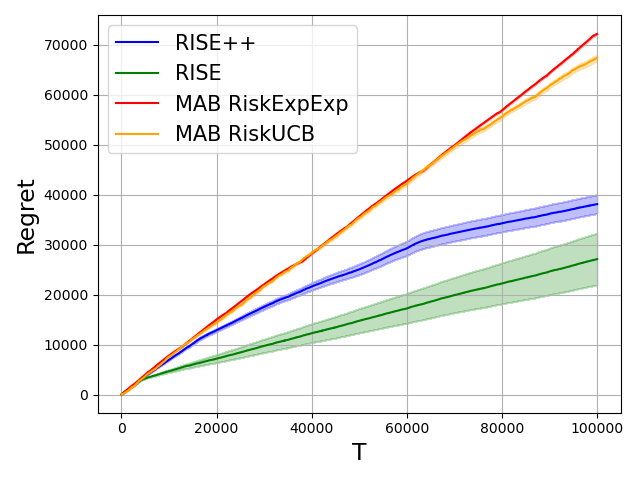} \label{fig:4_d6S8}	}
	\caption{Intermediate regrets of different algorithms over $T$. \rise~and \risepp~outperform the benchmark algorithms. }
	\label{fig:intermediate_regret}
\end{figure}

\noindent{\bf Implementation Details:}  
We note that in implementing \rise, rather than strictly adhering to \eqref{eq:n_a} and \eqref{eq:n}, we set the exploration phase length as $n = \lceil d T^{2/3} \rceil$ for practical efficiency. 


\noindent{\bf Criterion and Benchmark Algorithms:}  
To evaluate the performance of the algorithms, we generate 30 random instances and plot the average performance. 
We generate problem instances by sampling coefficients from a multivariate normal distribution.

We compare \rise~and \risepp~with two benchmark algorithms proposed by \cite{sani2012risk} for the MAB problem.
The first one is a variation of the UCB policy (referred to as \texttt{MAB RiskUCB}, and the second is a variation of the exploration-and-exploitation policy (referred to as \texttt{MAB RiskExpExp)}.
To apply these algorithms to the structured online resource allocation (SOR) problem, we treat all possible non-negative integer allocations of $S$ orders into $d$ venues as the action space. 
Moreover, we stack the quadratic terms to the action as in \eqref{eq:sor_mean_model} and \eqref{eq:sor_var_model}. 
Both benchmark algorithms are implemented following the descriptions in \cite{sani2012risk}.

\noindent{\bf Numerical Performance and Discussion:}  
The regrets in Figure~\ref{fig:intermediate_regret} illustrate the performance of the algorithms under varying problem dimensions $ d $, order sizes $ S $, and hence action space sizes $K$.
The solid curves represent the average regret values, while the shaded regions indicate 20\% of the standard deviation from the mean. 
To generate repeated trials, we first create a single instance of the bandit problem and fix its parameters for all experiments.
The algorithms are then run on this fixed instance 50 times, with randomness arising solely from the noise in each run.

\begin{itemize}
    \item \textbf{Figure 2(a):} When the action space is relatively small (\( d = 4, S = 4, K = 35 \)), the performance of \textbf{\texttt{RISE++}} (blue) and \textbf{\texttt{RISE}} (green) is comparable to the benchmarks \textbf{\texttt{MAB RiskUCB}} (orange) and \textbf{\texttt{MAB RiskExpExp}} (red). While \textbf{\texttt{MAB RiskUCB}} achieves the lowest regret in the long run, \textbf{\texttt{RISE}} and \textbf{\texttt{RISE++}} exhibit competitive performance with smaller variability.

    \item \textbf{Figure 2(b):} For a slightly larger problem (\( d = 5, S = 4, K = 70 \)), both \textbf{\texttt{RISE++}} and \textbf{\texttt{RISE}} maintain strong performance and closely match \textbf{\texttt{MAB RiskUCB}}. However, \textbf{\texttt{MAB RiskExpExp}} (red) continues to show higher regret compared to the other algorithms. The regret of \textbf{\texttt{RISE++}} remains slightly above that of \textbf{\texttt{RISE}} for most of the time horizon.

    \item \textbf{Figure 2(c):} As the action space grows significantly (\( d = 6, S = 6, K = 462 \)), a clear performance gap emerges. Both \textbf{\texttt{RISE}} and \textbf{\texttt{RISE++}} achieve lower regret compared to the benchmarks. In particular, the regret of \textbf{\texttt{MAB RiskUCB}} and \textbf{\texttt{MAB RiskExpExp}} grows at a much faster rate. The variability of \textbf{\texttt{RISE}} also decreases over time, showcasing its robustness to larger action spaces.

    \item \textbf{Figure 2(d):} When the action space is very large (\( d = 6, S = 8, K = 1287 \)), the performance differences are even more significant. 
    Specifically, since the length of its exploration phase is $K (T/14)^{2/3}$,  UCB(ExpExp) will  uniformly explore over the whole horizon when $K$ is large.  
On the contrary, \rise~and~\risepp~are able to fully utilize the linear structure of the problem and their regret growth rates do not rely on $K$, the size of action space.

\end{itemize}

\begin{remark} [Comparison between \rise~and \risepp]
    Theoretically, \risepp~ has a better asymptotic rate in $T$. Hence in theory, \risepp~ is preferred when the time horizon is long enough. 
    In Figure~\ref{fig:intermediate_regret}, we observe that \rise~ still outperforms \risepp~ in (b), (c) and (d). 
    We attribute this to two factors: 
        (1) the gaps in those cases are indeed small, since we did not deliberately select “easy” instances with large gaps; and 
        (2) the constant hidden in the $\tilde{O}$ bound may be large, so that the regret eventually exhibits a logarithmic pattern, but only at much later stages. 
        Indeed, faint traces of this transition can be observed when moving from panel (a) to (b) to (c) in Figure~\ref{fig:intermediate_regret}.

       From a practical standpoint, \rise~ is lightweight and reliable, making it a natural default choice. We therefore recommend starting with \rise~ and switching to \risepp~ only when there is empirical evidence of clear performance separation or when the time horizon is long enough for \risepp~ to realize its asymptotic gains. 
\end{remark}

\medskip

\noindent To conclude, the results demonstrate that while \textbf{\texttt{RISE}} and \textbf{\texttt{RISE++}} are not always superior in smaller action spaces (e.g., Figure~\ref{fig:intermediate_regret}(a)), they exhibit clear advantages as the action space size increases (Figures~\ref{fig:intermediate_regret}(c) and \ref{fig:intermediate_regret}(d)). In particular, both algorithms effectively leverage the problem's linear structure, leading to slower regret growth compared to the benchmarks. \textbf{\texttt{MAB RiskUCB}} and \textbf{\texttt{MAB RiskExpExp}} struggle to adapt to larger action spaces, with their regrets growing nearly linearly as $K$ increases.

\subsection{A Numerical Study Based on the NASDAQ ITCH Dataset}\label{sec:study_realdata}
Undoubtedly, the actual performance of algorithms is susceptible to the environment specification. 
In this section, we conduct a numerical study using the NASDAQ ITCH dataset to extract model parameters and evaluate the assumptions made in the framework. 
This approach ensures that the results of our study are representative of actual financial decision-making scenarios, and can be used to make more informed decisions in practice.


\noindent{\bf Data Source:} 
We extract model parameters based on historical \textit{order book messages} from three exchanges at NASDAQ on March 27, 2019.
The dataset is publicly available \citep{data}.
We only use data from 11:00 to 13:00, since the market is usually more stable during this period of time. 
We reconstruct the LOB 
from the message-level data using tools developed by \cite{https://doi.org/10.5281/zenodo.5209267}. 
We call the configuration of the LOB at a particular moment the \textit{snapshot} of the LOB. 
Each snapshot contains five queues on both the ask and bid side.
This dataset consists of 3 lit-pool venues: NASDAQ BX (BX), NASDAQ PSX (PSX) and the Nasdaq Stock Market (NASDAQ). 
Each of them is a stock exchange with different liquidity incentives. 
To be specific, NASDAQ has the conventional price/time matching mechanism, while PSX uses a price/pro-rata allocation rule and BX has an inverted pricing model. 
More details can be found at \eg, \cite{bx, psx, tradetalk}.  

\begin{figure}[htbp]
    \centering
    \includegraphics[width=12cm]{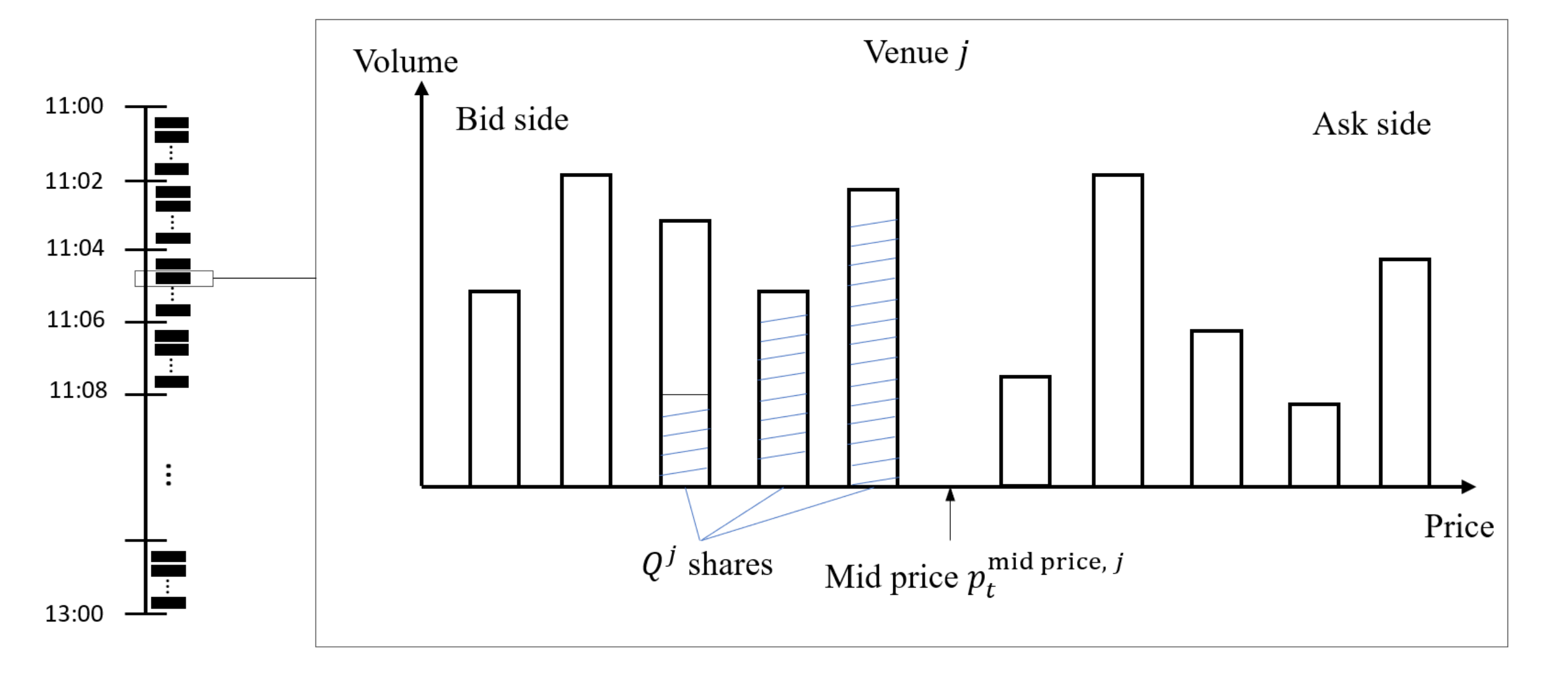}
    \caption{Visualization of the sampling procedure. We sample one snapshot of the LOB from each bucket uniformly over time. Then we obtain the empirical mean and variance PnL (cf. Eq.~\eqref{eq:estimation_p}) over 60 buckets. 
     }
    \label{fig:bucket_lob}
\end{figure}

\noindent{\bf Liquidity Simulation:}
To obtain a reasonable estimation of the parameters in Eqs.~\eqref{eq:sor_mean_model} and \eqref{eq:sor_var_model} for lit pools, we follow the sampling procedure outlined below. 
For a given stock, we group all the LOB snapshots from 11:00 to 13:00 to 60 2-min buckets.
To estimate the cost of submitting market orders to venue $j$, we sample uniformly \textit{over time} one snapshot at each venue $j$ from each bucket, with different values of \textit{test quantity} $Q^j$. 
Let $p^{\texttt{bid}, j}_{t, i}$ be the $i$th best bid price on the LOB snapshot at time $t$, and let $\bar{q}^{\texttt{bid}, j}_{t, i}$ be the corresponding queue size. 
We split $Q^j$ orders to the five queues, with as many orders to the more favorable prices as possible. 
To be precise,  let $q^{\texttt{bid}, j}_{t, i} \in [0, \bar{q}^{\texttt{bid}, j}_{t, i}]$ be the size of the market order submitted to the $i$th queue. 
We have $\sum_{i=1}^{5} q^{\texttt{bid}, j}_{t, i} = Q^j$.
Also, $ q^{\texttt{bid}, j}_{t, i} > 0 $ only when $ \sum_{i'=1}^{i - 1} q^{\texttt{bid}, j}_{t, i'} < Q^j$ for $ 2 \leq i \leq 5$.
Note that due to the nature of the problem, we do not need to set $Q^j$ too large so that the orders in the bid queues are not enough to fulfill. 
We denote $p^{\texttt{mid-price}, j }_{t}$ the mid-price of the $j$th venue.
Then the {\it relative} PnL of market orders is defined to be
\begin{equation} \label{eq:estimation_p}  
   \sum_{i=1}^{5} p^{\texttt{bid}, j}_{t, i} q^{\texttt{bid}, j}_{t, i}  -  p^{\texttt{mid-price}, j }_{t} \cdot Q^j,
\end{equation}
against the benchmark level $p^{\texttt{mid-price}, j }_{t} \cdot Q^j$. See Figure~\ref{fig:bucket_lob} for visualization. 

To sample uniformly over time 
we first uniformly sample a time point in the bucket, and then match it with the closest snapshot within the bucket. 
We note that this sampled time point is used across three venues. 
Finally, we obtain the empirical mean and variance of PnL (cf. Eq.~\eqref{eq:estimation_p}) over 60 buckets.

For dark pools, we do not have access to any data to simulate from. 
So based on the estimated coefficients of lit pools, we randomly generate moments of dark pools such that they generally have higher but more volatile returns. 
Also, we make sure that the first two moments of dark pools are on a scale comparable to the lit pools. 
For this experiment, we set to sell $S=10$ shares over $d=5$ venues. 
{ 
Table~\ref{tab:parameters_2} { in Online Companion~\ref{sec:figures_tables} } reports the parameters we used for our experiments. 
The rows $p$ and $\tilde{p}$ stand for the coefficients of the mean of the PnL for each venue, and it is estimated by regressing  PnL on $Q^j$ and ${(Q^j)}^2$. 
Similarly, the rows $\sigma^2$ and $\tilde{\sigma}^2$ stand for the estimated coefficients of the variance. 

To summarize, we try to estimate the coefficients in Eq.~\ref{eq:sor_mean_model} and Eq.~\ref{eq:sor_var_model} from real data to the greatest extent. 
For lit pools, we estimate the parameters for each venue individually. We select a time frame with high trade volume to ensure a representative sample for simulation. To estimate the mean and variance of order submission costs at different sizes, we simulate the process using limit order book snapshots, sampling them uniformly over time.
For dark pools, since no direct data sources are available, we set the parameters in a manner that aligns with real-world observations.

\noindent\textbf{Validation of Assumption \eqref{eq:sor_mean_model} and \eqref{eq:sor_var_model}:}
Figure~\ref{fig:tsla} in { Online Companion~\ref{sec:figures_tables}} plots the estimated means and variances of PnL for Tesla stock at different venues, as $Q^j$ varies.
We observe strong empirical evidence for a linear relationship between the mean trading profit and $Q^j$ for the considered scale of $Q^j$, which supports the approximation in Eq.~\eqref{eq:sor_mean_model}.
Also, there is approximately a quadratic relationship between the variance of trading profit and $Q^j$. 
Essentially, this critically hinges on the fact that we are trading a volume that will not eat up the best bid queue in a liquid market. 




\noindent{\bf Experiment Result:} Figure~\ref{fig:real_regret} shows the performance of the four algorithms in this setting. \textbf{\texttt{RISE}} (green) and \textbf{\texttt{RISE++}} (blue) both exhibit sublinear regret growth, showcasing their ability to adapt effectively to the problem structure. 
Interestingly, \textbf{\texttt{MAB RiskUCB}} (orange) performs reasonably well, exhibiting logarithmic regret growth. 
However, it suffers regret with greater magnitude than \textbf{\texttt{RISE}} and \textbf{\texttt{RISE++}}, particularly over longer horizons. 
In contrast, \textbf{\texttt{MAB RiskExpExp}} (red) suffers from a linear regret growth rate. This behavior can be attributed to its exploration phase, which scales linearly with the size of the action space, \( K \). 


\begin{figure}[htbp]
    \centering
    \includegraphics[width=9cm]{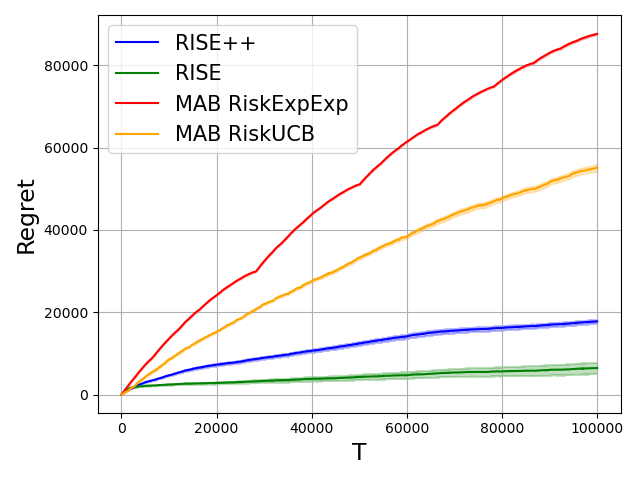}
    \caption{Intermediate regret. \rise~and \risepp~exhibit sublinear regret in the experiment based on the NASDAQ ITCH dataset. }
    \label{fig:real_regret}
\end{figure}


\section{Conclusion}

Linear bandit is a natural setting in online decision-making to handle large action sets which appear in many financial applications, meanwhile, mean-variance is one of the most popular criteria used by investors to balance the trade-off between the risk and expected returns of their investments. To the best of our knowledge, this is the first theoretical work that studies linear bandits under mean-variance framework. 
The numerical experiment justifies the linear approximation for the SOR problem, validates our theoretical finding, and demonstrates their strong performances compared with existing algorithms for the SOR problem. Motivated by other practical considerations in financial decision-making problems, future directions include risk-sensitive frameworks with other risk measures  \citep{simchi2023stochastic,coache2023reinforcement}, delayed feedback  \citep{pike2018bandits,bistritz2022no,blanchet2023delay}, and possibly an improved regret in non-stationary environment \citep{zhu2020demands,cheung2022hedging,simchi2023non}.

\SingleSpacedXI
\bibliographystyle{plainnat} 
\bibliography{ref}
\newpage


\ECSwitch


\ECHead{Online Companion}

\begin{APPENDICES} 
\DoubleSpacedXI

\section{Omitted Proofs}

\subsection{Proof of Proposition~\ref{prop:bound_ip}} \label{sec:prop:bound_ip}

From Eq. (21.1) of \cite{lattimore2020bandit}), we have for any fixed $n,$ $\delta > 0$, and every $a\in\cA,$
\begin{align} \label{eq:bound_inner_product_theta}
    \p{ \left|\langle\hat{\theta}_{n} - \theta_{*},a \rangle\right|  \geq  \sigma_{\max} \sqrt{2\|a\|^2_{V^{-1}_n}\log \left( 2\delta^{-1} \right)} } \leq \delta.
\end{align}
Then from Theorem 1 of \cite{stay2019},
we have for any fixed $n,$ $\delta > 0$, and every $a\in\cA,$
\begin{equation}
\left | \inner{ \hat{\phi}_{n} - \phi_{*}, a}  \right | \leq C_3\left \|a \right \|_{V_{n}^{-1}} d\sqrt{\log(d)}\log(2\delta^{-1})  \ . \label{eq:bound_inner_product_phi}
\end{equation}

By a union bound over \eqref{eq:bound_inner_product_theta} and \eqref{eq:bound_inner_product_phi}, we have for any fixed $n,$ $\delta>0$, and every $a\in\cA,$ it holds with probability at least $1 - \delta$, that
\begin{align}
 & \left | \inner{a, \phi_{*} - \rho \theta_{*}} - \inner{a, \hat{\phi}_{n} - \rho \hat{\theta}_{n}  } \right | \nonumber \\
\leq&  \rho \left | \inner{a, \theta_{*} - \hat{\theta}_{n} } \right | + \left | \inner{a, \phi_{*} - \hat{\phi}_{n} }\right |       \nonumber \\ 
\leq& \left \| a \right \|_{V_{n}^{-1}} \rho \sigma_{\max} \sqrt{2 \log \left(4\delta^{-1}\right)}  +  
C_3\left \|a \right \|_{V_{n}^{-1}} d\sqrt{\log(d)}\log(4\delta^{-1}) \nonumber \\
\leq&  \left \|a \right \|_{V_{n}^{-1}} \tilde{C} d \sqrt{  \log{(d)} }  \log( {\delta^{-1}}) \ , \label{eq:bound_ip}
\end{align}
where the second step follows from the triangle inequality and the last step follows from the definition of $\tilde{C}$, \textit{c.f.} Eq.~\eqref{eq:C_tilde}. 

Now recall the definition of $n_a$ in \eqref{eq:n_a}, we have
\begin{align}
V_n =& \sum_{a \in \mathcal{A}} n_a a a^\top  = \sum_{a \in \mathcal{A}} \left \lceil  \frac{ \tilde{C}^2 \log^2 \left( \delta^{-1}\right) d^2 \log{(d)} g(Q^*) Q^* (a)  }{\varepsilon^2}   \right \rceil a a^\top \nonumber  \\
\succeq& \sum_{a \in \mathcal{A}} \frac{ \tilde{C}^2 \log^2 \left( \delta^{-1}\right) d^2 \log{(d)} g(Q^*) Q^* (a)   }{\varepsilon^2}  a a^{\top} \nonumber \\ 
=& \frac{ \tilde{C}^2 \log^2 \left( \delta^{-1}\right) d^2 \log{(d)} g(Q^*)}{\varepsilon^2}\sum_{a\in\cA}Q^*(a)aa^{\top} \ ,
\end{align}
where we denote by $B \succeq 0$ if the matrix $B$ is positive semi-definite. Hence, we arrive at
\begin{align}
\left \|a \right \|_{V_{n}^{-1}}  =& \sqrt{ a^\top V_{n}^{-1} a }  \nonumber \leq \sqrt{ \frac{\varepsilon^2a^\top (\sum_{a\in\cA}Q^*(a)aa^{\top})^{-1} a}{ \tilde{C}^2 \log^2 \left( \delta^{-1}\right) d^2 \log( d)  g(Q^*)} }  \nonumber \\
\leq& \frac{\varepsilon}{ \tilde{C} d \sqrt{\log(d)}\log \left(\delta^{-1}\right)    }  \ ,  \label{eq:a_norm}
\end{align}
where the last step follows from the fact that $Q^*$ is an optimal solution to \eqref{eq:g-optimal}. The statement follows by combining \eqref{eq:bound_ip} and \eqref{eq:a_norm}.

\subsection{Proof of Proposition~\ref{prop:temporal_decomposition}} \label{sec:prop:temporal_decomposition}

For ease of notation, in this section we omit the superscript $\pi$ for the reward $X_t^\pi$ and action taken $A_t^\pi$ where no confusion arises.

The idea is to decompose the expected regret in time. 
The instantaneous expected regret at round $t, t\geq2$ can be written as  
\begin{eqnarray}
& &  \E{ r_\pi(t+1)}  \nonumber \\ 
&=& \E{ \mathcal{R}_{\pi}(t+1) - \mathcal{R}_{\pi}(t) } \nonumber \\
&=& \E{ \xi_{\pi} (t+1) - \xi_{\pi} (t) } - \left( \inner{\phi_{*}, a_*} + \omega  - \rho  \inner{\theta_* , a_*}  \right)  \nonumber \\
&=& \E{  \sum_{s=1}^{t+1} \left(X_s-\frac{1}{t+1}\sum_{\tau=1}^{t+1} X_{\tau} \right)^2
- \rho \sum_{s=1}^{t+1} X^{}_{s}
- \sum_{s=1}^{t} \left(X_s-\frac{1}{t}\sum_{\tau=1}^{t} X_{\tau} \right)^2
+ \rho \sum_{s=1}^{t} X^{}_{s}
}  - \left( \inner{\phi_{*}, a_*} + \omega  - \rho  \inner{\theta_* , a_*}  \right) \nonumber \\
&=& \E{  \frac{1}{t(t+1)} \left( t X_{t+1}^{} - \sum_{s=1}^{t} X_{s}^{} \right)^2 - \rho X_{t+1} } - \left( \inner{\phi_{*}, a_*} + \omega  - \rho  \inner{\theta_* , a_*}  \right) ,  \label{eq:per_round_regret_decomposition_1}
\end{eqnarray}
where the last equation is due to the following algebraic fact.

\begin{lemma} \label{lemma:variance_decomposition}
Denote by $\bar{x}_n = \frac{1}{n} \sum_{k=1}^{n} x_k $ and let $S_{n}=\sum_{k=1}^{n}\left(x_{k}-\bar{x}_{n}\right)^{2}$.
Then we have
\begin{equation}
    S_{n+1} =  S_{n}+\frac{\left(n x_{n+1}-n \bar{x}_n \right)^{2}}{n(n+1)} \ .
\end{equation}
\end{lemma}

Continuing from \eqref{eq:per_round_regret_decomposition_1}, we have 
\begin{eqnarray}
& & \E{ r_\pi(t+1)} \nonumber  \\
&=&    \E{  \frac{1}{t(t+1)} \left( t \inner{\theta_*, A_{t+1}}+ t \eta_{t+1}(A_{t+1})
- \sum_{s=1}^{t} \left( \inner{\theta_*, A_{s}} + \eta_s(A_s) \right) \right)^2
}
- \rho \E{\inner{\theta_*, A_{t+1}}} - \left( \inner{\phi_{*}, a_*} + \omega  - \rho  \inner{\theta_* , a_*}  \right) \nonumber  \\
&=&  \frac{1}{t(t+1)} \E{ \left( t \inner{\theta_*, A_{t+1}} - \sum_{s=1}^{t}  \inner{\theta_*, A_{s}}  \right)^2 }
+   \frac{1}{t(t+1)} \E{ \left(t \eta_{t+1}(A_{t+1}) - \sum_{s=1}^{t}  \eta_{s} (A_{s}) \right)^2 }  \nonumber  \\
& & - \rho \E{\inner{\theta_*, A_{t+1}}} - \left( \inner{\phi_{*}, a_*} + \omega  - \rho  \inner{\theta_* , a_*}  \right)  \nonumber \\
& & + \frac{2}{t(t+1)} \E{ \left( t \inner{\theta_*, A_{t+1}} - \sum_{s=1}^{t}  \inner{\theta_*, A_{s}}   \right) \left( t \eta_{t+1}(A_{t+1}) - \sum_{s=1}^{t}  \eta_{s} (A_{s})  \right)  }  \label{eq:per_round_regret_decomposition_2} \ . 
\end{eqnarray}
We proceed to deal with each term in \eqref{eq:per_round_regret_decomposition_2} respectively. 
\begin{itemize}
    \item For the first three terms, we note that 
\begin{eqnarray}
& &  \frac{1}{t(t+1)} \E{ \left( t \inner{\theta_*, A_{t+1}} - \sum_{s=1}^{t}  \inner{\theta_*, A_{s}}  \right)^2 }
+   \frac{1}{t(t+1)} \E{ \left(t \eta_{t+1}(A_{t+1}) - \sum_{s=1}^{t}  \eta_{s} (A_{s}) \right)^2 }  \nonumber \\
& & - \rho \E{\inner{\theta_*, A_{t+1}} } - \left( \inner{\phi_{*}, a_*} + \omega  - \rho  \inner{\theta_* , a_*}  \right)  \nonumber \\
&=&   \frac{1}{t(t+1)} \E{ \left( t \inner{\theta_*, A_{t+1}} - \sum_{s=1}^{t}  \inner{\theta_*, A_{s}}  \right)^2 }
+ \frac{1}{t(t+1)}  \sum_{i=1}^{t} \sum_{j=1}^{t} \left( \E{\eta_{t+1}^2} - \E{\eta_i \eta_{t+1}} - \E{\eta_j \eta_{t+1}} + \E{\eta_i \eta_j} \right)   \nonumber \\
& & - \rho \E{\inner{\theta_*, A_{t+1}} } - \left( \inner{\phi_{*}, a_*} + \omega  - \rho  \inner{\theta_* , a_*}  \right)  \nonumber \\
&=& \E{  \frac{1}{t(t+1)}  \left( t \inner{\theta_*, A_{t+1}} - \sum_{s=1}^{t}  \inner{\theta_*, A_{s}}  \right)^2
+ \frac{1}{t(t+1)}  \left( t^2 \left( \inner{\phi_*, A_{t+1}} + \omega \right)
+ \sum_{s=1}^{t}  \left( \inner{\phi_{*} , A_s} + \omega \right) \right)   }  \label{eq:per_round_regret_decomposition_3} \\
& & - \rho \E{\inner{\theta_*, A_{t+1}} } - \left( \inner{\phi_{*}, a_*} + \omega  - \rho  \inner{\theta_* , a_*}  \right)  \nonumber  \\
&\leq& \E{\frac{1}{t (t+1)} \left( \sum_{s=1}^{t} \left( \inner{\theta_*, A_{t+1} - A_{s} } \right)^2  \sum_{s=1}^{t} 1  \right)
+ \frac{1}{t(t+1)}  \left( t^2 \left( \inner{\phi_*, A_{t+1}} + \omega \right)
+ \sum_{s=1}^{t}  \left( \inner{\phi_{*} , A_s} + \omega \right) \right)   } \nonumber \\
& & - \rho  \E{\inner{\theta_*, A_{t+1}}} - \left( \inner{\phi_{*}, a_*} + \omega  - \rho  \inner{\theta_* , a_*}  \right)   \label{eq:per_round_regret_decomposition_4} \\
&=& \E{ \frac{1}{ (t+1)}  \sum_{s=1}^{t} \left( \inner{\theta_*, A_{t+1} - A_{s} } \right)^2
+ \frac{1}{t(t+1)} \left( t^2 \inner{\phi_*, A_{t+1} - a_*} + \sum_{s=1}^{t} \inner{\phi_*, A_s - a_*} \right)  - \rho \inner{\theta_* ,  A_{t+1} - a_* } }  \nonumber \\
&=& \E{ \frac{1}{ (t+1)}  \sum_{s=1}^{t} \left( \inner{\theta_*, A_{t+1} - A_{s} } \right)^2
+ \inner{\phi_* - \rho \theta_*, A_{t+1} - a_*}
   +  \frac{1}{t(t+1)}  \sum_{s=1}^{t} \inner{\phi_*, A_s - A_{t+1}} }  \label{eq:per_round_regret_decomposition_5} \ .
\end{eqnarray}
Equation~\eqref{eq:per_round_regret_decomposition_3} follows by noting that $\E{\eta_i(A_i) \eta_j(A_j)}=0$ for $i \neq j$.
Inequality~\eqref{eq:per_round_regret_decomposition_4} follows from Cauchy-Schwarz inequality. 
Let us observe that only the first two terms in \eqref{eq:per_round_regret_decomposition_5} contribute significantly to the regret. Since, 
\begin{eqnarray*}
  \E{\sum_{t=1}^{T}  \frac{1}{t(t+1)}  \sum_{s=1}^{t} \inner{\phi_*, A_s - A_{t+1}} }   &\leq&  2   \sum_{t=1}^{T}  \frac{1}{(t+1)}  = O(\log\paren{T}) \ . 
\end{eqnarray*}
    \item Regarding the last term in \eqref{eq:per_round_regret_decomposition_2}, which corresponds to the variance term caused by the accumulated noise, first we note that
\begin{equation} \label{eq:noise_term_can_ignore}
   \E{  \paren{ t \inner{\theta_*, A_{t+1}} - \sum_{s=1}^{t}  \inner{\theta_*, A_{s}}   }  \eta_{t+1}(A_{t+1}) } = \E{ \paren{ t \inner{\theta_*, A_{t+1}} - \sum_{s=1}^{t}  \inner{\theta_*, A_{s}} } \E{ \eta_{t+1}(A_{t+1}) \mid \mathcal{F}_t }}= 0 \ .  
\end{equation}
Also, we have 
\begin{eqnarray}
    & & \frac{2}{t(t+1)}   \E{ \left(  \sum_{s=1}^{t}  \inner{\theta_*, A_{s} - A_{t+1}}   \right)  \sum_{s=1}^{t}  \eta_{s} (A_{s})  } \nonumber \\ 
    &\leq& \frac{2}{t(t+1)}   \Gamma_{\max} t  \E{\left| \sum_{s=1}^{t}  \eta_{s} (A_{s}) \right| }  \label{eq:noise_term_1} \\ 
    &=& \frac{2 \Gamma_{\max}}{t+1}  \int_{0}^{\infty} \p{ \left| \sum_{s=1}^{t}  \eta_{s} (A_{s}) \right| > u} du  \nonumber \\ 
    &\leq& \frac{2 \Gamma_{\max}}{t+1}  \int_{0}^{\infty} 2 \exp \paren{ - \frac{u^2}{ 2 t \sigma_{\max}^2 } } du  \label{eq:noise_term_2} \\ 
    &=&  2  \sqrt{2 \pi \sigma_{\max}^2 \Gamma_{\max}^2 } \frac{ \sqrt{t}  }{t + 1} \label{eq:noise_term_3} \ . 
\end{eqnarray}
For \eqref{eq:noise_term_1}, we recall the fact that $\| \theta_* \|_2 \leq 1, \| a \|_2 \leq 1$, and $|xy| \leq |x| |y|$ for any $x, y \in \mathbb{R}$. 
Inequality~\eqref{eq:noise_term_2} follows since the noises are sub-Gaussian.
\end{itemize}

Putting everything together, we arrive at 
\begin{eqnarray}
    \E{\reg_\pi(T)} &\leq& \sum_{t=1}^{T-1} \E{ \frac{1}{ (t+1)}  \sum_{s=1}^{t} \left( \inner{\theta_*, A_{t+1} - A_{s} } \right)^2
+ \inner{\phi_* - \rho \theta_*, A_{t+1} - a_*} }  \nonumber \\
& & + \sum_{t=1}^{T} \paren{4 \sqrt{2 \pi \sigma_{\max}^2 \Gamma_{\max}^{2}} \frac{ \sqrt{t}  }{t + 1}  + \frac{1}{t+1}}  \nonumber \\
&=& \sum_{t=1}^{T-1} \E{ \frac{1}{ (t+1)}  \sum_{s=1}^{t} \left( \inner{\theta_*, A_{t+1} - A_{s} } \right)^2
+ \inner{\phi_* - \rho \theta_*, A_{t+1} - a_*} } + O\paren{\sqrt{T}}  \ . \label{eq:decomposition_final_eq}
\end{eqnarray}

\subsection{Statement and Proof of Lemma~\ref{lemma:trivial_rise_exploration_bound}} \label{sec:lemma:trivial_rise_exploration_bound}
\begin{lemma} \label{lemma:trivial_rise_exploration_bound}
$\E{ \sum_{t=1}^{n-1} \frac{1}{ (t+1)}  \sum_{s=1}^{t}  \left( \inner{\theta_*, A_{t+1}^{\rise} - A_{s}^{\rise} } \right)^2 } \leq \Gamma_{\max}^{2} ( n - \sum_{j=1}^{ |Q^*| } \frac{1}{j} ) \ . $
\end{lemma}
Let $N_i=\sum_{j=1}^{i} n_j$ and $N_0=0$. Let $|Q^*| = m$.
\begin{eqnarray*}
&& \E{ \sum_{t=1}^{n-1} \frac{1}{ (t+1)}  \sum_{s=1}^{t}  \left( \inner{\theta_*, A_{t+1}^{\rise} - A_{s}^{\rise} } \right)^2 } \\ 
    &=& \paren{ \frac{1}{2} + \frac{1}{3} + \cdots + \frac{1}{n_1} } \Gamma_{1,1}^2  + \paren{\frac{1}{n_1+1} + \cdots + \frac{1}{n_1+n_2}} n_1 \Gamma_{1,2}^{2}  \\ 
    & & + \paren{\frac{1}{n_1+n_2+1} + \cdots + \frac{1}{n_1+n_2+n_3}} \paren{ n_1 \Gamma_{1,3}^{2} + n_2 \Gamma_{2,3}^{2} }  + \cdots \\ 
    & & + \paren{\frac{1}{n_1+\cdots+n_{M-1}+1} + \cdots + \frac{1}{n_1+\cdots+n_{M}}} \paren{ n_1 \Gamma_{1,M}^{2} + \cdots + n_{M-1} \Gamma_{M-1,M}^{2} }    \\ 
    &\leq& \Gamma_{\max}^{2} \cdot \paren{ \sum_{i=2}^{M} N_{i-1} \sum_{j = N_{i-1}+1}^{N_i} \frac{1}{j}  }   \\
    &\leq& \Gamma_{\max}^{2} \cdot \paren{ \sum_{i=2}^{M}  \sum_{j = N_{i-1}+1}^{N_i} \frac{j-1}{j}  }  \quad \text{ as } j-1 \geq N_{i-1} \text{ for } j \in \braces{N_{i-1}+1, \cdots, N_i } \\
    &=& \Gamma_{\max}^{2} \cdot  \sum_{j= N_1+1}^{N_M} (1-\frac{1}{j}) \ . 
\end{eqnarray*}

\subsection{Proof of Proposition~\ref{lemma:phase_elimination_confidence_bound_theta_phi_any_t}}
\label{sec:proof_prop_phase_elimination_confidence_bound_theta_phi_any_t}

It is easy to extract from the proof of Theorem~2 in \cite{abbasi2011improved} that, in phase $\ell$, for given $\delta_{\ell} > 0$, we have 
\begin{equation}
    \p{ \forall~ x \in \mathbb{R}^d ~~ \inner{ \hat{\theta}_{(\ell)}-\theta_{*} , x } \leq \left \| x \right\|_{V_{(\ell)}^{-1}} 
    \paren{ \sigma_{\max}^2 \sqrt{d \log \paren{ \frac{T+\lambda}{\delta_{\ell} \lambda}}} + \sqrt{\lambda} } }  
    \geq 1 - \delta_{\ell} \ . 
\end{equation}
Choosing $\delta_{\ell} = \frac{\delta}{ 2 \ell^2 }$ and applying the union bound over $\ell \in \mathbb{N}$, we see that 
\begin{equation}
    \p{ \forall~ x \in \mathbb{R}^d, \forall~\ell \in \mathbb{N} ~~ \inner{ \hat{\theta}_{(\ell)}-\theta_{*} , x } \leq \left \| x \right\|_{V_{(\ell)}^{-1}} 
    \paren{ \sigma_{\max}^2 \sqrt{d \log \paren{ 2 \ell^2 \frac{T+\lambda}{\delta \lambda}}} + \sqrt{\lambda} } }  
    \geq 1 - \sum_{\ell=1}^{\infty} \frac{\delta}{2 \ell^2} >  1 - \delta_{} \ . 
\end{equation}

Also, in view of Theorem 1 of \cite{stay2019}, there exists a constant $C'$ that only depends on $\lambda$, $\sigma_{\max}$ such that in phase $\ell$, for given $\delta_{\ell} > 0$, we have 
\begin{equation}
    \p{ \forall~ x \in \mathbb{R}^d ~~ \inner{ \hat{\phi}_{(\ell)} - \phi_*  , x } \leq \left \| x \right\|_{V_{(\ell)}^{-1}} 
    \paren{ C' d \sqrt{\log(d)} \log(\frac{T}{\delta_{\ell}}) + \sigma_{\max}^{2} \sqrt{\lambda} } }  
    \geq 1 - \delta_{\ell} \ . 
\end{equation}

Combining the above two concentration results we know that there exists some constant $\hat{C}$ depending only on $\lambda, \sigma_{\max}, \rho$ such that for given $\delta> 0$, with probability at least $1-\delta$, 
\begin{eqnarray}
\left | \inner{ \rho \theta_{*} - \phi_{*} - \left( \rho \hat{\theta}_{(\ell)} - \hat{\phi}_{(\ell)} \right) , x } \right | 
&\leq& 
\left \|  \rho \theta_{*} - \phi_{*} - \left( \rho \hat{\theta}_{(\ell)} - \hat{\phi}_{(\ell)} \right) \right\|_{V_{(\ell)}} 
\left \| x \right\|_{V_{(\ell)}^{-1}} \nonumber \\
&\leq& \hat{C} d\sqrt{\log(d)} \log\left(\frac{ T}{\delta}\right) \left \| x \right\|_{V_{(\ell)}^{-1}}  \label{eq:risepp_bound_inner_product_MV_preliminary_1}  
\end{eqnarray}
for any $\ell \in \mathbb{N}$ and for any $x \in \mathbb{R}^d$.  

Next, we proceed to show that our design of the algorithm enables the control of $ \left \| x \right\|_{V_{(\ell)}^{-1}}$.

Recalling the definition of $n_{\ell,a}$ in \eqref{eq:phase_elimination_phase_length}, we have 
\begin{eqnarray*}
 V_{(\ell)}= \lambda I + \sum_{a \in \mathcal{A}_{\ell}} n_{\ell,a} a a^{\top} &=&  \lambda I + \sum_{a \in \mathcal{A}_{\ell}  } \left\lceil \hat{C}^2 \frac{d^{3}  \log(d) Q_{\ell}^{*}(a)}{\varepsilon_{\ell}^{2}} \log^{2} \left(\frac{  T }{\delta}\right)\right\rceil a a^{\top} \\
 &\succeq& \hat{C}^{2} \frac{ d^{3}  \log(d)}{\varepsilon_{\ell}^{2}} \log^{2} \left(\frac{  T}{\delta}\right) \sum_{a \in \mathcal{A}_{\ell} }  Q_{\ell}^{*}(a) a a^{\top}    .  
\end{eqnarray*}
Hence, by Lemma~\ref{lemma:A_geq_B}, we see that for any $x \in \operatorname{range}\paren{  \sum_{a \in \mathcal{A}_{\ell} }  Q_{\ell}^{*}(a) a a^{\top} }$, 
\begin{eqnarray}
  \left \| x \right\|_{V_{(\ell)}^{-1}} = \sqrt{ x^{\top} V_{(\ell)}^{-1} x }  
  &\leq& \frac{\varepsilon_{\ell}^{}}{\hat{C} d \sqrt{\log{d}}} \frac{1}{ \log \left(\frac{  T}{\delta}\right)  } \sqrt{\frac{x^{\top} \left(  \sum_{a \in \mathcal{A}_{\ell} }  Q_{\ell}^{*}(a) a a^{\top} \right)^{\dagger} x}{d}  } \label{eq:risepp_bound_inner_product_MV_preliminary_2} \ . 
\end{eqnarray}
To proceed, we must draw upon the following insights from algorithm design.
\begin{lemma} \label{lemma:g_optimal_obs}
    Denote $ V_\ell (Q) = \underset{ a \in \mathcal{A}_\ell }{ \sum }  Q(a) a a^\top$ and $\tilde{V}_\ell ( Q ) = \underset{a \in \tilde{\mathcal{A}}_\ell}{\sum}  Q(a) \tilde{a} \tilde{a}^\top$.
    By design of the algorithm, we have 
    (1) $\underset{a \in \mathcal{A}_\ell}{\max} ~ \| a \|^{2}_{ V_{\ell}( Q^*_{\ell} )^{\dagger}  } = m_\ell \leq d $;  and 
    (2) $\operatorname{range}\paren{  \sum_{a \in \mathcal{A}_{\ell} }  Q_{\ell}^{*}(a) a a^{\top} } = \operatorname{range} (\mathcal{A}_\ell)$. 
\end{lemma}
Given the lemma, we conclude that for any $x \in \mathcal{A}_{\ell}$, 
\begin{equation}
     \frac{x^{\top} \left(  \sum_{a \in \mathcal{A}_{\ell} }  Q_{\ell}^{*}(a) a a^{\top} \right)^{\dagger} x}{d}  \leq 1   \label{eq:risepp_bound_inner_product_MV_preliminary_3} \ . 
\end{equation}
The proof of proposition is complete, by combining \eqref{eq:risepp_bound_inner_product_MV_preliminary_1}, \eqref{eq:risepp_bound_inner_product_MV_preliminary_2} and \eqref{eq:risepp_bound_inner_product_MV_preliminary_3}.

\begin{myproof2}{Lemma~\ref{lemma:g_optimal_obs}}
    We consider two cases: 
    \begin{itemize}
    \item If $\operatorname{dim}( \operatorname{range} (\mathcal{A}_{\ell}) ) = d$: 
    Then everything follows from the General Equivalence Theorem of \cite{kiefer1960equivalence}. 
    \item If $\operatorname{dim}( \operatorname{range} (\mathcal{A}_{\ell}) ) = m_{\ell} < d$: 
    \begin{enumerate}
        \item By the property of G-optimal design, we know that $\tilde{Q}_{\ell}^{*}$ ensures $\tilde{V}( \tilde{Q}_{\ell}^{*} )$ to be invertible and  
        \begin{equation} \label{eq:max_norm_tilde}
            \max_{\tilde{a} \in \tilde{\mathcal{A}}_{\ell} } ~ \| \tilde{a} \|^{2}_{ \tilde{V}( \tilde{Q}_{\ell}^{*} )^{-1} } = m_{\ell} \ . 
        \end{equation}
        Since $B_\ell$ is orthonormal, we have 
        $$
         V_{\ell}(Q_{\ell}^{*})
         = \sum_{a \in \mathcal{A}_{\ell} }  Q_{\ell}^{*}(a) a a^{\top} 
         = \sum_{\tilde{a} \in \tilde{\mathcal{A}}_\ell } \tilde{Q}_{\ell}^{*}(a) B_{\ell}^{\top} \tilde{a} \tilde{a}^{\top} B_{\ell}
         =  B_{\ell}^{\top}  \paren{ \sum_{\tilde{a} \in \tilde{\mathcal{A}}_\ell } \tilde{Q}_{\ell}^{*}(a)\tilde{a} \tilde{a}^{\top} }  B_{\ell} 
         = B_{\ell}^{\top}   \tilde{V}_\ell ( \tilde{Q}^{*}_{\ell} ) B_{\ell}  \ . 
        $$
        Hence, $V_{\ell} ( Q_{\ell}^{*} )$ has pseudo-inverse $V_{\ell} ( Q_{\ell}^{*} )^{\dagger} = B_{\ell}^{\top} \tilde{V}_\ell ( \tilde{Q}^{*}_{\ell} )^{-1} B_{\ell}$. 
        Moreover, for every $a \in \mathcal{A}_{\ell}$, it is easy to see that 
        $
         \| a \|^{2}_{ V( Q_{\ell}^{*} )^{-1} } 
        = \| \tilde{a} \|^{2}_{ \tilde{V}( \tilde{Q}_{\ell}^{*} )^{-1} }  
        $ where $\tilde{a} = B_\ell a$. 
        Combing this with \eqref{eq:max_norm_tilde} yields the desired result.  
        \item 
        
        Now we observe that $\operatorname{range} \paren{ B_{\ell}^{\top} \tilde{V}_\ell ( \tilde{Q}^{*}_{\ell} )   B_{\ell} } 
        = \operatorname{range} (B_{\ell}^{\top} ) = \operatorname{range} (\mathcal{A}_{\ell})$.  
        To see this, first we notice that $\operatorname{range} \paren{ B_{\ell}^{\top}  \tilde{V}_\ell ( \tilde{Q}^{*}_{\ell} )   B_{\ell} }
        \subseteq \operatorname{range} (B_{\ell}^{\top} ) $. For the other direction, it suffices to realize that for any $v = B_{\ell}^{\top} z$ for some $v \in \mathbb{R}^{m_\ell}$, there exists $ x = B_\ell^\top  \tilde{V}_\ell ( \tilde{Q}^{*}_{\ell} ) ^{-1} z $ such that $B_{\ell}^{\top}  \tilde{V}_\ell ( \tilde{Q}^{*}_{\ell} )   B_{\ell} x = v $. 
    \end{enumerate}
    
\end{itemize}
\end{myproof2}

\begin{lemma} \label{lemma:A_geq_B}
    Given a symmetric positive definite matrix $A$, symmetric positive semidefinite matrix $B$, if $A \succeq B$, then $B^{\dagger} \succeq A^{-1}$ on $\operatorname{range}(B)$.
\end{lemma}
\begin{myproof2}{Lemma~\ref{lemma:A_geq_B}}
    For any $x \in \operatorname{range}(B)$, there exists some $v$ such that $x = B v$.
    Our goal is to show that 
    $$
        v^{\top} B v = v^{\top} B^{\top} B^{\dagger} B v 
        = x^{\top} B^{\dagger} x \succeq x^{\top} A^{-1} x 
        = v^{\top} B^{\top} A^{-1} B v \ . 
    $$
    It suffices to show that $B - B^{\top} A^{-1} B \succeq 0$.
    Recall the fact that given symmetric matrices $X$ and $Y$, if $X \preceq Y$ then $ C X C^\top \preceq C Y C^\top$ for any symmetric $C$.
    Hence, we know that $A- B \succeq 0$ implies that $I - A^{-\frac{1}{2}} B^{\frac{1}{2}} B^{\frac{1}{2}}   A^{-\frac{1}{2}} \succeq 0$.
    Lemma~\ref{lemma:AB_BA}, a standard fact in linear algebra yields that 
    $
        I - B^{\frac{1}{2}} A^{- \frac{1}{2}} A^{- \frac{1}{2} } B^{\frac{1}{2}} \succeq 0 \ . 
    $
    Finally, multiplying both sides by $B^{\frac{1}{2}}$ on two ends gives $B - B A^{-1} B \succeq 0$, as desired. 
    \begin{lemma} \label{lemma:AB_BA}
        If $A$ and $B$ are two square matrices, then characteristic polynomials of $AB$ and $BA$ coincide. 
    \end{lemma}
\end{myproof2}

\subsection{Proof of Theorem~\ref{thm:risepp_regret}}


We denote by $G_{\ell}^{a}$ the event that the MV estimate for action $a$ in phase $\ell$ deviates from the true value no more than $\varepsilon_{\ell}$, namely, 
$$
G_{\ell}^{a} = \left \{ \left| \inner{\rho \hat{\theta}_{(\ell)} - \hat{\phi}_{(\ell)} - \left( \rho \theta_{*} - \phi_{*} \right), a } \right|  \leq \varepsilon_{\ell} \right \} .
$$
We say the \textit{good event} $G$ happens if for any action $a$ in exploration basis $\mathcal{A}_{\ell}$ in all phases, $G_{\ell}^{a}$ happens. 
\textit{I.e.}, we let 
$G = \underset{\ell}{\midcap} \underset{a \in \mathcal{A}_{\ell}}{\midcap} G_{a}^{\ell}$. 
By Lemma~\ref{lemma:phase_elimination_confidence_bound_theta_phi_any_t}, we know the event $G$ happens with probability at least $1 - \delta$. 

We first observe that if $G$ happens, then in phase $\ell$, the optimal action $a_{*}$ will not be eliminated at step (4) in Algorithm~\ref{algo:phase_ elimination}. 
To see this, suppose in phase $\ell$, the optimal action $a_{*}$ were eliminated by some suboptimal action $b$, which would mean $\exists~b \in \mathcal{A}_{\ell}$ such that 
\begin{eqnarray}
&& 
\inner{\rho \hat{\theta}_{(\ell)} - \hat{\phi}_{(\ell)}, b-a_{*}} > 2 \varepsilon_{\ell} \nonumber  \\
 &\Leftrightarrow~& 
\langle \rho \hat{\theta}_{(\ell)} - \hat{\phi}_{(\ell)}, b  \rangle - \varepsilon_{\ell} > 
\langle \rho \hat{\theta}_{(\ell)} - \hat{\phi}_{(\ell)}, a_*  \rangle + \varepsilon_{\ell}. \label{eq:good_estimate_1}
\end{eqnarray}
Since $G$ happens, we have 
$$
\langle \rho \theta_{*} - \phi_{*}, b  \rangle \geq \langle \rho \hat{\theta}_{(\ell)} - \hat{\phi}_{(\ell)}, b  \rangle - \varepsilon_{\ell}  ~~\text{and}~~ \langle \rho \hat{\theta}_{(\ell)} - \hat{\phi}_{(\ell)}, a_*  \rangle + \varepsilon_{\ell} \geq \langle \rho \theta_{*} - \phi_{*}, a_*  \rangle
,$$
which, combined with \eqref{eq:good_estimate_1}, leads to the contradiction $\langle \rho \theta_{*} - \phi_{*}, b  \rangle >\langle \rho \theta_{*} - \phi_{*}, a_*  \rangle$.

Next, we argue that provided the good event $G$ happens, a suboptimal action $a$ with gap $\Delta_a$ will be eliminated before $\ell_a = \min \{\ell: 2 \varepsilon_\ell < \Delta_a \}$. 
Indeed, at step (3) of phase $\ell+1$, the estimated MV gap of action $a$ based on estimators in the previous phase $\ell$ is no more than  $2 \varepsilon_\ell$, \textit{i.e.}, 
$$
\max _{b \in \mathcal{A}_{\ell+1 }} \inner{ \rho \hat{\theta}_{(\ell)} - \hat{\phi}_{(\ell)}, b-a } \leq 2 \varepsilon_{\ell} = 4 \varepsilon_{l+1}.
$$
In particular, the optimal action $a_*$ remains in $\mathcal{A}_{\ell+1}$. Hence, we have 
\begin{equation} \label{eq:1}
    \inner{ \rho \hat{\theta}_{(\ell)} - \hat{\phi}_{(\ell)} , a_* - a } \leq 2 \varepsilon_{\ell}
\end{equation}
for any suboptimal action $a \in \mathcal{A}_{\ell+1}$.
On the other hand, the event $G$ implies that 
$-\varepsilon_{\ell}   \leq \inner{ \rho \hat{\theta}_{(\ell)} - \hat{\phi}_{(\ell)} - \left( \rho \theta_{*} - \phi_{*} \right) , a  } \leq \varepsilon_{\ell}$ 
and $-\varepsilon_{\ell} \leq \inner{  \rho \hat{\theta}_{(\ell)} - \hat{\phi}_{(\ell)} - \left( \rho \theta_{*} - \phi_{*} \right), a_*  } \leq \varepsilon_{\ell}$ hold, which means    
\begin{equation} \label{eq:2}
  -2\varepsilon_{\ell} \leq 
\inner{ \rho \theta_{*} - \phi_{*}  , a_* - a  } 
- \inner{ \rho \hat{\theta}_{(\ell)} - \hat{\phi}_{(\ell)} , a_* - a  }
\leq 2\varepsilon_{\ell}. 
\end{equation}
Combining \eqref{eq:1} and \eqref{eq:2} yields that 
$$
\Delta_a = \inner{ \rho \theta_{*} - \phi_{*}  , a_* - a } \leq 4 \varepsilon_{\ell} = 2 \varepsilon_{\ell+1}.
$$
Therefore, the true MV gap incurred by a suboptimal action in $\mathcal{A}_{\ell}$ is at most $2 \varepsilon_\ell$.
In other words, when $G$ happens, a suboptimal action $a$ with gap $\Delta_a$ will be eliminated before phase $\ell_a = \min \{\ell \geq 1: 2 \varepsilon_\ell < \Delta_a  \} = \lceil \log_2(2 / \Delta_a ) \rceil$. This also indicates 
\begin{align}\label{eq:eps_delta}
\varepsilon_{\ell_a}\geq\frac{\Delta_a}{4}\,.
\end{align}

Denote 
$
\Delta = \min_{a \in \mathcal{A}, ~a \neq a_* }~\Delta_a \ .
$
When event $G$ happens, all suboptimal arms except $a_*$ are eliminated before phase $\lceil \log_2(2 / \Delta ) \rceil$. 
After that, only the action $a_*$ remains. 
Denote by $\bar{t}$ the time after which there is only one arm left. Namely, 
\begin{eqnarray}
    \bar{t}  &=&  \sum_{\ell = 1}^{ \lceil \log_2(2 / \Delta ) \rceil }  
    n_{(\ell)}   \ . 
\end{eqnarray}
We remark that this quantity may exceed $T$.

Following the reasoning in Proposition~\ref{prop:temporal_decomposition}, we combine \eqref{eq:per_round_regret_decomposition_2} and \eqref{eq:per_round_regret_decomposition_5}, leading to the conclusion that the expected total MV regret can be bounded above by 
\begin{eqnarray}
     \E{\reg_\pi(T)} &\leq& \sum_{t=1}^{T-1} \Big (  \E{ \frac{1}{ (t+1)}  \sum_{s=1}^{t} \left( \inner{\theta_*, A_{t+1} - A_{s} } \right)^2 }
+ \E{  \inner{\phi_* - \rho \theta_*, A_{t+1} - a_*} }   \nonumber  \\ 
    & & + \E{ \frac{1}{t(t+1)}  \sum_{s=1}^{t} \inner{\phi_*, A_s - A_{t+1}} }   
+ \frac{2}{t(t+1)}   \E{ \left(  \sum_{s=1}^{t}  \inner{\theta_*, A_{s} - A_{t+1}}   \right)  \sum_{s=1}^{t}  \eta_{s} (A_{s})  }  \Big  )   \ . \nonumber 
\end{eqnarray}
We study how much each of these four terms contribute to the expected MV regret in the sequel. 
Recall that $\inner{\theta_*, a-b} \leq \Gamma_{\max}, \inner{\theta_*-\rho \phi_*, a-b} \leq \Delta_{\max}$ for any $a,b \in \mathcal{A}$, which we will frequently use in what follows. 
\begin{enumerate}  
    \item Regarding the first term, we observe that 
     \begin{eqnarray}
        & & \E{ \sum_{t=1}^{T-1}  \frac{1}{ (t+1)}  \sum_{s=1}^{t} \left( \inner{\theta_*, A_{t+1} - A_{s} } \right)^2 }  \nonumber \\
        &=& \E{ \sum_{t=1}^{T-1}  \frac{1}{ (t+1)}  \sum_{s=1}^{t} \left( \inner{\theta_*, A_{t+1} - A_{s} } \right)^2  ~\Big | ~ G }  \p{G} 
        + \E{ \sum_{t=1}^{T-1}  \frac{1}{ (t+1)}  \sum_{s=1}^{t} \left( \inner{\theta_*, A_{t+1} - A_{s} } \right)^2  ~\Big | ~ G^\complement }  \p{G^\complement } \nonumber \\
         &\stackrel{(a)}{\leq}&  \E{ \sum_{t=1}^{T-1}  \frac{1}{ (t+1)}  \sum_{s=1}^{t} \left( \inner{\theta_*, A_{t+1} - A_{s} } \right)^2  ~\Big | ~ G }  
         + T \cdot \delta \Gamma_{\max}^2  \nonumber \\ 
         &\stackrel{(b)}{=}&  \E{ \sum_{t=1}^{\bar{t}}  \frac{1}{ (t+1)}  \sum_{s=1}^{t} \left( \inner{\theta_*, A_{t+1} - A_{s} } \right)^2  ~\Big | ~ G }  
         + \E{ \sum_{t= \bar{t}+1 }^{T-1}  \frac{1}{ (t+1)}  \sum_{s=1}^{t} \left( \inner{\theta_*, A_{t+1} - A_{s} } \right)^2  ~\Big | ~ G }  
         + T \cdot \delta \Gamma_{\max}^2  \nonumber \\ 
         &\stackrel{(c)}{\leq}& \bar{t} \cdot \Gamma_{\max}^{2}  + \E{ \sum_{t= \bar{t}+1 }^{T-1}  \frac{1}{ (t+1)}  \sum_{s=1}^{t} \left( \inner{\theta_*, A_{t+1} - A_{s} } \right)^2  ~\Big | ~ G }  
         + T \cdot \delta \Gamma_{\max}^2  \nonumber \\ 
         &\stackrel{(d)}{=}& \bar{t} \cdot \Gamma_{\max}^{2}  + \E{ \sum_{t= \bar{t}+1 }^{T-1}  \frac{1}{ (t+1)}  \sum_{s=1}^{ \bar{t} } \left( \inner{\theta_*, A_{t+1} - A_{s} } \right)^2  ~\Big | ~ G }  
         + T \cdot \delta \Gamma_{\max}^2  \nonumber  \\
         &\stackrel{(e)}{\leq}& \bar{t} \cdot \Gamma_{\max}^{2}  + \bar{t}  \sum_{t= \bar{t}+1 }^{T-1}  \frac{1}{ (t+1)}    \Gamma_{\max}^2    
         + T \cdot \delta \Gamma_{\max}^2  \nonumber  \\
         &\lesssim& \bar{t} \cdot \Gamma_{\max}^{2}  + \bar{t}  \paren{ \log (T) - \log (\bar{t}) }  \cdot \Gamma_{\max}^2    
         + T \cdot \delta \Gamma_{\max}^2  \nonumber \ , ~ \text{if } \bar{t} \leq T \ .  
    \end{eqnarray}
 Inequalities (a), (c) and (e) follow simply from the definition of $\Gamma_{\max}$. 
 In (b), we decompose the sum to two parts. 
 Equation (d) holds since there is only one arm left provided that the event $G$ happens. 
If $\bar{t} > T$, then the bound becomes vacuous. 

    \item For the second term, we have  
    \begin{eqnarray}
        & & \E{ \sum_{t=1}^{T}  \inner{\phi_* - \rho \theta_*, A_{t} - a_*}  }   \nonumber  \\ 
        &\stackrel{(a)}{\leq}&  \sum_{t=1}^{T}  
       \E{ \inner{\phi_* - \rho \theta_*, A_{t} - a_*}   \mid G } 
        +  \p{ G^{\complement} } \Delta_{\max} 
        \nonumber  \\ 
        &\stackrel{(b)}{\leq}&  \sum_{\ell=1}^{ \lceil \log_2(2 / \Delta ) \rceil } n_{(\ell)} 2 \varepsilon_{\ell}  + T  \cdot   \p{ G^{\complement} } \Delta_{\max}     \nonumber   \\ 
        &\stackrel{(c)}{=}& \sum_{\ell=1}^{ \lceil \log_2(2 / \Delta ) \rceil } 
        \sum_{a \in \mathcal{A}_{\ell}} 
        \left\lceil \hat{C}^2 \frac{d^{3}   \log(d) Q_{\ell}^{*} (a) }{\varepsilon_{\ell}^{2}} \log^2 \left(\frac{ T }{\delta}\right)\right\rceil 
        \varepsilon_{\ell}  + T  \cdot   \p{ G^{\complement} } \Delta_{\max}  \nonumber  \\ 
        &\stackrel{(d)}{\lesssim}& 
        \sum_{\ell=1}^{ \lceil \log_2(2 / \Delta ) \rceil } 
        \paren{ 
        \hat{C}^2 \frac{d^{3}   \log(d)  }{\varepsilon_{\ell}^{2}} \log^2 \left(\frac{  T }{\delta}\right)
        + d^2 }
        \varepsilon_{\ell}  + T  \cdot   \delta \Delta_{\max}   \nonumber \\ 
        &=&  \sum_{\ell=1}^{ \lceil \log_2(2 / \Delta ) \rceil } 
        \hat{C}^2 \frac{d^{3}   \log(d)  }{\varepsilon_{\ell}^{}} \log^2 \left(\frac{  T }{\delta}\right)
        + d^2 \sum_{\ell=1}^{ \lceil \log_2(2 / \Delta ) \rceil } \varepsilon_{\ell}
          + T  \cdot   \delta \Delta_{\max} \nonumber \\ 
        &\lesssim&  \frac{1}{\Delta} \hat{C}^2 d^3 \log(d) \log^2 \left(\frac{  T }{\delta}\right)
         + d^2 \Delta 
         + T  \cdot   \delta \Delta_{\max}  \ . 
    \end{eqnarray}
In (a), we condition on the event $G$. 
In (b), we use the fact that the MV gap incurred by a suboptimal arm in $\mathcal{A}_{\ell}$ is at most $2 \varepsilon_{\ell}$, provided that the event $G$ happens. 
In addition, we note that $\lceil \log_2 \paren{ \frac{2}{\Delta} } \rceil$ may exceed $\ell_{\max}$.
In (c), we invoke the definition of length of $\ell$th phase \eqref{eq:n_a}. 
Inequality (d) follows since there are at most $d(d+1)/2$ non-zero entries in the optimal design.
The final step simply follows from sum of geometric series since we set $\varepsilon_{\ell}=\frac{1}{2^{\ell}}$.

\item The third term can be easily controlled 
$$
\E{\sum_{t=1}^{T}  \frac{1}{t(t+1)}  \sum_{s=1}^{t} \inner{\phi_*, A_s - A_{t+1}} }   \leq  2   \sum_{t=1}^{T}  \frac{1}{t+1}  \lesssim \log \paren{T}  \ . 
$$
\item Regarding the forth term, we have 
\begin{eqnarray}
    & & \E{  \sum_{t=1}^{T-1} \frac{2}{t(t+1)}   \left(  \sum_{s=1}^{t}  \inner{\theta_*, A_{s} - A_{t+1}}   \right)  \sum_{s=1}^{t}  \eta_{s} (A_{s})  }   \nonumber \\ 
    &\stackrel{(a)}{\lesssim}&  \sum_{t=1}^{T-1}  \frac{1}{t^2} \E{ \abs{ \sum_{s=1}^{t}  \inner{\theta_*, A_{s} - A_{t+1}}  } \abs{  \sum_{s=1}^{t}  \eta_{s} (A_{s}) } } \nonumber  \\ 
    &\stackrel{(b)}{\leq}& \sum_{t=1}^{T-1}  \frac{1}{t^2} \sqrt{ \E{ \paren{ \sum_{s=1}^{t}  \inner{\theta_*, A_{s} - A_{t+1}}  }^2 } } \sqrt{ \E{ \paren{  \sum_{s=1}^{t}  \eta_{s} (A_{s}) }^2  } } \nonumber \\ 
    &\stackrel{(c)}{\lesssim}&  \sigma_{\max} \sum_{t=1}^{T-1}   \frac{1}{t \sqrt{t} } \sqrt{ \E{ \paren{ \sum_{s=1}^{t}  \inner{\theta_*, A_{s} - A_{t+1}}  }^2 } }  \label{eq:fourth_first_sim} \ . 
\end{eqnarray}
In (a), we push the absolute signs inside. 
Inequality (b) follows from Holder's inequality. 
Step (c) follows since $ \E{ \paren{ \sum_{s=1}^{t} \eta_s }^2  } = \E{ \sum_{i=1}^{t}\sum_{j=1}^{t} \eta_i \eta_j   } = \sum_{s=1}^{t} \E{\eta_s^2} \leq t \sigma_{\max}^{2} $. 

To bound \eqref{eq:fourth_first_sim}, we first note that the sum up to $\bar{t}$ is 
\begin{equation}
    \sigma_{\max} \sum_{t=1}^{ \bar{t} }   \frac{1}{t \sqrt{t} } \sqrt{ \E{ \paren{ \sum_{s=1}^{t}  \inner{\theta_*, A_{s} - A_{t+1}}  }^2 } }
    \lesssim  \sigma_{\max} \sum_{t=1}^{ \bar{t} }   \frac{1}{t \sqrt{t} }  \Gamma_{\max} t   
    \lesssim  \sigma_{\max} \Gamma_{\max} \sqrt{ \bar{t} }  \ . 
\end{equation}
For the second half of the sum in \eqref{eq:fourth_first_sim}, we have 
\begin{eqnarray}
    & &   \sigma_{\max} \sum_{t= \bar{t}+1}^{T-1}   \frac{1}{t \sqrt{t} } \sqrt{ \E{ \paren{ \sum_{s=1}^{t}  \inner{\theta_*, A_{s} - A_{t+1}}  }^2 } }   \nonumber \\ 
    &=&   \sigma_{\max} \sum_{t= \bar{t}+1}^{T-1}   \frac{1}{t \sqrt{t} } \sqrt{ 
    \E{ \paren{ \sum_{s=1}^{t}  \inner{\theta_*, A_{s} - A_{t+1}}  }^2 \Big| ~ G } \p{ G } 
   +  \E{ \paren{ \sum_{s=1}^{t}  \inner{\theta_*, A_{s} - A_{t+1}}  }^2 \Big| ~ G^{\complement} } \p{ G^{\complement} } 
    }  \nonumber  \\ 
    &\stackrel{(a)}{=}&  \sigma_{\max} \sum_{t= \bar{t}+1}^{T-1}   \frac{1}{t \sqrt{t} } \sqrt{ 
    \E{ \paren{ \sum_{s=1}^{\bar{t}}  \inner{\theta_*, A_{s} - A_{t+1}}  }^2 \Big| ~ G } \p{ G } 
   +  \E{ \paren{ \sum_{s=1}^{t}  \inner{\theta_*, A_{s} - A_{t+1}}  }^2 \Big| ~ G^{\complement} } \p{ G^{\complement} } 
    }  \nonumber \\ 
    &\stackrel{(b)}{\lesssim}&  \sigma_{\max} \sum_{t= \bar{t}+1}^{T-1}   \frac{1}{t \sqrt{t} }  \bar{t} \Gamma_{\max}^{} 
    +  \sigma_{\max} \sum_{t= \bar{t}+1}^{T-1}   \frac{1}{t \sqrt{t} } t \Gamma_{\max} \sqrt{ \p{ G^{\complement} }  }  \nonumber \\
    &\lesssim& \sigma_{\max}    \Gamma_{\max}  \bar{t} \cdot \paren{ \frac{1}{\sqrt{ \bar{t}}}  - \frac{1}{\sqrt{T}} } 
    +  \sigma_{\max}   \Gamma_{\max}  \paren{ \sqrt{ T } - \sqrt{\bar{t}} } \sqrt{\delta}   \ , 
\end{eqnarray}
where step (a) is due to the fact that there is only one arm left provided that the event $G$ happens. 
Inequality (b) follows since $\inner{\theta_*, A_{s} - A_{t+1}}  \leq \Gamma_{\max}$ and $\sqrt{a+b}\leq \sqrt{a}+\sqrt{b}$.

\end{enumerate}

Putting these together, we conclude that 
\begin{eqnarray}
    \E{\reg_\pi(T)}  &\lesssim&   \bar{t}  \log (T) \cdot \Gamma_{\max}^2   + \sqrt{\bar{t}} \cdot  \sigma_{\max} \Gamma_{\max}  
    +  \frac{1}{\Delta} \hat{C}^2 d^3 \log(d) \log^2 \left(\frac{ T }{\delta}\right)
    + d^2 \Delta 
       \nonumber \\
    & & + \sqrt{T \delta}   \sigma_{\max}   \Gamma_{\max} + T \cdot \delta \Gamma_{\max}^2 + T  \cdot   \delta \Delta_{\max} \ . 
\end{eqnarray}

We further note that 
\begin{eqnarray}
     \bar{t}  = \sum_{\ell = 1}^{ \lceil \log_2(2 / \Delta_a ) \rceil }  
    n_{(\ell)}  &=& \sum_{\ell = 1}^{ \lceil \log_2(2 / \Delta_a ) \rceil }  
    \sum_{a \in \mathcal{A}_{\ell} }  \left\lceil \hat{C}^2 \frac{d^{3}   \log(d) Q_{\ell}^{*} (a) }{\varepsilon_{\ell}^{2}} \log^2 \left( \frac{  T }{\delta} \right)\right\rceil  \nonumber \\
    &\leq&  \sum_{\ell = 1}^{ \lceil \log_2(2 / \Delta_a ) \rceil }  
    \sum_{a \in \mathcal{A}_{\ell} }  \paren{ \hat{C}^2 \frac{d^{3}   \log(d) Q_{\ell}^{*} (a) }{\varepsilon_{\ell}^{2}} \log^2 \left( \frac{  T }{\delta} \right) + 1 }  \nonumber \\ 
    &\lesssim& \sum_{\ell = 1}^{ \lceil \log_2(2 / \Delta_a ) \rceil }  
     \paren{ \hat{C}^2 \frac{d^{3}   \log(d)  }{\varepsilon_{\ell}^{2}} \log^2 \left( \frac{  T }{\delta} \right) + d^2 }  \nonumber \\
    &\lesssim& \frac{1}{\Delta^2} d^3 \log(d) \log^2 \left( \frac{  T }{\delta} \right) + d^2 \log_2 \paren{\frac{1}{\Delta}}   \nonumber \ . 
\end{eqnarray}

Therefore, by setting $\delta=1/T$, we have 
\begin{eqnarray}
    \E{\reg_\pi(T)}  &\lesssim& \paren{ \Gamma_{\max}^{2} \frac{1}{\Delta^2} + \frac{1}{\Delta}  }  d^3 \log(d) \log^3(T)  + d^2 \log_2 \paren{\frac{1}{\Delta}} + d^2 \Delta  \ . \nonumber  
\end{eqnarray}

\newpage

\section{Supplementary Materials for Section~\ref{sec:numerical} }  \label{sec:figures_tables}

\subsection{Figures and Tables}

\begin{table}[htbp] 
  \centering
  \caption{Model parameters used for the experiment}
    \begin{tabular}{cccccc}
    \toprule
          & PSX    & BX   & NASDAQ & Dark pool 1 & Dark pool 2 \\
    \midrule
    $p$   & - 1.1439  & - 1.4709  &  - 0.07  & -0.05    & - 0.01 \\
    $\tilde{p}$  & 0.0008  & -0.000001   & -0.0002  & 0.0003    & 0.0002 \\
    $\sigma^2$  & -0.031  & - 0.000001   & -0.000004  & -0.0191    & -0.0091 \\   
    $\tilde{\sigma}^2$  & 0.0391 &  0.000001 & 0.00056 &  0.06  &  0.05 \\
    \bottomrule
    \end{tabular}%
  \label{tab:parameters_2}%
\end{table}%

\begin{figure}[htbp]
	\centering
	\subfigure[BX  mean with R-squared $R^2=0.99$]{\includegraphics[width=8cm,height=5.5cm]{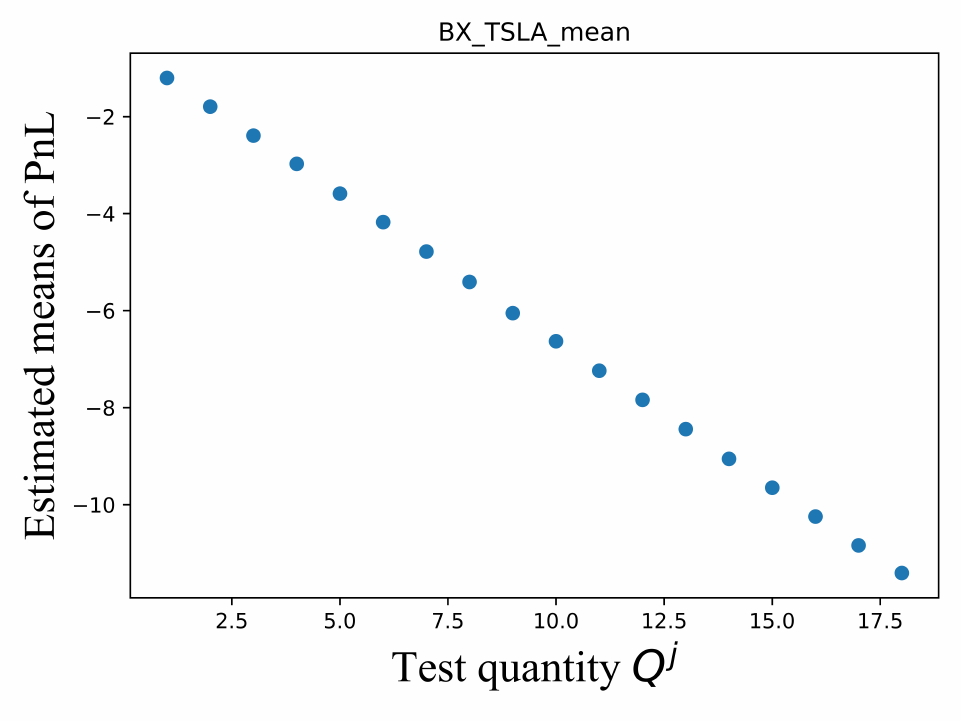} \label{fig:BX_TSLA_mean} }
	\subfigure[BX  var with R-squared $R^2=0.94$]{\includegraphics[width=8cm,height=5.5cm]{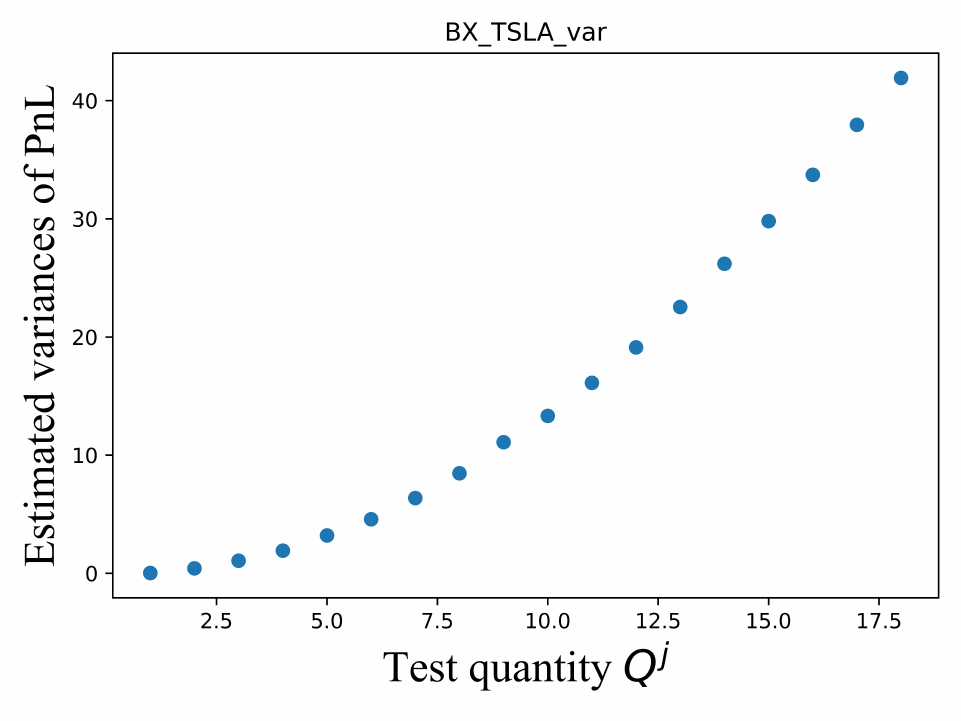} \label{fig:BX_TSLA_var}	}
	\subfigure[PSX  mean with R-squared $R^2=0.99$]{\includegraphics[width=8cm,height=5.5cm]{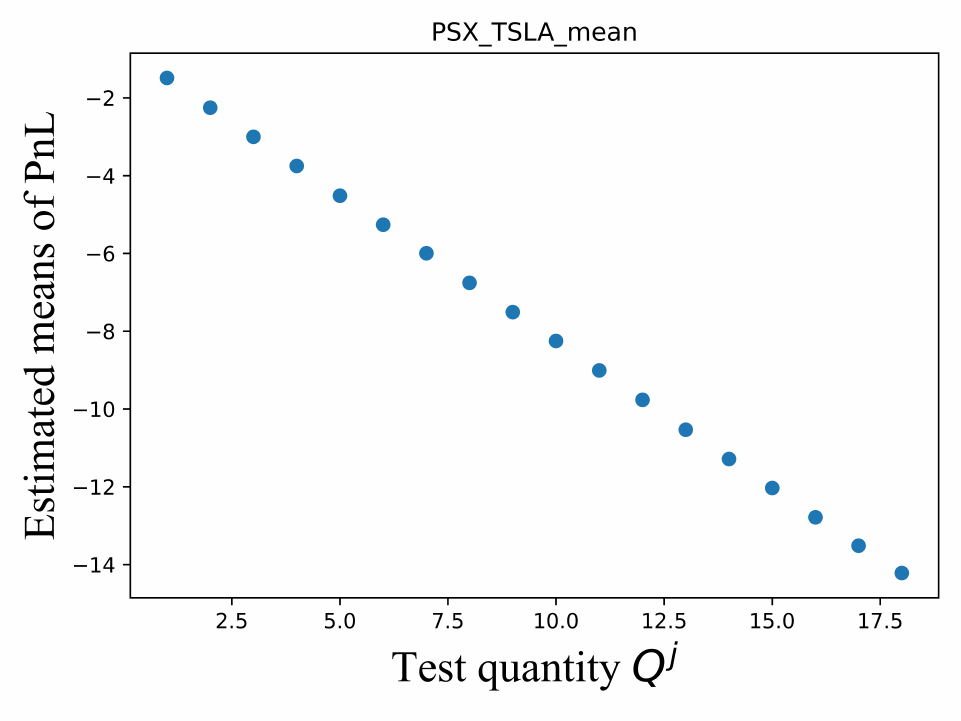} \label{fig:PSX_TSLA_mean}	}
	\subfigure[PSX  var]{\includegraphics[width=8cm,height=5.5cm]{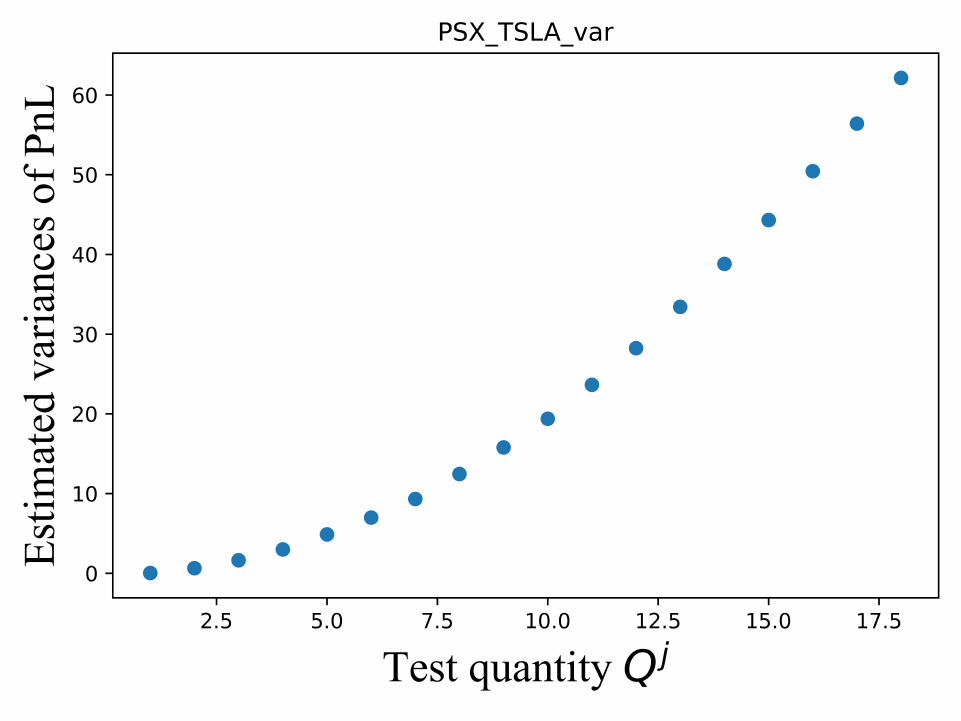} \label{fig:PSX_TSLA_var with R-squared $R^2=0.94$}	}
	\subfigure[NASDAQ  mean with R-squared $R^2=0.99$]{\includegraphics[width=8cm,height=5.5cm]{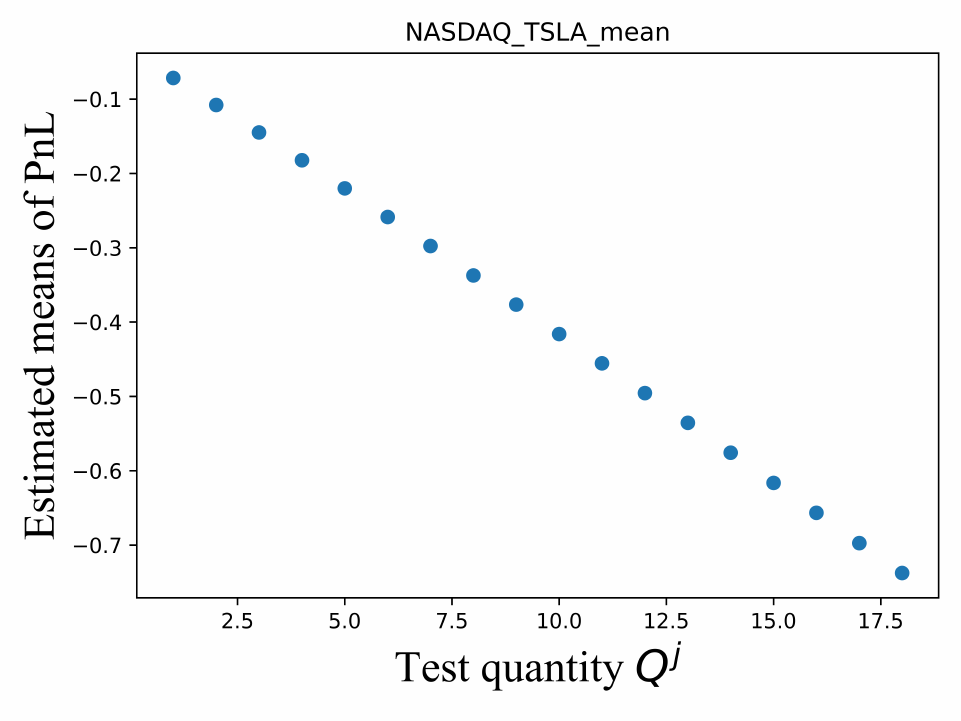} \label{fig:NASDAQ_TSLA_mean} }
	\subfigure[NASDAQ  var with R-squared $R^2=0.93$]{\includegraphics[width=8cm,height=5.5cm]{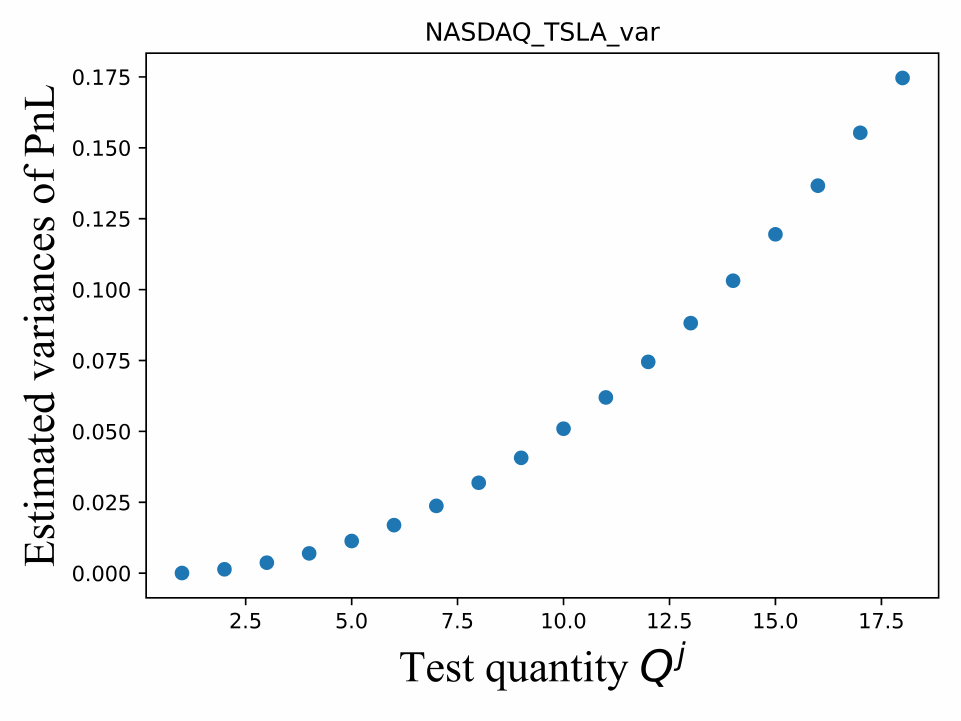} \label{fig:NASDAQ_TSLA_var}	}
	\caption{
The estimated means and variances of PnL for Tesla stock at different venues, as $Q^j$ varies.
A clear linear relationship prevails for mean and a linear relationship for variance roughly holds.
This is an empirical justification for our linear form approximation Eq.~\eqref{eq:sor_mean_model} and Eq.~\eqref{eq:sor_var_model}. 
 }
	\label{fig:tsla}
\end{figure}

In Figure~\ref{fig:tsla} and Figure~\ref{fig:spy}, we validate the modeling assumptions specified in \eqref{eq:sor_mean_model} and \eqref{eq:sor_var_model} using real data, considering the case where all $ n $ units are allocated to a single venue while the others receive none.  
Let $ a(n) $ denote this allocation strategy. We plot $ \widehat{\mathbb{E}} [ X_t(a(n)) ] $ against $ n $ and $ \widehat{\operatorname{Var}} ( X_t(a(n)) ) $ against $ n^2 $, respectively.  
The data sampling methodology used to estimate these quantities is detailed in Section~\ref{sec:study_realdata}.  
Our results confirm that the reward structure is well captured by linear and quadratic terms in the allocation vector.

\subsection{Detailed Discussion on Dependence of $\Delta$ on $K$}
\label{sec:discussion_gap_K}

The section focuses on the combinatorial action set that arises in the SOR applications (allocating $S$ shares to $d$ venues) in Section \ref{sec:model_SOR_linear_bandit} where the MV has a quadratic structure. We show that in this case, under a fixed margin condition, the gap $\Delta$ does not deteriorate merely because $K$ is large.

To recall, we use the action set  
\begin{equation} \label{eq:action_set_sor_example}
    \mathcal{A} = \braces{ a \in \mathbb{Z}_{\geq 0}^{d} : \sum_{i=1}^{d} a_i = S } ,
\end{equation}
where $a_i$ is the $i$th coordinate of the vector $a$. 
One can verify that $\| a \|_2 \leq S$ for all $a \in \mathcal{A}$.
The number of actions is
\begin{equation}
    K=\binom{S+d-1}{d-1} . 
\end{equation}

Following the arguments in Section \ref{sec:model_SOR_linear_bandit}, we assume that MV has the following structure
\begin{equation} \label{eq:MV_sor_example}
    \operatorname{MV}(a) = \sum_{k=1}^d\left(\alpha_k a_k+\beta_k a_k^2\right) ~\text{ for }~  
    a \in \mathbb{Z}_{\geq 0}^d, ~ \sum_{k=1}^d a_k=S . 
\end{equation}
The coefficients $\alpha_k$'s and $\beta_k$'s are scalars. 
\begin{remark}[Connection with the example in Section~\ref{sec:model_SOR_linear_bandit}]\label{remark:connection_example}
We clarify that, throughout this section, the action stems from the allocation vector over the venues belonging to the set defined in \eqref{eq:action_set_sor_example}. The MV in \eqref{eq:MV_sor_example} contains both the linear and squared allocation terms $a_k$ and $a_k^2$. In Section~\ref{sec:model_SOR_linear_bandit}, the squared terms $a_k^2$ directly show up in the action vector. Namely, we considered the following action set 
$$ \left\{(a, v) \in \mathbb{Z}_{\geq 0}^d \times \mathbb{Z}_{\geq 0}^d: \sum_{i=1}^d a_i=S, v_i=a_i^2 ~~\forall i\right\} .  $$
However, these two representations are equivalent. For notational convenience, we adopt \eqref{eq:action_set_sor_example} and  \eqref{eq:MV_sor_example} in this section. 
\end{remark}

We note that this aligns with the structure we used in the numerical experiments. 
For the following discussion, to ease the computation, we impose the condition that $\beta_k \geq 0$.
This condition reflects that trading more through a single venue increases its volatility, capturing the empirically observed “volume–volatility” relationship.

Let $e_k \in \mathbb{R}^d$ be the $k$th standard Euclidean basis vector. 
For venue $k$, the one-share addition at action $a$ changes MV by
\begin{equation}
    c_k^{+}(a) \defeq \operatorname{MV}\left(a+   e_k \right)-\mathrm{MV}(a)=\alpha_k+\beta_k\left(2 a_k+1\right),
\end{equation}
and one-share removal at action $a$ changes MV for venue $k$ by 
\begin{equation}
    c_k^{-}(a) \defeq \mathrm{MV}\left(a-e_k\right) - \mathrm{MV}(a) 
    = - \alpha_k - \beta_k\left(2 a_k-1\right) \quad \text{ for action } a \text{ such that } a_k \geq 1  . 
\end{equation}
Let $a^* = \arg\min_{a \in \mathcal{A}} ~ \operatorname{MV}(a) $ be the optimal allocation. 
We define the margin at $a^*$ to be 
\begin{equation} \label{eq:margin}
    \gamma 
    \defeq \min _{\substack{i: a_i^* \geq 1 \\ j \neq i}} ~ c_j^{+}(a^*)+c_i^{-}(a^*)
\end{equation}
and \textit{assume} that $\gamma > 0$. 
We note that $\gamma$ is guaranteed to be nonnegative due to the optimality of $a^*$. 
The strict positive gap $\gamma > 0$ can also be  satisfied by many instances. 
For example, let $S=d$, $\alpha_k=\alpha$ for all $k$, and $\beta_k = \beta~(>0)$ for all $k$, then, the optimal action is $a^*=(1,1,\ldots,1)$ and $\gamma = 2 \beta > 0$.

\begin{claim}
    Under the fixed margin assumption that $\gamma > 0$, for any $a \neq a^*$, we have 
    \begin{equation}
        \Delta_a = \operatorname{MV}(a)-\operatorname{MV}\left(a^*\right) \geq \gamma . 
    \end{equation}
\end{claim}

\begin{myproof}
    
Let $\delta \defeq a - a^* \in \mathbb{R}^d$.
For each venue $k$, we denote $\delta_k \defeq a_k-a_k^* \in \mathbb{R}$, and 
$\delta_k^{+} = \max \left\{0, \delta_k\right\},  \delta_k^{-} = \max \left\{0,-\delta_k\right\}$.
One can verify that 
\begin{equation} \label{eq:halp_diff_a_a_star}
    \sum_{k=1}^d \delta_k^{+} = \sum_{k=1}^d \delta_k^{-} = \frac{1}{2} \| a - a^* \|_1 .
\end{equation}
Note we have $\| a - a^* \|_1 \geq 2$ for $a \neq a^*$. 
For brevity, we denote $g_k(x) \defeq \alpha_k x + \beta_k x^2$. 
For given venue $k$, if $\delta_k > 0$, then by telescoping sum, we can write 
\begin{eqnarray}
    g_k(a_k) - g_k(a_k^*) &=& \sum_{i=0}^{\delta_k^+ -1} \brackets{ g_k(a_k^* + i + 1) - g_k(a_k^* + i) }  \nonumber  \\ 
    &=& \sum_{i=0}^{\delta_k^+ -1} \brackets{ \alpha_k + \beta_k \paren{ 2 a_k^* + 1 } + 2 \beta_k i }  \nonumber \\
    &=& \delta_k^+ \cdot \brackets{ \alpha_k + \beta_k \paren{ 2 a_k^* + 1 } } + 2 \beta_k \sum_{i=0}^{\delta_k^+ -1} i \nonumber   \\
    &=& \delta_k^{+} c_k^{+}\left(a_k^*\right)+\beta_k \delta_k^{+}\left(\delta_k^{+}-1\right) ,
    \label{eq:diff_g_pos}
\end{eqnarray}
where in the last equality we denote (with slight abuse of notation)
$$
c_k^{+}\left(a_k^*\right) = \alpha_k + \beta_k \paren{ 2 a_k^* + 1 } .
$$
Similarly, for venue $k$ with $\delta_k < 0$, then we have 
\begin{equation} \label{eq:diff_g_neg}
    g_k\left(a_k\right)-g_k\left(a_k^*\right)
    = \delta_k^{-} c_k^{-}\left(a_k^*\right)+\beta_k \delta_k^{-}\left(\delta_k^{-}-1\right) , 
\end{equation}
where we denote (with slight abuse of notation)
$$
c_k^{-}\left(a_k^*\right) = - \alpha_k - \beta_k \paren{ 2 a_k^* - 1 } .
$$

Therefore, combining \eqref{eq:diff_g_pos} and \eqref{eq:diff_g_neg} we have
\begin{eqnarray}
    \operatorname{MV}(a)-\operatorname{MV}\left(a^*\right)
    &=&  \sum_{k=1}^{d}  \brackets{ g_k(a_k) - g_k(a_k^*) } \\ 
    &=&  \sum_{k=1}^{d}  \Big [  \paren{ \delta_k^{+} c_k^{+}\left(a_k^*\right)+\beta_k \delta_k^{+}\left(\delta_k^{+}-1\right) } \nonumber \\
    & & +  \paren{ \delta_k^{-} c_k^{-}\left(a_k^*\right)+\beta_k \delta_k^{-}\left(\delta_k^{-}-1\right) } \Big ] . 
\end{eqnarray}

Since $\left|\delta_k\right|=\delta_k^{+}+\delta_k^{-}$, and only one of $\delta_k^{+}$ and $\delta_k^{-}$ is nonzero, we have 
$$
\delta_k^{+}\left(\delta_k^{+}-1\right)+\delta_k^{-}\left(\delta_k^{-}-1\right)=\left|\delta_k\right|\left(\left|\delta_k\right|-1\right) . 
$$
Moreover, note that $\delta_k \in \mathbb{Z}$ and since we assume $\beta_k \geq 0$, it is always true that 
\begin{equation}
    \sum_{k=1}^d \beta_k\left|\delta_k\right|\left(\left|\delta_k\right|-1\right) \geq 0 .
\end{equation}
Hence, we conclude that 
\begin{eqnarray}
    \operatorname{MV}(a)-\operatorname{MV}\left(a^*\right)
    &=& \sum_{k=1}^d \delta_k^{+} c_k^{+}\left(a_k^*\right) 
    + \sum_{k=1}^d \delta_k^{-} c_k^{-}\left(a_k^*\right)
    + \sum_{k=1}^d \beta_k\left|\delta_k\right|\left(\left|\delta_k\right|-1\right) \nonumber \\ 
    &\geq& \sum_{k=1}^d \delta_k^{+} c_k^{+}\left(a_k^*\right) 
    + \sum_{k=1}^d \delta_k^{-} c_k^{-}\left(a_k^*\right) .
\end{eqnarray}

To proceed, we define the sets of indices 
\begin{equation}
    P=\left\{j: a_j>a_j^*\right\}, \quad N=\left\{i: a_i<a_i^*\right\} .
\end{equation}
Namely, $P$ is the set of indices of coordinates where $a$ has strictly more units than $a^*$, and $N$ is the set of indices where $a$ has strictly fewer units than $a^*$. 
There must exist nonnegative integers $\{ m_{i, j} \}_{ (i,j) \in N \times P } $'s such that 
\begin{equation}
    \sum_{j \in P} m_{i, j}=a_i^*-a_i=\delta_i^- \quad \forall i \in N , 
\end{equation}
and 
\begin{equation}
    \sum_{i \in N} m_{i, j}=a_j-a_j^* =\delta_j^+ \quad \forall j \in P . 
\end{equation}
Namely, $m_{i, j}$ denotes the number of shares moved from venue $i \in N$ to $j \in P$ when transforming the vector $a^*$ to $a$. 

Then, we can write
\begin{equation} \label{eq:m_pos}
    \sum_{k=1}^d \delta_k^{+} c_k^{+}\left(a^*\right)
    =\sum_{j \in P} \left(\sum_{i \in N} m_{i, j}\right) ~ c_j^{+}\left(a^*\right)
    =\sum_{i \in N, j \in P} m_{i, j} ~ c_j^{+}\left(a^*\right),
\end{equation}
\begin{equation} \label{eq:m_neg}
    \sum_{k=1}^d \delta_k^{-} c_k^{-}\left(a^*\right)
    = \sum_{i \in N}\left(\sum_{j \in P} m_{i, j}\right) ~ c_i^{-}\left(a^*\right)
    = \sum_{i \in N, j \in P} m_{i, j} ~ c_i^{-}\left(a^*\right) .
\end{equation}
Summing up \eqref{eq:m_pos} and \eqref{eq:m_neg} yields 
\begin{equation}
    \sum_{k=1}^{d}  \delta_k^{+} c_k^{+}\left(a^*\right)
    + \sum_{k=1}^{d}  \delta_k^{-} c_k^{-}\left(a^*\right)
    = \sum_{i \in N, j \in P} m_{i, j}\left(c_j^{+}\left(a^*\right)+c_i^{-}\left(a^*\right)\right) .
\end{equation}
Recalling the definition of margin in \eqref{eq:margin}, we have
\begin{equation}
    \sum_{i \in N, j \in P} m_{i, j}\left(c_j^{+}\left(a^*\right)+c_i^{-}\left(a^*\right)\right)
    \geq \gamma \sum_{i \in N, j \in P} m_{i, j}
    = \gamma \sum_{i \in N} \delta_i^{-} 
    =  \frac{\gamma}{2} \| a - a^* \|_1 ,
\end{equation}
where the last equation follows from \eqref{eq:halp_diff_a_a_star}.
Hence, we conclude that the gap $\Delta$ is of order $\Omega(\gamma)$.

\end{myproof}

\newpage

\section{A Non-Stationary Setting}
\label{appendix:non-stationary}

In this section, we consider a piecewise stationary setting similar to \cite{garivier2008upper, yu2009piecewise, auer2008near}, to showcase that our analysis can be potentially extended to more challenging settings than the stationary setting. 
This setting is not only well-studied in the literature but also aligns with practical industry practices, where models are typically retrained on a weekly basis or over even longer intervals. Such retraining schedules naturally accommodate potential shifts in the environment, further supporting the relevance of our proposed approach.

As we will see, Proposition~\ref{prop:temporal_decomposition} serves as a critical foundation for the analysis.
In a non-stationary setting, the best arm changes frequently, which makes arm-wise analysis infeasible.
This challenge posed by the changing nature of the environment can be effectively addressed by the temporal decomposition  because it tracks regret over time rather than per arm.

\textbf{Problem Formulation. }
For simplicity, we assume that the variances of different arms are the same. 
We note that it is possible to extend the analysis to a more general setting where the variances of different actions are not identical, following the approach outlined in the main body.
In what follows, we assume that the reward at time $t$ is generated by 
\begin{equation}
    X_t = \inner{\theta_t, A_t} + \eta_t \ , 
\end{equation}
where we assume the noise is conditionally $\sigma$-subgaussian. Precisely, $ \E{\eta_t \big | \mathcal{F}_t} = 0$ and
$$
\E{ \exp \paren{ t \eta_t } ~\big |~ \mathcal{F}_t} \leq \exp \paren{ \frac{t^2 \sigma^2}{2}}~~a.s.
$$
Recall $\mathcal{F}_t = \sigma (A_1, X_1, \cdots, A_t, X_t)$.
We assume that $\normtwo{A_t} \leq L_a, \normtwo{\theta_t} \leq L_{\theta}$. 
The coefficient $\theta_t$ is allowed to change (abruptly) up to $\Upsilon_T$ many times throughout the horizon of $T$ steps. 
We assume that we have the knowledge of $T$ and $\Upsilon_T$ in advance.
Note that $\Upsilon_T$ serves as an upper bound on the number of changes, making its knowledge a reasonable assumption.
Given a policy $\pi$, the expected MV is  
\begin{eqnarray}
    & & \E{ \sum_{t=1}^T\left(X_t^\pi-\frac{1}{T}\sum_{s=1}^T X_{s}^\pi \right)^2 - \rho \sum_{t=1}^{T} X^{\pi}_{t}}   \ .  \nonumber 
\end{eqnarray}
In this nonstationary environment, instead of the best single-arm policy, it is more sensible to compare with a policy that adaptively selects the best arm based on the mean reward, taking into account the changing coefficient.  
Namely, at time $t$, it selects $A_t^*$ according to 
\begin{equation}
    A_t^* = \arg \max_{a \in \mathcal{A} } ~ \inner{ \theta_t , a  } \ . 
\end{equation}


\begin{algorithm}  \label{algo:non_stationary_main}
\caption{Non-Stationary ExpExp with GSE Exploration }

    \begin{algorithmic}[1]
    \label{algo:nonstationary}
    \Require Total time horizon $T$, 
    cycle length $H = n + \tilde{n}$, 
    number of cycles $P = \lceil T / H \rceil$, 
    elimination parameter $\kappa$, budget $n$ for exploration,
    action set $\mathcal{A}$
    
    \For{each cycle $p = 1, \dots, P$}
        \State \textbf{Exploration Phase:} Use GSE Algorithm~\ref{algo:gse} to identify the best arm $\hat{a}_p$ for cycle $p$
        \State \textbf{Exploitation Phase:} Play the best arm $\hat{a}_p$ for $\tilde{n}$ steps
    \EndFor
    
    \end{algorithmic}
\end{algorithm}

\begin{algorithm}
\caption{GSE: Generalized Successive Elimination}
\label{algo:gse}
\begin{algorithmic}[1]

\Require Elimination hyper-parameter $\kappa=2$, budget $n$, action set $\mathcal{A}$
\State \textbf{Initialization:} $\mathcal{A}_1 \gets \mathcal{A}$, $t \gets 1$, $s \gets \lceil \log_{2} K \rceil$ where $K = |\mathcal{A}|$ 

\While{$t \leq s$}
    \State \textbf{Projection:} Project $\mathcal{A}_t$ to $d_t$ dimensions, such that $\mathcal{A}_t$ spans $\mathbb{R}^{d_t}$
    \State \textbf{Exploration:} Explore $\mathcal{A}_t$ using the G-optimal design $\Pi_t$
    \State \textbf{Estimation:} Calculate $(\hat{\mu}_{i,t})_{i \in \mathcal{A}_t}$ based on observed rewards 
    \State \textbf{Elimination:} Update action set:
    \[
    \mathcal{A}_{t+1} = \underset{\mathcal{A} \subseteq \mathcal{A}_t : |\mathcal{A}| = \lfloor |\mathcal{A}_t| / 2 \rfloor}{\arg\max} \sum_{i \in \mathcal{A}} \hat{\mu}_{i,t}
    \]
    \State $t \gets t + 1$
\EndWhile

\State \textbf{Output:} Best arm $\hat{a} = \mathcal{A}_{s+1}$

\end{algorithmic}
\end{algorithm}

\begin{figure}
    \centering
    \includegraphics[width=0.9\linewidth]{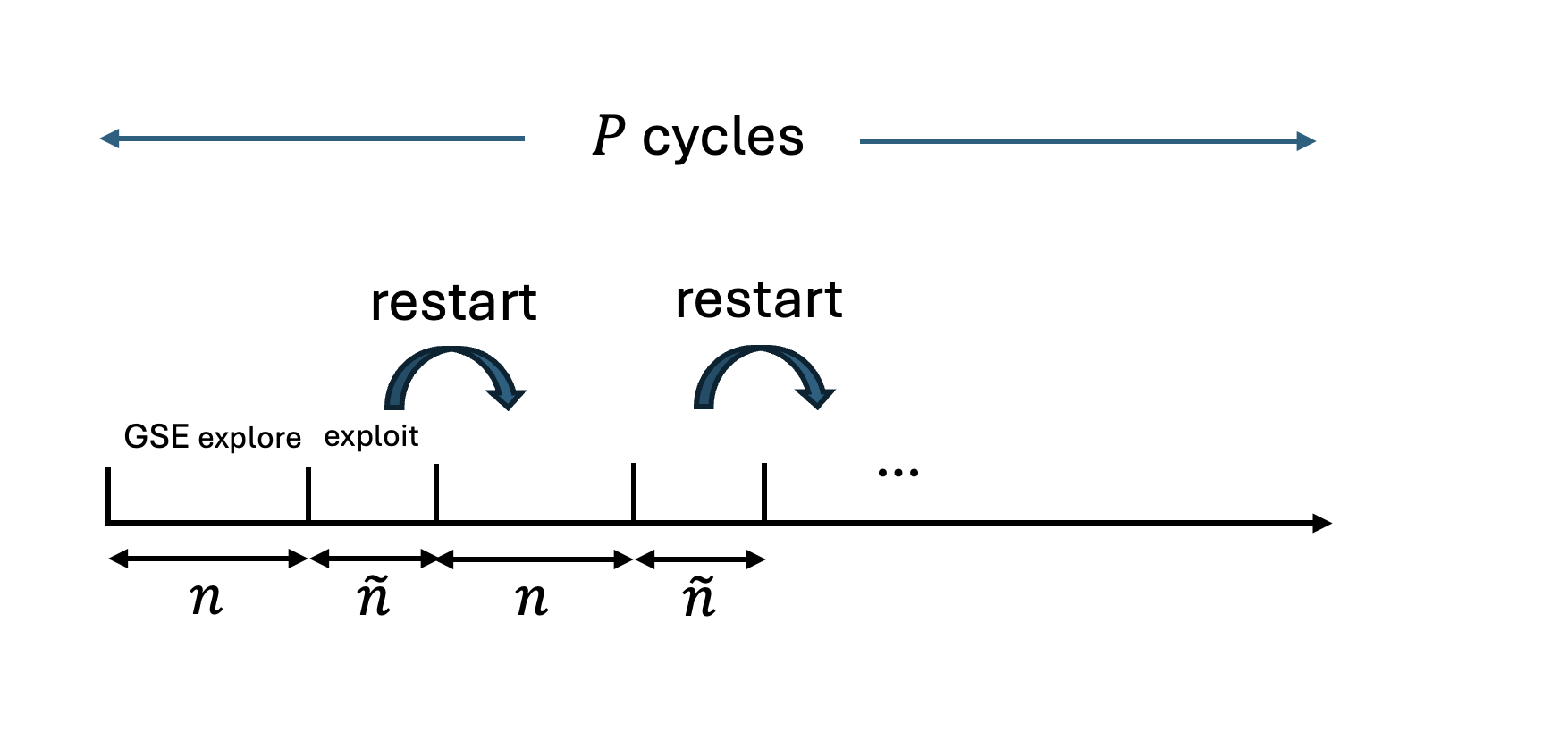}
    \caption{Illustration of Non-Stationary ExpExp with GSE Exploration}
    \label{fig:nonstationary_algo}
\end{figure}

\textbf{Algorithm Design. }
The overall design of the algorithm is depicted in Figure~\ref{fig:nonstationary_algo}. 
Given the piecewise stationary structure in time, where the $\theta_t$'s are assumed to change abruptly, we restart the algorithm every $H$ steps to adapt to this non-stationarity. The $T$ rounds are evenly divided into $P = \lceil \frac{T}{H} \rceil$ cycles, each comprising $H$ steps. We refer to these $H$ steps as a cycle, which includes both exploration and exploitation phases.

Within each cycle, we run the best-arm-identification algorithm Generalized Successive Elimination (GSE) proposed by \cite{azizi2021fixed} with a budget of $n$ and an elimination parameter $\kappa=2$ during the exploration phase, followed by exploiting the identified arm for $\tilde{n}$ steps, which we detail in Algorithm~\ref{algo:nonstationary}.
The budget $n$ is split evenly over $\lceil \log_2 (K) \rceil$ pieces. In each piece, GSE pulls arms for $ \lfloor \frac{n}{\lceil \log_{2} (K) \rceil} \rfloor$ times and eliminates $ \frac{1}{2} $ fraction of them at the end.

\begin{theorem}
   If $
\Delta_{\min}^{2}  \geq  \frac{\sigma^2 d \log(K)}{ \log (\frac{\log(K)}{\log(T)} ) + \frac{4}{9} \log(T)  } \frac{1}{T^{\frac{2}{9}}} 
$, the expected regret of Algorithm~\ref{algo:nonstationary} is at most 
\begin{equation}
   \E{\reg_{}(T)} \lesssim L_a^2 L_\theta^2 \cdot \paren{T^{ \frac{8}{9} } \log(T) 
    +   \Upsilon_T  T^{\frac{2}{3}} 
    + \sigma^2 \sqrt{\Upsilon_T T }}  \ . 
\end{equation}
\end{theorem}

\begin{myproof}
    
To set the stage, we introduce some notations first. 
Let $\mathcal{N}_p$ denote the set of time indices corresponding to the exploration phase of the $p$-th cycle, and let $\tilde{\mathcal{N}}_p$ represent the set of time indices corresponding to the exploitation phase of the $p$-th cycle.
Let $\abs{ \mathcal{N}_p } = n, \abs{ \tilde{\mathcal{N}}_p } = \tilde{n}$, and hence the length of a complete cycle is given by $H = n + \tilde{n}$.

Let $\Theta \in \mathbb{R}^T$ denote a random sequence of coefficients.
Consider a particular realization of $\Theta$, denoted as $\bftheta = [\theta_1, \theta_2, \cdots, \theta_T]$.
For the purpose of the following discussion, we treat $\bftheta$ as deterministically given, though we do not know its value.
Let $S_p(\bftheta)$ denote the event that there is no change in cycle $p$. 
Namely, $\one{S_p (\bftheta) }=1$ if and only if $\braces{\theta_t ~|~ t \in \mathcal{N}_p \cup \tilde{\mathcal{N}}_p } $ is a singleton set.  
For brevity, we may suppress the dependency on $\bftheta$ when it is clear from the context. 

We define the good event $G$ to be  
\begin{equation}
    G( \bftheta ) = \braces{
        \forall ~p \in [P] : ~ \paren{ \one{S_p} = 1 } \implies \left( \forall t \in \tilde{\mathcal{N}}_p, \, a_t = a_t^* \right)   ~\big|~ \Theta = \bftheta
    } \ .
\end{equation}
Let us unpack the definition. 
Under event $G$, we require that the event that there is no change of coefficient in cycle $p$ implies that we identify the ``correct" arm in that particular cycle. 
The randomness of this event comes from the noises and the algorithm. 
In words, event $G$ corresponds to the scenario in which, during each cycle without a change, the algorithm successfully identifies the correct arm.


Using Theorem~1 in \cite{azizi2021fixed}, the probability of the complement event $G(\Theta)^\complement$ is bounded as: 
\begin{eqnarray}
    \p{  G( \Theta )^\complement } &=& 
    \p{  \exists ~p \in [P] : ~ \paren{ \one{S_p} = 1 } \implies \left( \exists ~ t \in \tilde{\mathcal{N}}_p, \, a_t \neq a_t^* \right)   ~\big|~ \Theta = \bftheta  
    }  \nonumber \\ 
    &\leq& \sum_{p=1}^{P} \one{ S_p } 
    \p{ \exists~ t \in \tilde{\mathcal{N}}_p  ~ a_t \neq a_t^*   ~\big|~ \Theta = \bftheta }  \\ 
    &\leq&  P \cdot \kappa \log(K) \exp \paren{ - \frac{n \Delta_{\min}^2}{\sigma^2 d \log(K)} }    \ ,
\end{eqnarray}
where $\Delta_{\min}$ is an instance-dependent parameter. It is the minimum gap between the mean of the best arm and the second-best arm. Precisely, 
\begin{equation}
    \Delta_{\min} = \min_{p} \min_{a,b \in \mathcal{A}} \abs{ \theta_p^{\top} (a-b)} \ . 
\end{equation}

\textbf{The Mean Term.} 
First, we estimate the regret from the mean term. 
\begin{eqnarray}
    \E{ \sum_{t=1}^{T} X_t^* - X_t^\pi }   &=&  \sum_{t=1}^{T} \E{ \inner{ \theta_t, a_t^* - a_t} }  \nonumber \\ 
    &=& \E{ \sum_{p=1}^{P}
    \paren{  \sum_{t \in \mathcal{N}_p } 
    + \sum_{ t \in \tilde{\mathcal{N}}_p }  }
    \paren{ \inner{ \theta_t, a_t^* - a_t } }
    }  \nonumber \\ 
    &\lesssim&  L_a L_\theta n P + \E{ \sum_{p=1}^{P} \sum_{ t \in \tilde{\mathcal{N}}_p }  \inner{ \theta_t, a_t^* - a_t }   }  \label{eq:non_mean_1}  \ . 
\end{eqnarray}
The inequality follows from the simple fact that 
$$
\inner{ \theta_t, a_t^* - a_t } \leq \normtwo{\theta_t} \normtwo{a_t^* - a_t} \leq 2 L_a L_\theta \ . 
$$
Conditioned on the event $G$, we have 
\begin{eqnarray}
    & &   \E{ \sum_{p=1}^{P} \sum_{ t \in \tilde{\mathcal{N}}_p }  \inner{ \theta_t, a_t^* - a_t }  }  \nonumber  \\ 
    &=&  \E{ \sum_{p=1}^{P} \sum_{ t \in \tilde{\mathcal{N}}_p }  \inner{ \theta_t, a_t^* - a_t } ~\Big|~ G  }  \p{G}
    +  \E{ \sum_{p=1}^{P} \sum_{ t \in \tilde{\mathcal{N}}_p }  \inner{ \theta_t, a_t^* - a_t } ~\Big|~ G^\complement  } \p{G^\complement} \nonumber  \\ 
    &\lesssim& \E{ \sum_{p=1}^{P} \sum_{ t \in \tilde{\mathcal{N}}_p }  \inner{ \theta_t, a_t^* - a_t } ~\Big|~ G  } 
    +  L_a L_\theta  T \delta  \ ,   \label{eq:non_mean_2}
\end{eqnarray}
where in the last inequality, we pessimistically upper bound $ \inner{ \theta_t, a_t^* - a_t }$ by $2 L_a L_\theta$. 
We further note that the former term in \eqref{eq:non_mean_2} can be upper bounded by 
\begin{eqnarray}
    && \E{ \sum_{p=1}^{P} \sum_{ t \in \tilde{\mathcal{N}}_p }  \inner{ \theta_t, a_t^* - a_t } ~\Big|~ G  }  \nonumber \\ 
    &=&  \E{ \sum_{p=1}^{P} 
    \paren{ \one{S_p} \sum_{ t \in \tilde{\mathcal{N}}_p }  \inner{ \theta_t, \underbrace{a_t^* - a_t}_{=0}  }
    + \one{S_p^\complement} \sum_{ t \in \tilde{\mathcal{N}}_p }  \inner{ \theta_t, a_t^* - a_t } }
    ~\Big|~ G  }   \nonumber \\
    &=& \E{ \sum_{p=1}^{P} 
    \one{S_p^\complement} \sum_{ t \in \tilde{\mathcal{N}}_p }  \inner{ \theta_t, a_t^* - a_t } 
    ~\Big|~ G  } \nonumber  \\ 
    &\lesssim&  L_a L_\theta \Upsilon_T H  \ .   \label{eq:non_mean_3}
\end{eqnarray}
The inequality follows from observing that conditioned on event $G$, we can still pessimistically upper bound $\inner{ \theta_t, a_t^* - a_t}$ by $2 L_a L_\theta$ within cycles in which there is any change in coefficient.

Combining \eqref{eq:non_mean_1}, \eqref{eq:non_mean_2} and \eqref{eq:non_mean_3}, we conclude that 
\begin{equation}
    \E{ \sum_{t=1}^{T} X_t^* - X_t^\pi } 
      \lesssim   L_a L_\theta \cdot \paren{ n P +  \Upsilon_T H  +   T \delta }  \ . 
\end{equation}


\textbf{The Variance Term.} 
In what follows, we carefully study the variance term
\begin{eqnarray}
    & & \E{ \sum_{t=1}^T\left(X_t^\pi-\frac{1}{T}\sum_{s=1}^T X_{s}^\pi \right)^2 } 
    - \E{ \sum_{t=1}^T\left(X_t^*-\frac{1}{T}\sum_{s=1}^T X_{s}^*\right)^2 } \nonumber   \\ 
    &=&  \sum_{t=1}^{T-1} \E{ \frac{1}{t(t+1)} \paren{ t X_{t+1} - \sum_{s=1}^{t} X_s }^2 } 
     - \sum_{t=1}^{T-1} \E{ \frac{1}{t(t+1)} \paren{ t X_{t+1}^* - \sum_{s=1}^{t} X_s^* }^2 } \nonumber \\ 
    &=&   \sum_{t=1}^{T-1} \Big (   \E{  \frac{1}{t(t+1)} \paren{ t \inner{\theta_{t+1}, a_{t+1}} -  \sum_{s=1}^{t} \inner{\theta_s, a_s} }^2
    +  \frac{1}{t(t+1)}  \paren{ t \eta_{t+1} - \sum_{s=1}^{t} \eta_s   }^2  }  \nonumber \\ 
    & & -  \E{  \frac{1}{t(t+1)} \paren{ t \inner{\theta_{t+1}, a^*_{t+1}} -  \sum_{s=1}^{t} \inner{\theta_s, a^*_s} }^2
    +  \frac{1}{t(t+1)}  \paren{ t \eta_{t+1} - \sum_{s=1}^{t} \eta_s   }^2 }  \nonumber \\
    && + \E{  \frac{2}{t(t+1)}  \paren{t \inner{\theta_{t+1}, a_{t+1}} -  \sum_{s=1}^{t} \inner{\theta_s, a_s}  } \paren{  t \eta_{t+1} - \sum_{s=1}^{t} \eta_s  }
    }
    \nonumber \\
    && - \E{   \frac{2}{t(t+1)}  \paren{t \inner{\theta_{t+1}, a^*_{t+1}} -  \sum_{s=1}^{t} \inner{\theta_s, a^*_s}  } \paren{  t \eta_{t+1} - \sum_{s=1}^{t} \eta_s  }
    }  \Big )  \nonumber  \\
    &=&  \sum_{t=1}^{T-1} \E{  \frac{1}{t(t+1)} \paren{ t \inner{\theta_{t+1}, a_{t+1}} -  \sum_{s=1}^{t} \inner{\theta_s, a_s} }^2 -  \frac{1}{t(t+1)} \paren{ t \inner{\theta_{t+1}, a^*_{t+1}} -  \sum_{s=1}^{t} \inner{\theta_s, a^*_s} }^2   } \label{eq:big_term_1}  \\
    & &  +  2 \sum_{t=1}^{T-1}  \frac{1}{t(t+1)} \E{  \paren{  t \inner{\theta_{t+1},  a_{t+1}^{*} - a_{t+1}^{} } - \sum_{ s=1 }^{t }  \inner{\theta_s,  a_s^* - a_s}  }  \paren{ \sum_{s=1}^{t}  \eta_s  }  } 
  \label{eq:big_term_2}  \ . 
\end{eqnarray}
We decompose the variance in terms of time using Lemma~\ref{lemma:variance_decomposition} in the first equation. 
In the second equation, we plug in the definiton of $X_t$. 
The third equation follows since 
\begin{equation}
   \E{ \E{  \paren{ \inner{\theta_{t+1}, a_{t+1} } - \sum_{s=1}^{t} \inner{\theta_s, a_s}  } \eta_{t+1}  ~\Big|~ \mathcal{F}_t } }
    = \E{  \paren{ \inner{\theta_{t+1}, a_{t+1} } - \sum_{s=1}^{t} \inner{\theta_s, a_s}  } \E{   \eta_{t+1}  ~\Big|~ \mathcal{F}_t } }
    = 0 \ . 
\end{equation}
Again, we emphasize that the expectation is taken with respect to the randomness induced by the noise and the algorithm, rather than the coefficients.
The term \eqref{eq:big_term_1} arises from the variation of coefficients and actions, while \eqref{eq:big_term_2} is also attributed to noises. 
In what follows, we analyze these two terms separately. 
\begin{enumerate}
    \item 
We further upper bound \eqref{eq:big_term_1} with two terms. 
\begin{small}    
\begin{eqnarray}
    & &  \eqref{eq:big_term_1}        \nonumber  \\ 
    &=& \sum_{t=1}^{T-1} \mathbb{E} \Bigg [ \frac{1}{t(t+1)} \paren{ t \inner{\theta_{t+1}, a_{t+1}} -  \sum_{s=1}^{t} \inner{\theta_s, a_s} - t \inner{\theta_{t+1}, a^*_{t+1}} +  \sum_{s=1}^{t} \inner{\theta_s, a^*_s}    }  \nonumber \\ 
    & &      \cdot \paren{   t \inner{\theta_{t+1}, a_{t+1}} -  \sum_{s=1}^{t} \inner{\theta_s, a_s}  + t \inner{\theta_{t+1}, a^*_{t+1}} -  \sum_{s=1}^{t} \inner{\theta_s, a^*_s}  }  \Bigg ]   \nonumber \\ 
    &\lesssim& L_a L_\theta \sum_{t=1}^{T-1} \E{ \frac{1}{t+1} \abs{ t \inner{ \theta_{t+1}, a_{t+1} - a^*_{t+1} } -  \sum_{s=1}^{t} \inner{\theta_s, a_s - a_s^*}  }   }  \\ 
    &=&  L_a L_\theta \sum_{t=1}^{T-1} 
    \E{ \frac{1}{t+1} \abs{
    \sum_{s=1}^{t} 
    \paren{ \inner{ \theta_{t+1} - \theta_{s}, a_{t+1} - a^*_{t+1} } 
    + \inner{ \theta_s,  a_{t+1} - a^*_{t+1} }
    -  \inner{\theta_s, a_s - a_s^*} } }  } \\ 
    &\leq& L_a L_\theta  \sum_{t=1}^{T-1} \frac{1}{t+1} 
    \E{ \sum_{s=1}^{t} 
    \abs{ \inner{ \theta_{t+1} - \theta_{s}, a_{t+1} - a^*_{t+1} } } } \nonumber \\ 
    & & +  L_a L_\theta \sum_{t=1}^{T-1} \frac{1}{t+1} 
    \E{ \sum_{s=1}^{t} 
    \abs{  \inner{ \theta_s,  a_{t+1} - a^*_{t+1} }
    -  \inner{\theta_s, a_s - a_s^*} } } \ ,   \label{eq:nonstationary_action_2_terms}
\end{eqnarray}
\end{small}
where in the first inequality, we simply bound each inner product term by $L_a L_\theta$. 

For the first sum in the display above, we note that 
\begin{small}
\begin{eqnarray}
    & & \sum_{t=1}^{T-1} \frac{1}{t+1} 
    \E{ \sum_{s=1}^{t} 
    \abs{ \inner{ \theta_{t+1} - \theta_{s}, a_{t+1} - a^*_{t+1} } } }  \nonumber \\
    &=& \sum_{t=2}^{T} \frac{1}{t} 
    \E{ \sum_{s=1}^{t-1} 
    \abs{ \inner{ \theta_{t} - \theta_{s}, a_{t} - a^*_{t} } }  }  \nonumber \\ 
    &\overset{(a)}{=}& \sum_{p=1}^{P}  \paren{ \sum_{ t \in \mathcal{N}_p } + \sum_{ t \in \tilde{\mathcal{N}}_p } } 
    \paren{ \frac{1}{t} \sum_{s=1}^{t-1} 
    \E{ \abs{ \inner{ \theta_{t} - \theta_{s}, a_{t} - a^*_{t} } }  }  }   \nonumber  \\ 
&\overset{(b)}{\lesssim}& L_a L_\theta   \sum_{p=1}^{P} n + 
      \sum_{p=1}^{P} \sum_{ t \in \tilde{\mathcal{N}}_p }
       \paren{ \frac{1}{t} \sum_{s=1}^{t-1} 
    \E{ \abs{ \inner{ \theta_{t} - \theta_{s}, a_{t} - a^*_{t} } }  }  }  \nonumber \\ 
&\overset{(c)}{=}& L_a L_\theta    n P
    + \sum_{p=1}^{P} \one{S_p} \sum_{ t \in \tilde{\mathcal{N}}_p }
    \paren{ \frac{1}{t} \sum_{s=1}^{t-1} 
 \E{ \abs{ \inner{ \theta_{t} - \theta_{s}, a_{t} - a^*_{t} } }  }  } 
    + \sum_{p=1}^{P} \one{S_p^\complement} \sum_{ t \in \tilde{\mathcal{N}}_p }
 \paren{ \frac{1}{t} \sum_{s=1}^{t-1} 
\E{ \abs{ \inner{ \theta_{t} - \theta_{s}, a_{t} - a^*_{t} } }  }  }   \nonumber   \\ 
&\overset{(d)}{\lesssim}& L_a L_\theta    n P 
    + \sum_{p=1}^{P} \one{S_p} \sum_{ t \in \tilde{\mathcal{N}}_p }
    \paren{ \frac{1}{t} \sum_{s=1}^{t-1} 
 \E{ \abs{ \inner{ \theta_{t} - \theta_{s}, a_{t} - a^*_{t} } }  }  } 
    + L_a L_\theta \Upsilon_T H   \nonumber \\
    &\overset{(e)}{=}& L_a L_\theta  \cdot \paren{ n P + \Upsilon_T H  } 
    \nonumber \\ 
    & &    + \sum_{p=1}^{P} \one{S_p} \sum_{ t \in \tilde{\mathcal{N}}_p }
    \paren{ \frac{1}{t} \sum_{s=1}^{t-1} 
 \E{ \abs{ \inner{ \theta_{t} - \theta_{s}, \underbrace{a_t^* - a_t}_{=0}  } } ~\Big |~ G } \p{G} 
 + \E{ \abs{ \inner{ \theta_{t} - \theta_{s}, a_{t} - a^*_{t} } } ~\Big |~ G^\complement } \p{G^\complement}   }    \nonumber \\ 
    &\lesssim& L_a L_\theta  \cdot \paren{ n P + \Upsilon_T H + T \delta  }  \ . 
\end{eqnarray}
\end{small}
In equation (a), we decompose $T$ steps to $P$ exploration-then-exploitation cycles. 
In (b), we simply use the fact 
$$
\inner{ \theta_{t} - \theta_{s}, a_{t} - a^*_{t} }
\leq \normtwo{ \theta_{t} - \theta_{s} } 
\normtwo{a_{t} - a^*_{t}} 
\leq 4 L_a L_\theta \ . 
$$
for all $t \in \mathcal{N}_p$. 
In step (c), we categorize the cycles based on whether a change occurs within the cycle.
In step (d), for all cycles where changes occur, we conservatively bound the regret by $L_a L_\theta \Upsilon_T H$. 
In step (e), conditioned on event \(G\), we can accurately identify the correct arm during a cycle with no changes, ensuring that \(a_t = a_t^*\) during exploitation.


For the second sum in \eqref{eq:nonstationary_action_2_terms}, we observe that conditioned on $G$, 
\begin{eqnarray}
& &  \sum_{t=1}^{T-1} \frac{1}{t+1} 
    \E{ \sum_{s=1}^{t} 
    \abs{  \inner{ \theta_s,  a_{t+1} - a^*_{t+1} }
    -  \inner{\theta_s, a_s - a_s^*} } ~\Big |~ G  }   \nonumber \\ 
&=& \sum_{t=2}^{T} \frac{1}{t} 
    \E{ \sum_{s=1}^{t-1} 
    \abs{  \inner{ \theta_s,  a_{t} - a^*_{t} }
    -  \inner{\theta_s, a_s - a_s^*} } ~\Big |~ G  }   \nonumber   \\ 
&\overset{(a)}{=}& \sum_{p=1}^{P} \mathbb{E} \brackets{  \paren{ \sum_{t \in \mathcal{N}_p } + \sum_{ t \in \tilde{\mathcal{N}}_p }  } \paren{ \frac{1}{t} \sum_{s=1}^{t-1}
    \abs{  \inner{ \theta_s,  a_{t} - a^*_{t} }
    -  \inner{\theta_s, a_s - a_s^*} } }  ~\Big |~ G   }  \nonumber \\ 
&\overset{(b)}{\lesssim}&  
    \sum_{p=1}^{P} \paren{ L_a L_\theta \cdot n 
    + \sum_{t \in \tilde{\mathcal{N}}_p }  \frac{1}{t} \sum_{s=1}^{t-1} \E{ \abs{ 
    \inner{ \theta_s,  a_{t} - a^*_{t} }
    -  \inner{\theta_s, a_s - a_s^*} } ~\Big |~ G  } } \nonumber \\ 
&=& L_a L_\theta n P 
+ \sum_{p=1}^{P} \one{S_p} \sum_{t \in \tilde{\mathcal{N}}_p }  \frac{1}{t} \sum_{s=1}^{t-1} \E{ \abs{ 
    \inner{ \theta_s,  \underbrace{a_{t} - a^*_{t}}_{=0} }
    -  \inner{\theta_s, a_s - a_s^*} } ~\Big |~ G  }  \nonumber \\ 
& & + \underbrace{\sum_{p=1}^{P} \one{S_p^\complement} \sum_{t \in \tilde{\mathcal{N}}_p }  \frac{1}{t} \sum_{s=1}^{t-1} \E{ \abs{ 
    \inner{ \theta_s,  a_{t} - a^*_{t} }
    -  \inner{\theta_s, a_s - a_s^*} } ~\Big |~ G  } }_{ \lesssim L_a L_\theta \Upsilon_T H } \nonumber \\
&\overset{(c)}{\lesssim}&    L_a L_\theta \cdot \paren{ n P + \Upsilon_T H } 
+  \sum_{p=1}^{P} \one{S_p} \sum_{t \in \tilde{\mathcal{N}}_p }  \frac{1}{t} \sum_{k=1}^{p-1}  \sum_{s \in \mathcal{N}_k }  \E{ 
  \underbrace{ \abs{ \inner{\theta_s, a_s - a_s^*} } }_{\lesssim L_a L_\theta }  ~\Big |~ G  }  \nonumber   \\ 
&\lesssim&    L_a L_\theta \cdot \paren{ n P + \Upsilon_T H } 
+  \sum_{p=1}^{P}  \sum_{t \in \tilde{\mathcal{N}}_p }  \frac{1}{t} (p-1) n  \nonumber  \\
&\lesssim& L_a L_\theta \cdot \paren{ n P + n P^2 \log(T)   +  \Upsilon_T  H }   \ . 
\end{eqnarray}
In (a), we decompose $T$ steps to $P$ exploration-then-exploitation cycles. 
In (b), we pessimistically bound the summand by $L_a L_\theta $ for each $t$ in exploration phases. 
For (c), again we use the fact that, for cycles when there are changes, we pessimistically bound the regret by $L_a L_\theta \Upsilon_T H$. 
Conditioned on event $G$, we are able to identify the correct arm in a cycle when there is no change, and hence $a_t = a_t^*$ in exploitation. 
Finally, we simply note that on the complement, 
$$
\sum_{t=1}^{T-1} \frac{1}{t+1} 
    \E{ \sum_{s=1}^{t} 
    \abs{  \inner{ \theta_s,  a_{t+1} - a^*_{t+1} }
    -  \inner{\theta_s, a_s - a_s^*} } ~\Big |~ G^\complement  }
\lesssim L_a L_\theta T \delta \ . 
$$

    \item 
Regarding the term \eqref{eq:big_term_2}, we decompose it in the following way:  
\begin{eqnarray}
   & & \frac{1}{2} \eqref{eq:big_term_2} \nonumber \\ 
    &=& \sum_{t=1}^{T-1}  \frac{1}{t(t+1)} \E{  \paren{  
  \sum_{s=1}^{t} \inner{ \theta_{t+1} - \theta_s , a_{t+1}^{*} - a_{t+1}^{} } 
  + \sum_{s=1}^{t} \inner{ \theta_s , a_{t+1}^{*} - a_{t+1}^{} - (a_s^* - a_s) }
   }  \paren{ \sum_{s=1}^{t}  \eta_s  }  }  \nonumber \\
   &\leq& \sum_{t=1}^{T-1}  \frac{1}{t(t+1)} \E{  \abs{  
  \sum_{s=1}^{t} \inner{ \theta_{t+1} - \theta_s , a_{t+1}^{*} - a_{t+1}^{} } 
   }  \abs{ \sum_{s=1}^{t}  \eta_s  }  }   \label{eq:noise_1}  \\ 
   && + \sum_{t=1}^{T-1}  \frac{1}{t(t+1)} \E{ 
   \abs{  
  \sum_{s=1}^{t} \inner{ \theta_s , a_{t+1}^{*} - a_{t+1}^{} - (a_s^* - a_s) }
   } 
   \abs{ \sum_{s=1}^{t}  \eta_s  }  }  \label{eq:noise_2}
\end{eqnarray}
To bound \eqref{eq:noise_1}, it is helpful to introduce the following notations. 
The coefficient $\theta_t$ remains constant for the first $m_1$ time steps, then changes to a new value for the next $m_2$ steps, and so on. Formally,
\begin{equation} 
\theta_1 = \theta_2 = \cdots = \theta_{m_1}, \quad 
\theta_{m_1+1} = \theta_{m_1+2} = \cdots = \theta_{m_1+m_2}, \quad 
\theta_{m_1+m_2+1} = \cdots     \nonumber 
\end{equation}
with each block of constant values differing from the previous one, i.e., 
$\theta_{m_1} \neq \theta_{m_1+1}, \; \theta_{m_1+m_2} \neq \theta_{m_1+m_2+1}, \; \dots$
Now, we make the following observations: 
\begin{small}
\begin{eqnarray}
    && \sum_{t=1}^{T-1}  \frac{1}{t(t+1)} \E{  \abs{  
  \sum_{s=1}^{t} \inner{ \theta_{t+1} - \theta_s , a_{t+1}^{*} - a_{t+1}^{} } 
   }  \abs{ \sum_{s=1}^{t}  \eta_s  }  }  \nonumber \\ 
   &\lesssim& L_a \sum_{t=1}^{T-1}  \frac{1}{t(t+1)}    
  \sum_{s=1}^{t} 
  \normtwo{\theta_{t+1} - \theta_s} 
     \E{  \abs{ \sum_{s=1}^{t}  \eta_s  }  }   \\ 
  &\overset{(a)}{\lesssim}& \sigma^2 L_a \sum_{t=1}^{T-1}  \frac{1}{\sqrt{t} (t+1)}    
  \sum_{s=1}^{t} 
  \normtwo{\theta_{t+1} - \theta_s}  \\ 
  &=& \sigma^2  L_a \sum_{t=1}^{m_1-1}  \frac{1}{\sqrt{t} (t+1)}    
 \cdot 0 
 + L_a \sum_{t=m_1}^{m_1+m_2-1}  \frac{1}{\sqrt{t} (t+1)}   \sum_{s=1}^{m_1}  \normtwo{\theta_{t+1} - \theta_s}  \nonumber \\ 
 && +   L_a \sum_{t=m_1+m_2}^{m_1+m_2+m_3-1}  \frac{1}{\sqrt{t} (t+1)}   \sum_{s=1}^{m_1+m_2}  \normtwo{\theta_{t+1} - \theta_s}   \nonumber \\
 & & + \cdots + L_a \sum_{t= m_1+m_2+\cdots+ m_{\Upsilon_T-1} }^{m_1+m_2+\cdots+m_{\Upsilon_T}-1}  \frac{1}{\sqrt{t} (t+1)}   \sum_{s=1}^{m_1+m_2+\cdots+m_{\Upsilon_T-1} }  \normtwo{\theta_{t+1} - \theta_s}  \\ 
 &\overset{(b)}{\lesssim}& \sigma^2  L_a L_\theta \cdot \paren{ m_1 \cdot \paren{ \sqrt{\frac{1}{m_1}} - \sqrt{\frac{1}{m_1+m_2}}}  }  
 + L_a L_\theta \cdot \paren{ \paren{m_1+m_2} \cdot \paren{ \sqrt{\frac{1}{m_1+m_2}} - \sqrt{\frac{1}{m_1+m_2+m_3}}}  }   \nonumber \\
 & & + \cdots  +  L_a L_\theta \cdot \paren{  \paren{ m_1+m_2+\cdots+m_{\Upsilon_T-1} } 
 \cdot \paren{ \sqrt{\frac{1}{m_1+\cdots+m_{\Upsilon_T-1}} } 
 - \sqrt{ \frac{1}{ m_1+\cdots+m_{\Upsilon_T} }} }  }    \\
 &\overset{(c)}{=}& \sigma^2  L_a L_\theta \cdot  \paren{ m_1 \sqrt{\frac{1}{m_1}} + m_2 \sqrt{\frac{1}{m_1+m_2}} + \cdots + m_{\Upsilon_T-1} \sqrt{ \frac{1}{ m_1+\cdots+m_{\Upsilon_T-1} }} 
 + m_{\Upsilon_T} \sqrt{ \frac{1}{ m_1+\cdots+m_{\Upsilon_T} }} 
 }  \nonumber \\ 
&& - L_a L_\theta \cdot \paren{ m_1+\cdots+m_{\Upsilon_T-1} +  m_{\Upsilon_T}  } \sqrt{ \frac{1}{ m_1+\cdots+m_{\Upsilon_T} } }  \nonumber \\
&\overset{(d)}{\leq}& \sigma^2  L_a L_\theta \cdot \sqrt{ \paren{ \paren{ \sqrt{ \frac{m_1}{m_1} }  }^2 + \paren{ \sqrt{ \frac{m_2}{m_1+m_2} }  }^2 + \cdots + \paren{ \sqrt{\frac{ m_{\Upsilon_T} }{m_1+\cdots+m_{\Upsilon_T}}} }^2  } \paren{  m_1 + \cdots + m_{\Upsilon_T} } } - L_a L_\theta \cdot \sqrt{T}  \nonumber \\ 
&\leq& \sigma^2  L_a L_\theta  \sqrt{ \Upsilon_T \cdot T}  \ . 
\end{eqnarray}
\end{small}
Inequality (a) follows the same reasoning as \eqref{eq:noise_term_3}. 
Inequality (b) follows since $ \sum_{t=a}^{b} \frac{1}{\sqrt{t} (t+1)} \lesssim \int_{a}^{b} \frac{1}{t^{\frac{3}{2}}} d t = 2 \paren{ \frac{1}{\sqrt{a}} - \frac{1}{\sqrt{b}} }$.
Step (c) holds by telescoping the terms. 
Step (d) is due to Cauchy-Schwarz inequality. 

It remains to bound \eqref{eq:noise_2}, we simply note that 
\begin{eqnarray}
    & & \sum_{t=1}^{T-1}  \frac{1}{t(t+1)} \E{ 
   \abs{  
  \sum_{s=1}^{t} \inner{ \theta_s , a_{t+1}^{*} - a_{t+1}^{} - (a_s^* - a_s) }
   } 
   \abs{ \sum_{s=1}^{t}  \eta_s  }  } \nonumber   \\ 
   &\leq& \sum_{t=1}^{T-1}  \frac{1}{t(t+1)} \E{ 
  \sum_{s=1}^{t} \abs{ \inner{ \theta_s , a_{t+1}^{*} - a_{t+1}^{} - (a_s^* - a_s) }
   } 
   \abs{ \sum_{s=1}^{t}  \eta_s  }  } \nonumber  \\ 
   &\lesssim& L_a L_\theta \sigma^2  \sqrt{T}  \ , 
\end{eqnarray}
where the last step follows the same reasoning as \eqref{eq:noise_term_3}. 
\end{enumerate}

\textbf{Putting the Mean and Variance Together. }
Putting all these together,
by setting $H = T^\beta, n = H^\alpha = T^{\alpha \beta}$,
we conclude that the expected regret is of order 
\begin{eqnarray}
   &&  \E{\reg_{}(T)}  \nonumber \\ 
   &=&  \E{ \sum_{t=1}^T\left(X_t^\pi-\frac{1}{T}\sum_{s=1}^T X_{s}^\pi \right)^2 } 
    - \E{ \sum_{t=1}^T\left(X_t^*-\frac{1}{T}\sum_{s=1}^T X_{s}^*\right)^2 }  
    - \rho \cdot
    \E{ \sum_{t=1}^{T}  X_t^\pi - X_t^*  }   \nonumber \\ 
    &\lesssim& L_a^2 L_\theta^2 \cdot \paren{ n P + n P^2 \log(T)   + \Upsilon_T  H  
    + T \delta  
    + \sigma^2  \sqrt{\Upsilon_T  T } } \nonumber  \\
    &\lesssim& L_a^2 L_\theta^2 \cdot  \paren{
        T^{ 2\paren{1-\beta} + \alpha \beta } \log(T) 
    +   \Upsilon_T  T^{\beta} 
    + \sigma^2 \sqrt{\Upsilon_T T }
    + T^{2-\beta} \log(K) \exp \paren{ - \frac{ T^{\alpha \beta} \Delta_{\min}^2}{\sigma^2 d \log(K)} } 
    }   \ .  \nonumber  
\end{eqnarray}
We recall that $\Delta_{\min}$ is a problem-dependent parameter. 
Whenever 
$$
\Delta_{\min}^{2}  \geq  \frac{\sigma^2 d \log(K)}{ \log (\frac{\log(K)}{\log(T)} ) + \beta (1-\alpha) \log(T)  } \frac{1}{T^{\alpha \beta}} \ , 
$$
the term  $T^{2-\beta} \log(K) \exp \paren{ - \frac{ T^{\alpha \beta} \Delta_{\min}^2}{\sigma^2 d \log(K)} } $ is dominated by  $T^{ 2\paren{1-\beta} + \alpha \beta } \log(T) $. 
Furthermore, if we set $\alpha=\frac{1}{3}, \beta=\frac{2}{3}$, then $T^{ 2\paren{1-\beta} + \alpha \beta } 
= T^{\frac{8}{9}}$; hence 
$$
 \E{\reg_{}(T)} \lesssim L_a^2 L_\theta^2 \cdot \paren{T^{ \frac{8}{9} } \log(T) 
    +   \Upsilon_T  T^{\frac{2}{3}} 
    + \sigma^2 \sqrt{\Upsilon_T T }}  \ . 
$$




Finally, we realize that the above argument holds true for all possible realization of $\bftheta$'s.
The proof is now complete. 
\end{myproof}


\begin{remark} \label{rmk:nonstationary_order_T}
    We expect that there is room for improvement in the $T$-dependence of the current bound $\tilde{O}\left(T^{8/9}\right)$ in the non-stationary setting. Nevertheless, we anticipate that the optimal rate in a non-stationary environment will remain worse than $\tilde{O}(T^{2/3})$, mirroring the risk-neutral case where the optimal rate in non-stationary setting deteriorates relative to the stationary setting. 
    \cite{wei2021non} provide a black-box approach to subsume many existing optimistic bandit and RL algorithms in a stationary setting, and turns it into a parameter-free algorithm with optimal dynamic regret in non-stationary environments. 
    Their analysis does not directly apply in our problem per se, as we are considering MV regret instead of the risk-neutral regret. 
    Nonetheless, the schedule-and-restart scheme proposed therein is conceptually appealing and may prove useful in the MV regret setting as well.   
    We consider this an interesting direction worth future effort.
\end{remark}
 
\vspace{0.5cm}

Next, we implement the algorithm in a setting identical to Section~\ref{subsec:synthetic} except that the environment follows a piecewise-stationary model with abrupt parameter changes (gray dashed lines), while restarts occur every $H$ steps (orange dashed dots) and exploration windows are shaded.
Figure~\ref{fig:nonstationary} shows that the MV regret (shown in (a)) grows most steeply precisely when the per-round MV gap in (b) hovers much beyond zero level (as negative per-round gap increments the regret). 
In (b), we observe that the positive increments are concentrated either in the shaded exploration window, or at abrupt change points. 
Moreover, in panel (b) the increments typically drop toward zero within each exploration window, indicating effective learning.

\begin{figure}[htbp] 
	\centering
	\subfigure[MV regret (ExpExp+GSE) with restarts (orange) and exploration windows (light blue). Grey dashed lines indicate moments when the parameter abruptly changes. ]{\includegraphics[width=8cm,height=5.5cm]{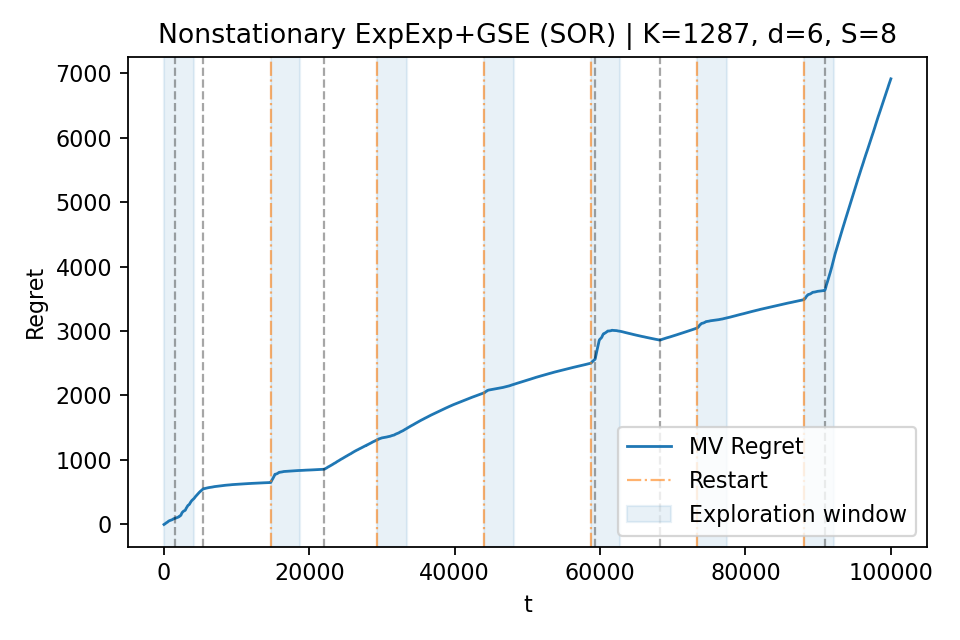} \label{fig:nonstationary_regret} }
	\subfigure[Per-step MV gap compard to the fixed best arm. ]{\includegraphics[width=8cm,height=5.5cm]{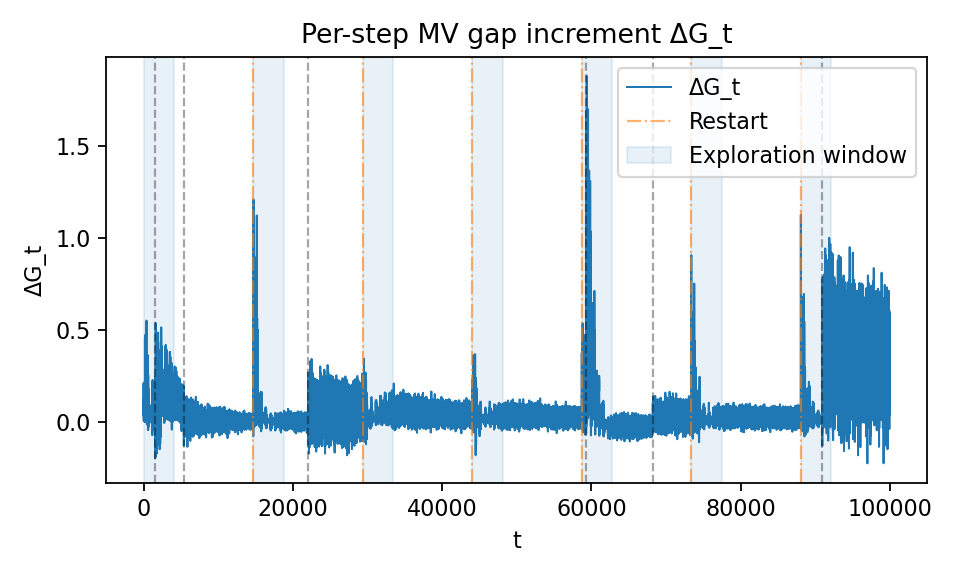} \label{fig:nonstationary_gap_increments}	}
	\caption{
    Performance in the piecewise-stationary (abrupt-change) setting. 
    }
	\label{fig:nonstationary}
\end{figure}

\newpage

\section{Robustness Check: Autocorrelated Noise }
\label{append:auot_cor}

In this section, we numerically demonstrate the robustness of the proposed algorithms to autocorrelated noise.
Unlike the $i.i.d.$ environment (i.e., \eqref{eq:var_def}) considered in the main body, we conduct synthetic experiments where the noise follows a specific autocorrelation structure. 
To be precise, let $\varepsilon_t$ be $i.i.d.$ samples from $\mathcal{N}(0,1)$. 
For $\varrho \in (-1,1)$, we define the autocorrelated noise $\tilde{\eta}_t$ as
\begin{equation}
    \tilde{\eta}_t = \varrho \tilde{\eta}_{t-1} + \sqrt{1-\varrho^2} \, \varepsilon_t \ .
\end{equation}
It is straightforward to verify that $\operatorname{corr}(\tilde{\eta}_t, \tilde{\eta}_{t-1}) = \varrho$. 

Finally, the noise in the rewards for taking action $A_t$ is given by  $ \left( \langle \phi_*, A_t \rangle + \omega \right) \cdot \tilde{\eta}_t $.

\begin{figure}[htbp]
	\centering
\subfigure[$\varrho = -0.6$]{\includegraphics[width=8cm,height=5.5cm]{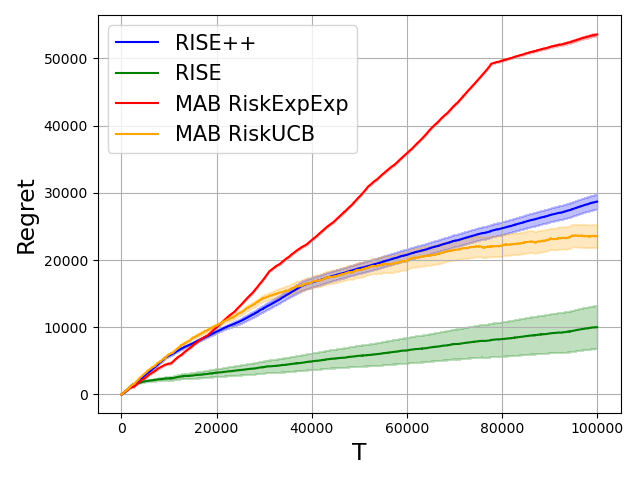} \label{} }
\subfigure[$\varrho = -0.2$]{\includegraphics[width=8cm,height=5.5cm]{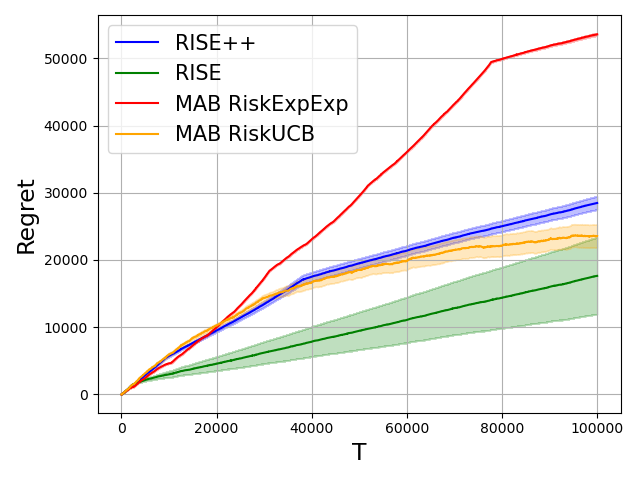} \label{fig:BX_TSLA_var}	}
\subfigure[$\varrho = 0.2$]{\includegraphics[width=8cm,height=5.5cm]{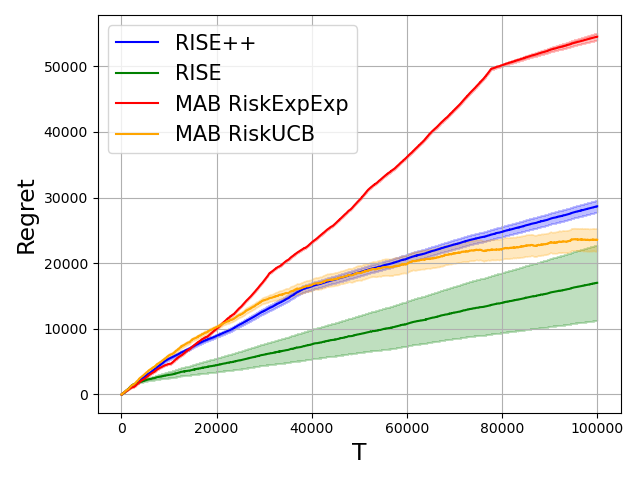} \label{fig:PSX_TSLA_mean}	}
\subfigure[$\varrho = 0.6$]{\includegraphics[width=8cm,height=5.5cm]{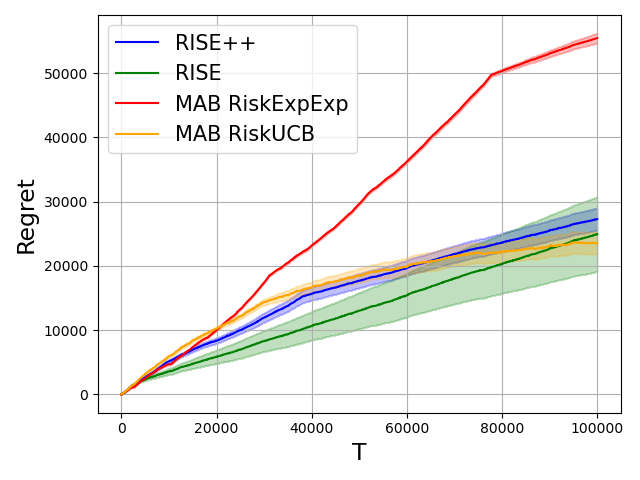} \label{fig:PSX_TSLA_mean}	}
\subfigure[$\varrho = 0$]               
   {\includegraphics[width=8cm,height=5.5cm]{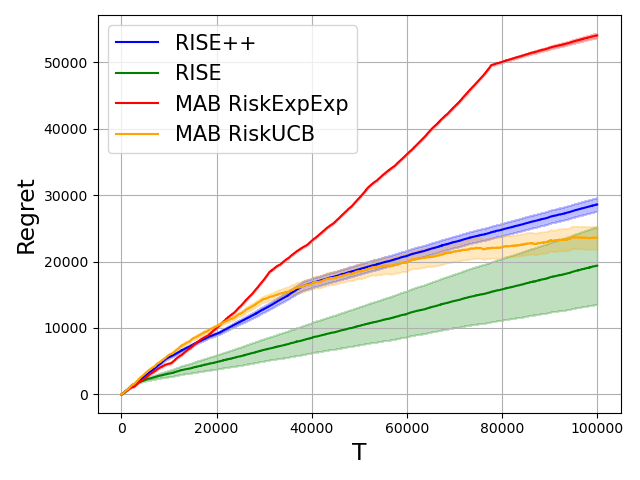} \label{fig:NASDAQ_TSLA_var}	}
\caption{ 
$d=5, S=6$ under correlated noises.   }
	\label{fig:corr_noises}
\end{figure}

We compare the regret of four algorithms: \texttt{RISE++} (blue), \texttt{RISE} (green), \texttt{MAB RiskExpExp} (red), and \texttt{MAB RiskUCB} (orange), under different levels of noise correlation, parameterized by \(\varrho \in \{-0.6, -0.2, 0, 0.2, 0.6\}\). The results demonstrate that \texttt{RISE} consistently achieves the lowest regret across all values of \(\varrho\), highlighting its robustness to autocorrelated noise. \texttt{RISE++} performs comparably to \texttt{RISE}, although it occasionally shows slightly higher regret than \texttt{MAB RiskUCB}, particularly over longer horizons. \texttt{MAB RiskUCB} exhibits competitive performance but is generally outperformed by both \texttt{RISE} and \texttt{RISE++}. In contrast, \texttt{MAB RiskExpExp} performs significantly worse than the other algorithms, with regret that remains high and largely unaffected by changes in \(\varrho\).

The effect of noise correlation varies across scenarios. For positively correlated noise (\(\varrho = 0.2, 0.6\)), \texttt{MAB RiskExpExp} fails to adapt effectively, while negatively correlated noise (\(\varrho = -0.6, -0.2\)) amplifies the performance gap between the proposed algorithms and the benchmarks. In the $i.i.d.$ case (\(\varrho = 0\)), all algorithms show smoother trends, with \texttt{RISE} and \texttt{RISE++} maintaining their advantages over the baselines. These results further demonstrate the robustness of our algorithms to a wide range of noise structures.

\section{Further Discussion of Existing Work}
\label{sec:further_discussion}

In this section, we present a more detailed discussion of existing work on MV formulations in the multi-armed bandit setting.
We give explicit bounds if we apply existing results to our setting.

\cite{sani2012risk} propose a two-step approach. 
First, they relate the MV regret $\reg (T)$ to the pseudo-MV-regret $\tilde{\reg} (T) $ and develop algorithms that provide upper bounds on the pseudo-MV-regret. 
Lemma~1 therein shows that with probability at least $1-\delta$, 
$$
\reg (T)  \leq \tilde{\reg} (T) + (5+\rho) \sqrt{ 2 K T \log(6 T K / \delta) } + 4 \sqrt{2} K \log(6TK / \delta) . 
$$
The additive $\tilde{O}(K)$ term is due to the fact that they decompose the MV regret by arms, and then apply concentration inequalities to each arm. 
Then, the instance-independent bound in Theorem~2 therein guarantees that 
$$
\E{\tilde{\reg} (T) } \leq 2 K T^{2/3} . 
$$
So following the same analysis framework of \cite{sani2012risk} would result in an additive term of order $\tilde{O}(K)$. 

Later, \cite{vakili2016} provide an instance-dependent MV regret bound that reads 
$$
    \E{\reg (T) } \leq  \sum_{ i \neq *} \paren{ \frac{4  \log T}{\min \left\{\Delta_i^2, 4(2+\rho)^2\right\}} + 5 } \paren{ \Delta_i + \Gamma_{i, *}^2} + \sigma_{*}^{2} . 
$$ 
This bound inevitably sums over all arms, leading to an additive order $\tilde{O}(K)$ in the worst case.

Our approach circumvents this issue by deploying the G-optimal design and decomposing the MV regret by time steps (Proposition~\ref{prop:temporal_decomposition}) instead of arms. 
The former allows us to focus the exploration on a small subset of arms; while the latter allows us to avoid the explicit sum over arms.

\end{APPENDICES}

\end{document}